\pdfoutput=1
\documentclass[10pt, logo, copyright]{nvidiatechreport}

\usepackage{mdframed}

\usepackage{graphicx}%
\usepackage{multirow}%
\usepackage{amsmath,amssymb,amsfonts}%
\usepackage{amsthm}%
\usepackage{mathrsfs}%
\usepackage[title]{appendix}%
\usepackage{xcolor}%
\usepackage{textcomp}%
\usepackage{manyfoot}%
\usepackage{booktabs}%
\usepackage{algorithm}%
\usepackage{algorithmicx}%
\usepackage{algpseudocode}%
\usepackage{listings}%

\usepackage{pgfplots}
\pgfplotsset{compat=1.18}
\usepackage{adjustbox}
\usepackage{placeins}

\usetikzlibrary{calc, positioning, shapes.geometric, shadows, backgrounds, fit}

\definecolor{userFill}{RGB}{220, 248, 198} 
\definecolor{userDraw}{RGB}{180, 220, 160}
\definecolor{aiFill}{RGB}{240, 240, 240}   
\definecolor{aiDraw}{RGB}{200, 200, 200}
\definecolor{codeBg}{RGB}{40, 44, 52}      
\definecolor{codeText}{RGB}{171, 178, 191}

\newmdenv[
    backgroundcolor=userFill,
    linecolor=userDraw,
    linewidth=0.8pt,
    roundcorner=8pt,
    leftmargin=0.10\linewidth,
    rightmargin=0pt,
    innerleftmargin=8pt,
    innerrightmargin=8pt,
    innertopmargin=7pt,
    innerbottommargin=7pt,
    skipabove=6pt,
    skipbelow=6pt
]{userbubble}
\newmdenv[
    backgroundcolor=aiFill,
    linecolor=aiDraw,
    linewidth=0.8pt,
    roundcorner=8pt,
    leftmargin=0pt,
    rightmargin=0.04\linewidth,
    innerleftmargin=8pt,
    innerrightmargin=8pt,
    innertopmargin=7pt,
    innerbottommargin=7pt,
    splittopskip=7pt,
    splitbottomskip=7pt,
    skipabove=6pt,
    skipbelow=6pt
]{aibubble}

\title{NV-Reason-CT: 3D Visual Language Model for CT Analysis}

\author[1]{Andriy Myronenko}
\author[1]{Dong Yang}
\author[1]{Yucheng Tang}
\author[2]{Baris Turkbey}
\author[2,3]{Benjamin Simon}
\author[2]{Stephanie Harmon}
\author[2]{Rikhil Makwana}
\author[4]{Mariam Aboian}
\author[5]{Sena Azamat}
\author[6]{Ibrahim Ethem Hamamci}
\author[6]{Sezgin Er}
\author[7]{Bjoern Menze}
\author[8]{Zongwei Zhou}
\author[8]{Wenxuan Li}
\author[1]{Marc Edgar}
\author[1]{Yufan He}
\author[1]{Pengfei Guo}
\author[1]{Daguang Xu}

\correspondingauthor{amyronenko@nvidia.com}

\medskip

\affil[1]{NVIDIA, Santa Clara, CA} 
\affil[2]{NIH, Bethesda, MD}
\affil[3]{University of Oxford, Oxford, UK}
\affil[4]{CHOP/UPenn, Philadelphia, PA}
\affil[5]{Basaksehir Cam and Sakura City Hospital, Istanbul, Turkey}
\affil[6]{Forithmus, San Francisco, CA}
\affil[7]{University of Zurich, Zurich, Switzerland}
\affil[8]{Johns Hopkins University, Baltimore, MD}

\begin{abstract}
We present NV-Reason-CT, a generative vision--language model for chest and
abdominal CT that combines native 3D visual encoding with radiologist-guided
reasoning. CT interpretation requires integrating spatially distributed
findings into a coherent account of anatomy, disease, and diagnostic
uncertainty. Our model couples a native 3D vision transformer with a language
model, passing all encoded visual tokens and their explicit 3D coordinates
into language decoding without further spatial token merging. This design
retains volumetric spatial information both within the vision encoder and
through the language model's positional encoding during joint processing with
text.
\smallskip

We train on a curated corpus of approximately 550,000 multimodal instruction
examples derived from 70,111 unique CT image inputs. The corpus combines
standardized reports, abnormality-focused and anatomy-specific questions,
multi-turn interactions, and radiologist-authored reasoning collected through
recorded and transcribed expert CT interpretations. These expert annotations
provide direct supervision and guide the generation of additional
report-grounded synthetic reasoning. End-to-end supervised fine-tuning (SFT)
is followed by Group Relative Policy Optimization (GRPO), using verifiable
rewards over chest and abdominal abnormality sets.
\smallskip

The model supports direct abnormality classification, report generation, and
interactive reasoning, presenting reviewable observations, differential
diagnoses, and uncertainty alongside its conclusions. Evaluation spans public
CT benchmarks and a held-out NIH cohort. On CT-RATE, NV-Reason-CT achieves a
macro-F1 of 0.614 and a macro-AUROC of 0.871 without a task-specific
classification head; generated reports achieve a report-derived macro-F1 of
0.592. In a
preliminary study with expert radiologists, AI-assisted review received
favorable confidence ratings and was associated with a 50\% reduction in average
reported interpretation and reporting time. We release the model and training
code to support reproducible research on explainable AI for volumetric medical
imaging.

\smallskip
\textbf{GitHub Code:}
\href{https://github.com/NVIDIA-Medtech/NV-Reason-CT}{https://github.com/NVIDIA-Medtech/NV-Reason-CT}

\smallskip
\textbf{Model:} \href{https://huggingface.co/nvidia/NV-Reason-CT}{https://huggingface.co/nvidia/NV-Reason-CT}

\end{abstract}

\begin{document}

\maketitle


\section{Introduction}
\label{introduction}

\begin{figure}[!htbp]
\centering
\includegraphics[width=\textwidth]{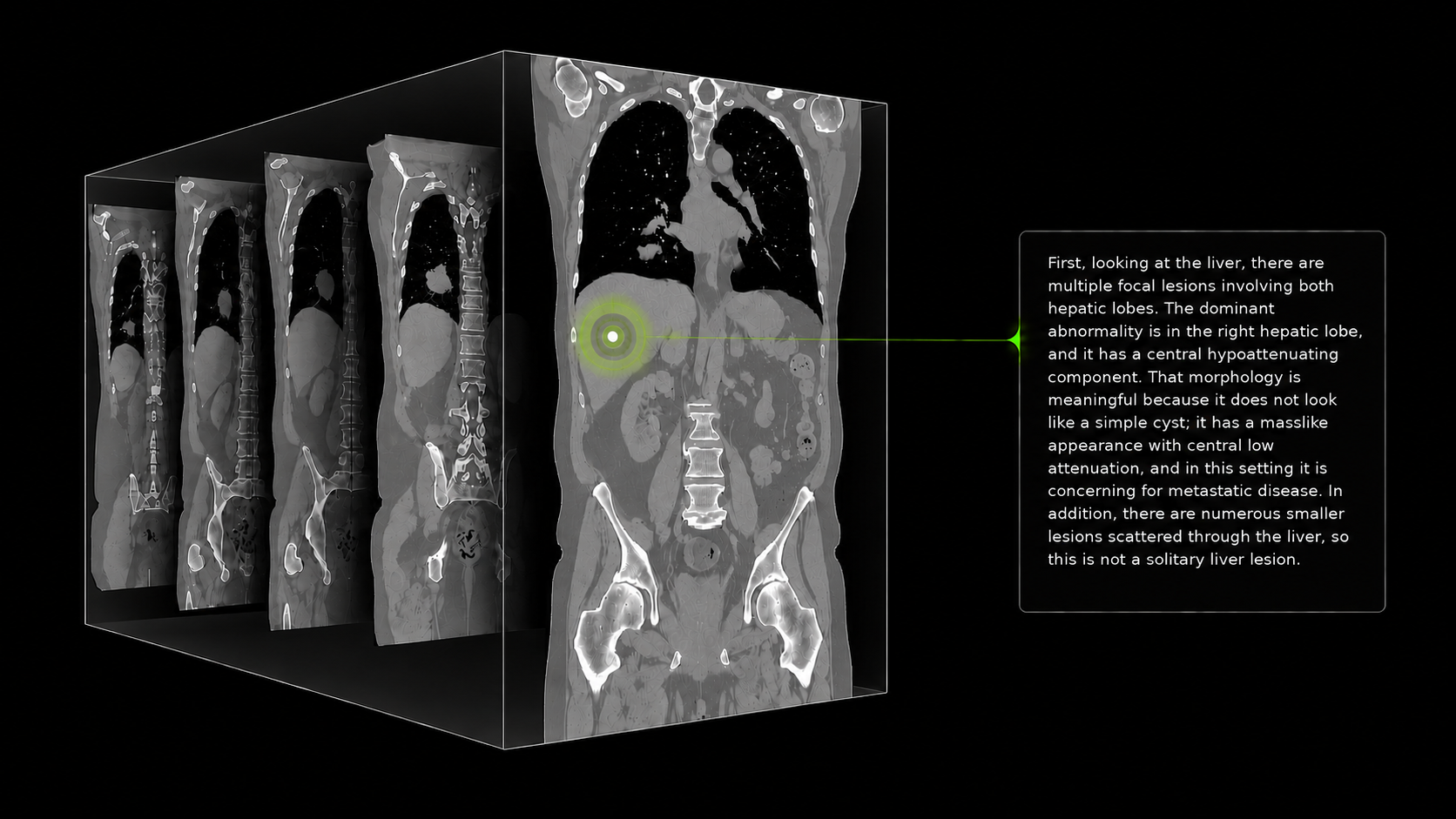}
\caption{A volumetric 3D CT examination is
provided as a native 3D input, and the model produces an image-conditioned
natural-language interpretation. The displayed target marker and example
narrative are illustrative.}
\label{fig:nv_reason_ct_overview}
\end{figure}

Computed tomography (CT) is intrinsically volumetric. A single examination can
contain hundreds of cross-sectional images, and clinically important evidence
depends on the morphology, extent, laterality, and anatomical distribution of
findings across slices. Interpretation requires connecting these observations
across a 3D volume and synthesizing them into a coherent account of normal and
abnormal anatomy, diagnostic uncertainty, and an actionable impression. For a
vision--language model, this creates two connected requirements: representing
volumetric anatomy and communicating the clinical significance of the observed
findings. Figure~\ref{fig:nv_reason_ct_overview}
illustrates this volumetric image-to-language objective.

Recent CT foundation models have demonstrated the value of learning from
volumetric images and associated clinical text. CT-CLIP and CT-CHAT
\cite{hamamci2026generalist}, Merlin \cite{blankemeier2026merlin}, and COLIPRI
\cite{wald2026colipri} provide important foundations for CT representation
learning, abnormality recognition, and generative interpretation. Within these
systems, the interface between the visual encoder and language model is an
important architectural choice. Pooling or resampling can reduce the spatial
granularity of the visual representation, while serializing features does not
automatically retain their explicit 3D coordinates in the language decoder.
Although visual features can encode geometry, preserving their depth, height,
and width indices provides an additional means of representing spatial
relationships during joint processing with text. This motivates an interface
that retains both volumetric features and their spatial indexing.

Clinical assistance also requires communicating how observations support an
interpretation. Radiologists organize their review around anatomical
structures, relate findings to diagnostic possibilities, acknowledge
uncertainty, and revisit the examination in response to follow-up questions.
These activities motivate supervision that includes the observations and
diagnostic considerations accompanying a report. We treat generated reasoning
as a clinical explanation that can be reviewed alongside the image and final
answer. Its value depends on factual grounding and usefulness to the reader;
a plausible explanation alone does not establish that it faithfully represents
the model's internal computation \cite{turpin2023unfaithful}.

Our previous work, NV-Reason-CXR-3B \cite{myronenko2025nvreasoncxr}, introduced
a reasoning-centered framework for chest radiography using radiologist-authored
demonstrations, report-grounded synthetic reasoning, and reinforcement learning
with verifiable abnormality rewards. The present work extends this methodology
to chest and abdominal CT. This extension requires connecting expert-guided
interpretation with a native volumetric architecture and accounting for
differences in anatomical coverage, report structure, and finding vocabularies
across regions.

Here we present NV-Reason-CT, a generative vision--language model that combines
native 3D visual encoding with radiologist-guided supervision and verifiable
reinforcement learning. The model couples the Qwen3.5-4B language model
\cite{qwen2026qwen35} with a native 3D vision transformer initialized from
COLIPRI pretrained weights \cite{wald2026colipri}. It processes regional chest
or abdominal volumes and passes all encoded visual tokens through a learned
projector to the language model without further spatial token merging. Explicit
depth, height, and width coordinates are incorporated through the decoder's
positional encoding, retaining volumetric spatial indexing during joint
processing of visual and text tokens.

We construct a curated training corpus of approximately 550,000 multimodal
instruction examples from 70,111 unique CT image inputs. A limited set of
recorded and transcribed expert-narrated CT interpretations provides direct
supervision and examples of systematic anatomical review, diagnostic
considerations, and uncertainty. These expert examples also guide the rewriting
of source reports into synthetic narrated interpretations, with the source
reports constraining the clinical content. The resulting corpus combines expert
and synthetic narrations with standardized reports, finding- and
anatomy-specific questions, multi-turn interactions, and refusal examples. Its
construction involves extensive rewriting and curation of structured reports,
synthetic narrations, and question--answer pairs, together with harmonization of
finding terminology and anatomical coverage. This process broadens the available
supervision while preserving the distinction between expert-authored examples
and report-derived synthetic content.

Training proceeds in two stages. End-to-end supervised fine-tuning (SFT) jointly
adapts the pretrained 3D encoder, learned projector, and language model to the
curated tasks. Subsequent Group Relative Policy Optimization (GRPO)
\cite{shao2024deepseekmath} uses region-specific abnormality and report-structure
rewards to optimize finding agreement while maintaining the intended reporting
format. This enables reinforcement learning from verifiable finding labels
without requiring an expert-authored reasoning trace for every training volume.

We evaluate abnormality classification and report generation across CT-RATE,
Merlin, RAD-ChestCT, and a held-out NIH cohort, complemented by qualitative
analysis of reasoning and follow-up interaction and a preliminary assessment by
expert radiologists. Together, these evaluations examine finding recognition,
the clinical content of generated reports, and the usefulness of explanations
for human review. We release NV-Reason-CT and its SFT and GRPO training code to
support reproducible research on explainable AI for volumetric medical imaging.

\section{Related work}
\label{sec:related_work}

Large-scale image--text learning and visual instruction tuning have made
open-ended image interpretation increasingly feasible. Vision Transformers
\cite{dosovitskiy2021vit} represent images as patch tokens; CLIP
\cite{radford2021clip} demonstrated that natural-language supervision can
produce transferable visual representations; and models such as Flamingo
\cite{alayrac2022flamingo} and LLaVA \cite{liu2023llava} showed how visual
features can condition a general-purpose language model. Medical generalist
models such as Med-PaLM M \cite{tu2024towards} have extended this paradigm to
clinical question answering and report generation. VILA-M3
\cite{nath2025vilam3} further showed how knowledge from domain-specific medical
expert models can be incorporated into a medical VLM. Most of this progress,
however, has been developed around two-dimensional inputs. Applying the same
interface to a CT by selecting a small number of slices, processing slices
independently, or aggressively resampling visual features creates a bottleneck
precisely where volumetric context is most important.

Early volumetric CT foundation models primarily pursued representation learning:
they sought an image encoder whose features transferred across classification
and other downstream tasks, rather than an end-to-end system for open-ended
clinical generation. Most followed the CLIP paradigm by aligning CT volumes with
text derived from radiology reports or diagnostic labels
\cite{hamamci2026generalist,blankemeier2026merlin,wald2026colipri,agrawal2025pillar0}.
A parallel line of
work learned image representations without paired text, commonly through
self-supervised objectives such as DINO \cite{morenoaguado2026voxelfm}. More
recent systems have increasingly
added dedicated report decoders or connected CT encoders to general-purpose
language models. We organize this progression into representation-learning
methods, CT report-generation systems, and generative CT vision--language
models. These categories overlap because a strong representation encoder can
later serve as the visual component of a generative system.

\subsection{Contrastive and representation-learning methods}

This class of methods is defined primarily by how the visual representation is
learned. Contrastive approaches jointly embed a CT volume and its associated
report, report sentence, or label description, encouraging matched image--text
pairs to be close and unmatched pairs to be separated. The resulting encoder
can support zero-shot classification by comparing an image with positive and
negative text prompts, or it can be transferred to a specific task by training
a linear or task-specific prediction head on top of its features. Although the
text encoder provides supervision during contrastive pretraining, it is not
always part of the deployed model. For linear-probe or task-specific transfer,
the text encoder is typically discarded and only the trained image encoder is
retained. It remains necessary for zero-shot image--text scoring and cross-modal
retrieval, where it encodes the comparison prompts or reports
\cite{hamamci2026generalist,agrawal2025pillar0,khlaut2026jolia}. Image-only
self-supervised methods pursue the same goal of a reusable CT representation but
replace paired text with objectives such as teacher--student consistency or
masked-image learning. Either type of encoder can subsequently be coupled to a
report decoder or an LLM, so the contrastive or self-supervised objective
describes the representation-learning stage rather than the full capabilities
of the final system.

\noindent\textbf{CT-CLIP.}
CT-CLIP introduced large-scale contrastive alignment between volumetric chest
CT and radiology reports using the CT-RATE training set
\cite{hamamci2026generalist}. This patient-disjoint training cohort contains
47,149 reconstructed chest CT volumes from 24,128 studies and 20,000 patients.
Its factorized vision transformer first models
in-plane patches and then aggregates through-plane information. The native
classification benchmark covers all 18 CT-RATE abnormalities through paired
positive and negative text prompts. CT-CLIP also provides the visual foundation
for CT-CHAT, which follows a LLaVA-style design: attention pooling compresses
the encoder output into 256 image tokens, and an MLP maps them into the
Llama~3.1 embedding space for instruction following and report generation. The
language model receives this 1D token sequence without an explicit
depth--height--width coordinate grid.

\noindent\textbf{Merlin.}
Merlin learns an abdominal CT representation by aligning a 3D
ConvNet with reports and structured clinical data
\cite{blankemeier2026merlin}. Representation learning uses 15,331 abdominal
CTs from the Stanford cohort, together with their reports and approximately
1.8 million structured clinical codes. Its native zero-shot classification task contains
30 report-derived findings, with unmentioned labels excluded and positive and
explicitly negative cases balanced separately for each finding. Although its
main contribution is representation learning, Merlin also attaches a language
model for section-wise abdominal report generation. A linear adapter projects
the encoder's last $7\times7\times10$ feature grid into RadLlama-7B input tokens;
the resulting flattened sequence does not retain an explicit 3D coordinate grid
inside the language model.

\noindent\textbf{COLIPRI.}
COLIPRI combines CT--report contrastive learning with image-only
self-supervision, opposite-sentence discrimination, and report-generation
objectives in a native 3D vision transformer
\cite{wald2026colipri}. It is developed entirely on chest CT: paired
vision--language training uses 22,676 CT-RATE volumes, while image-only
self-supervision additionally uses approximately 72,000 low-dose NLST scans.
Its CT-RATE classification experiments use the original
18-label ontology, while its RAD-ChestCT experiments report 16-label results
without fully enumerating the mapping. COLIPRI-CRM further connects the visual
encoder to a Qwen2.5 language model for report generation through a two-layer
MLP following the LLaVA framework. Qwen2.5 receives the projected patch
embeddings as a 1D token sequence rather than with their original
depth--height--width coordinates. Our model uses the released COLIPRI Primus
encoder weights as a strong initialization, rather than as a permanently frozen
feature extractor; the vision encoder, multimodal merger, and language model are
subsequently optimized together in our end-to-end training.

\noindent\textbf{Pillar-0.}
Pillar-0 introduced Atlas, a hierarchical volumetric encoder whose multiscale
attention makes processing full-resolution CT and MRI volumes computationally
efficient \cite{agrawal2025pillar0}. Its four modality-specific encoders are
pretrained on chest, abdomen--pelvis, and head CT examinations and breast MRI
examinations from one academic-center archive.
The encoders receive
multiple window-level views of each scan and are pretrained with an asymmetric
CLIP-style objective that aligns the Atlas volume representation with a frozen
Qwen3-Embedding-8B text encoder. Pillar-0 is a non-generative representation
model: downstream tasks use frozen embeddings with learned prediction heads.

\noindent\textbf{Jolia.}
Jolia adopts Atlas as its principal visual backbone and follows Pillar-0's use
of multi-window CT inputs and a frozen Qwen3-Embedding-8B text encoder, but it is
trained as a separate model~\cite{khlaut2026jolia}. Its headline configuration uses a smaller Atlas encoder
trained from scratch on 74,434 public CT--report pairs from CT-RATE and INSPECT
chest scans and Merlin abdominal scans. The
study also evaluates 3D ResNet-101 and ViT-B backbones. Jolia's main innovation,
ConQuer, augments global CLIP alignment with learnable anatomy-specific queries
and a separate contrastive loss for each concept, producing localized visual
representations without segmentation supervision. It supports 18-label CT-RATE
classification and, like Merlin, also adds a language-model decoder for report
generation. That pathway projects a single pooled 576-dimensional global image
feature into one Qwen3.5 input token; the localized ConQuer tokens and their 3D
relationships are not passed to the language model.

\noindent\textbf{VoxelFM.}
VoxelFM is an image-only 3D transformer pretrained with DINO-style
self-supervision and adapted with lightweight supervised heads
\cite{morenoaguado2026voxelfm}. Pretraining uses 137,107 CT studies spanning
the thorax, abdomen, and head and neck, drawn from CT-RATE, INSPECT, Merlin,
NLST, and other public collections. Its downstream classifiers cover all 18
CT-RATE findings and all 30 Merlin findings, and a separate Q-Former plus
Qwen3-8B pathway supports CT-RATE report generation. The Q-Former converts the
patch features into learned query outputs that the language model consumes as a
1D sequence, without an explicit dense 3D coordinate grid.

\noindent\textbf{CT-SSG.}
CT-SSG uses a supervised 2.5D convolutional encoder with graph-based slice
aggregation \cite{dipiazza2026ctssg}. It is trained and evaluated by fivefold
cross-validation on the 25,692-acquisition CT-RATE chest cohort, retaining one
reconstruction per acquisition. 

\noindent\textbf{fVLM.}
fVLM learns organ-conditioned contrastive representations using
segmentation-derived anatomical masks \cite{shui2025fvlm}. Its pretraining
resource, MedVL-CT69K, contains 272,124 multiregion CT scans from 69,086
patients, with 64,476 patients assigned to training; it includes chest,
abdomen, pelvis, and other body regions. 

\subsection{CT report-generation systems}

Dedicated CT reporting systems differ in how directly report text is generated
from the image and in how clinical content is measured.

\noindent\textbf{CT2Rep.}
CT2Rep generates CT-RATE findings with a volumetric encoder and
relational-memory transformer decoder \cite{hamamci2024ct2rep}. It is trained
on the CT-RATE chest training split, comprising 47,149 reconstructed volumes
from 20,000 patients.

\noindent\textbf{CT-CHAT.}
CT-CHAT connects the CT-CLIP encoder to Llama~3.1 through a learned multimodal
projector \cite{hamamci2026generalist,hamamci2025crg}. Its projector and
language-model adapters are trained on more than 2.7 million question--answer
pairs derived from the CT-RATE chest training cohort.

\noindent\textbf{BTB3D.}
BTB3D first compresses the volume into discrete causal 3D tokens before
language decoding \cite{hamamci2025btb3d}. Its report generator is likewise
trained on the 47,149 reconstructed chest volumes in the CT-RATE training split.

These systems are evaluated on CT-RATE through the 18 report-derived
abnormalities; CT-CHAT and BTB3D are also evaluated on the 16-label RAD-ChestCT
mapping after abnormalities are extracted from generated reports.

\Needspace{3\baselineskip}
\noindent\textbf{MS-VLM.}
MS-VLM combines a self-supervised 2D slice encoder with an inter-slice
transformer and a language-model bridge \cite{lee2024msvlm}. Its chest model
uses 47,149 CT-RATE training volumes, approximately 180,000 synthetic QA pairs,
and selected QA from the 1.3-million-pair RadGenome-ChestCT resource; a separate
rectal MRI experiment uses 186 training patients.

\noindent\textbf{CT-AGRG.}
CT-AGRG takes a more explicitly label-driven approach: an 18-label CT-RATE
classifier first selects positive abnormalities, after which a decoder generates
one sentence for each selected finding \cite{dipiazza_2025_ctagrg}. It is
trained on an earlier CT-RATE chest split containing 34,781 reconstructed
volumes from 17,799 patients.

These designs illustrate an important distinction between open-ended
image-conditioned reporting and reports whose content is gated by a separately
supervised abnormality classifier.

\noindent\textbf{DCP-PD.}
DCP-PD uses the Pillar-0 Atlas encoder and can condition its report generator on
CT-RATE-trained discriminative predictions \cite{wang2026dcppd}. It uses the
official 24,128-study CT-RATE chest training partition after removing
approximately 800 head-only studies.

\noindent\textbf{EXACT-CHAT.}
EXACT-CHAT derives 18 voxel-level anomaly maps and passes their diagnostic
predictions to its report decoder; its RAD-ChestCT evaluation uses a modified
16-label mapping \cite{bai2026exact}. Its weakly supervised vision pretraining
and report-generation training use 24,128 CT-RATE chest volumes.

\noindent\textbf{MonteRET.}
MonteRET retrieves condition- and region-matched cases to support report
generation and refinement \cite{lin2026monteret}. It is trained on 24,128
RadGenome-ChestCT scans---the CT-RATE chest training studies paired with
anatomically organized reports---and uses the same training reports as its
retrieval database.

\noindent\textbf{Astra.}
Astra combines a Merlin encoder with supervised and reinforcement-learning
stages \cite{wang2026astra}. It is developed on CTRgDB, a harmonized corpus of
90,678 thoracoabdominal CT--report pairs assembled from CT-RATE, INSPECT,
Merlin, AbdomenAtlas~3.0, and BIMCV.

\noindent\textbf{Resolution Meets Reduction.}
Resolution Meets Reduction couples a high-resolution slice encoder to a
language model through a Perceiver resampler \cite{suprijadi2026resolution}.
Its controlled report-generation experiments train separately on 22,590
CT-RATE chest studies and 19,361 Merlin abdominal cases.

The CT-RATE clinical-content results reported for these systems are generally
derived from the original 18-label ontology, although evaluation units, label
extractors, and averaging conventions differ.

As noted above, Merlin and Jolia straddle the representation and reporting
categories. We retain them in the first group because their principal novelty
is the learned CT representation, while explicitly recognizing that both also
attach language-model decoders through the projected, non-grid-aware interfaces
described above and provide report-generation results.

\subsection{Generative CT vision--language models}

General-purpose medical vision--language models aim to support classification,
question answering, and free-form generation with a common interface.

\noindent\textbf{RadFM.}
RadFM uses a 3D vision transformer and Perceiver-style resampler with an
autoregressive medical language model \cite{wu2023radfm}. It is pretrained on
MedMD, a broad multimodal mixture containing 15.5 million 2D images and 500,000
3D scans across CT, MRI, PET, radiography, and anatomical regions from brain to
pelvis, then fine-tuned on a three-million-pair radiology subset.

\noindent\textbf{M3D-LaMed.}
M3D-LaMed combines a native 3D transformer with staged image--text alignment and
multimodal instruction tuning \cite{bai2024m3dlamed}. Training uses M3D-Data,
which contains 120,092 CT image--text pairs covering anatomy throughout the body
and collected from Radiopaedia, plus 662,000 instruction--response pairs spanning report
generation, VQA, localization, and segmentation.

Both models can be prompted independently for all 18 CT-RATE abnormalities;
later studies also evaluate their generated reports against 16-label
RAD-ChestCT targets.

\noindent\textbf{MedGemma~1.5.}
MedGemma~1.5 represents a CT as a sequence of sampled 2D slices and uses a
general medical multimodal decoder \cite{sellergren2026medgemma15}. Its added
volumetric training data include 282,963 internal CT examples covering head,
chest, and abdomen, alongside 167,674 multiregion MRI examples and the broader
MedGemma medical mixture.

\noindent\textbf{ClinFusion.}
ClinFusion combines several 2D encoders with a native 3D pathway before fusing
the representations into Qwen3-VL \cite{yuan2026clinfusion}. Its progressive
training corpus contains approximately 22.2 million mixed samples across 12
medical imaging modalities; the volumetric stages draw on CT-RATE, Merlin,
RadGenome-ChestCT, INSPECT, M3D-Cap, AMOS, and synthetic chest and abdominal CT
collections.

We evaluate both model families on all 18 CT-RATE findings using
per-abnormality binary questions. Their broader generative interfaces make them
useful comparisons for instruction-following behavior, but their visual
interfaces and spatial assumptions differ from native dense 3D token
processing.

Chain-of-thought prompting \cite{wei2022chainofthought} and reasoning-oriented
reinforcement learning, exemplified by DeepSeek-R1 \cite{deepseek2025}, have
shown that language models can produce structured, multi-step solutions. In a
medical setting, however, a generated rationale should not be assumed to reveal
the model's faithful internal computation \cite{turpin2023unfaithful}.

A common design across language-connected CT systems is to project or resample
visual features into a 1D sequence that the language model then
consumes alongside text
\cite{hamamci2026generalist,wu2023radfm,bai2024m3dlamed,morenoaguado2026voxelfm,suprijadi2026resolution}.
Some of these models encode 3D geometry
inside the vision tower; this should not be confused with carrying the original
3D token coordinates through the language model itself. After the connector,
the language decoder commonly receives flattened visual embeddings without an
explicit dense depth--height--width grid in its own positional mechanism.

\noindent\textbf{NV-Reason-CT.}
Our model is designed to preserve this structure at both stages. Its 3D vision
transformer uses three-axis rotary positional encoding and produces a
$24\times24\times24$ grid of 13,824 visual tokens. We initialize the Primus
visual encoder \cite{wald2026primus} from COLIPRI pretrained weights
\cite{wald2026colipri} and connect it to Qwen3.5-4B \cite{qwen2026qwen35}
through a learned projector that retains every visual token without further
spatial merging. The original grid dimensions accompany these tokens, allowing
the language model's multimodal rotary positional encoding (MRoPE)
\cite{bai2025qwen25vl} to assign each token explicit depth, height, and width
indices. Volumetric position therefore remains represented throughout
language-decoder self-attention during joint processing with text, in addition
to its representation within the visual encoder. This design is intended to
support spatially specific interpretation, such as distinguishing left- from
right-sided renal lesions, describing craniocaudal extent, or localizing
findings to different thoracic regions. End-to-end SFT jointly adapts the
encoder, projector, and language model using curated chest and abdominal CT
tasks, including radiologist-guided reasoning. Subsequent GRPO uses
region-specific verifiable rewards to optimize abnormality agreement and
report structure.

\section{Methods}
\label{methods}

\subsection{Overview}
NV-Reason-CT combines a native 3D vision--language architecture with
radiologist-guided supervision and verifiable reinforcement learning for chest
and abdominal CT interpretation. The model represents a regional CT volume as
spatially indexed visual tokens and processes them alongside a natural-language
instruction to generate structured reports, finding-specific answers, and
clinical explanations.

Training adapts the reasoning-centered strategy of
NV-Reason-CXR~\cite{myronenko2025nvreasoncxr} to volumetric CT. A limited set of
recorded and transcribed expert interpretations provides direct supervision and
guides the construction of synthetic narrated interpretations from source
reports. These examples form part of a curated corpus that also includes
standardized reports, abnormality questions, localization and attribute
questions, follow-up interactions, and refusal examples. End-to-end supervised
fine-tuning (SFT) jointly adapts the visual encoder, projector, and language
model to these complementary tasks. Subsequent Group Relative Policy
Optimization (GRPO)~\cite{shao2024deepseekmath} uses region-specific abnormality
labels and output-structure requirements to optimize finding agreement and
report organization without requiring an expert-authored reasoning trace for
every training volume.

Generated reasoning is presented as a reviewable clinical explanation that
organizes observations, relevant normal findings, uncertainty, and conclusions.
Its usefulness must be assessed through factual grounding, report quality,
downstream task performance, and expert evaluation; it is not assumed to reveal
the model's internal computation.

\subsection{Model architecture and volumetric interface}
\label{sec:methods_architecture}

\begin{figure}[!htbp]
\centering
\includegraphics[width=\textwidth]{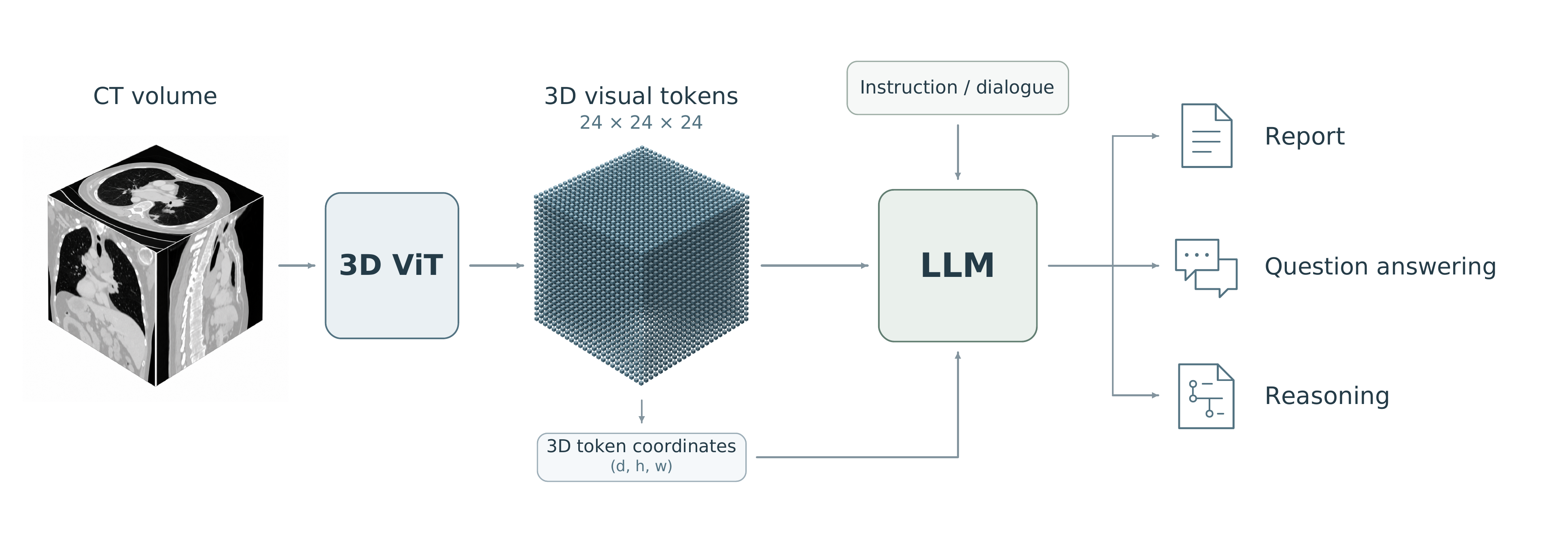}
\caption{Architecture overview of NV-Reason-CT. A native 3D visual encoder,
shown together with its learned projection as the 3D ViT block, maps a CT
volume to a $24\times24\times24$ grid of visual tokens. The LLM receives the
projected tokens and instruction or dialogue context. Each visual token is
assigned depth, height, and width coordinates $(d,h,w)$ derived from the token
grid and incorporated through multimodal rotary positional encoding (MRoPE).
Outputs include reports, question answering, and reasoning explanations. The
CT cube is an illustrative rendering based on orthogonal views extracted from
a CT scan.}
\label{fig:nv_reason_ct_architecture}
\end{figure}

NV-Reason-CT couples a Primus 3D vision transformer~\cite{wald2026primus},
initialized from released COLIPRI pretrained weights~\cite{wald2026colipri},
with the Qwen3.5-4B language model~\cite{qwen2026qwen35} through a learned
projector. Figure~\ref{fig:nv_reason_ct_architecture} summarizes the architecture
and its volumetric interface.

For a preprocessed single-channel volume
$x\in\mathbb{R}^{1\times192\times192\times192}$, the visual encoder uses
non-overlapping $8\times8\times8$-voxel patches and three-axis rotary positional
encoding to produce a $24\times24\times24$ grid of 864-dimensional features.
The projector maps each feature to the language model's 2,560-dimensional
embedding space without spatial merging, preserving all 13,824 visual tokens.

The projected tokens are serialized for language-model input, accompanied by
their original grid dimensions. The language model's multimodal rotary
positional encoding (MRoPE)~\cite{bai2025qwen25vl} uses these dimensions to
assign each visual token explicit depth, height, and width indices. Volumetric
position is therefore represented both within the visual encoder and during
language-decoder self-attention as visual and text tokens are processed
together. This interface retains the spatial indexing of the encoded volume
through language decoding.

\subsection{Volumetric preprocessing}
\label{sec:methods_preprocessing}

Each NIfTI volume is loaded as a single-channel tensor and processed using
MONAI transforms~\cite{cardoso2022monai}. Volumes are reoriented to
left--posterior--superior (LPS) orientation and resampled to 2-mm isotropic
spacing using trilinear interpolation. After resampling, dimensions smaller
than 192 voxels are padded with $-1{,}000$ Hounsfield units (HU) before the
anatomy-aware region selection described in the following subsection. The
resulting $192\times192\times192$-voxel input corresponds to a
$384\times384\times384$-mm field of view. Intensities are clipped to
$[-1000,1000]$ HU and divided by 1,000, yielding values in $[-1,1]$.

\subsection{Anatomy-aware cropping: chest and abdomen}
\label{sec:methods_anatomy_crop}

An anatomical region of interest could be obtained using an organ-segmentation
model or a learned 3D bounding-box detector. Both approaches introduce an
additional model into preprocessing, with associated integration,
computational, and validation requirements. Here, we use an automatic,
deterministic heuristic to select regional inputs for the chest and abdomen,
the two anatomical regions represented during training. Each input is assigned
a chest or abdominal region designation that determines the cropping procedure.

After resampling, the method thresholds the CT at $-500$ HU to form a body
mask. Slice-wise morphological hole filling identifies low-attenuation air
enclosed by the body, and 3D connected-component analysis groups these regions.
Component selection uses size and superior location to identify candidate
lung regions, which are accepted only when their combined physical volume is
at least 250~mL.

The detected lung bounds provide a common anatomical landmark for both
supported regions. For chest inputs, the $192\times192\times192$-voxel crop
is centered craniocaudally on the detected lungs. For abdominal inputs, a
target interval is defined using a 300-mm length extending primarily
inferiorly from the lung base, with up to 100~mm of overlap above it, subject
to the available anatomical coverage. This interval determines the center of
the fixed $192\times192\times192$-voxel crop. The in-plane crop is centered
in both cases.

\begin{figure}[!htbp]
\centering
\includegraphics[width=0.8\textwidth]{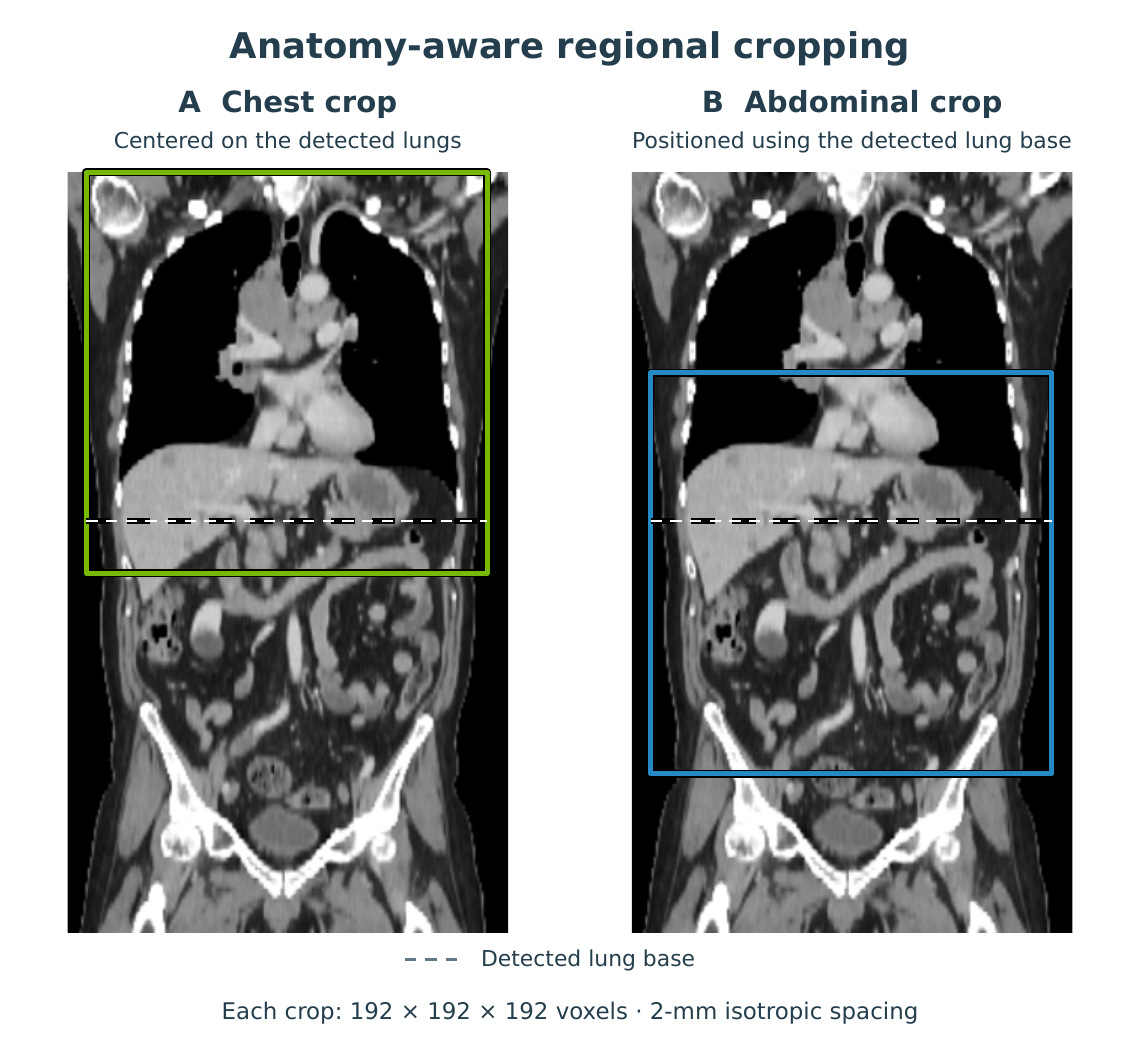}
\caption{Anatomy-aware regional cropping. The same coronal CT slice is shown
at the same scale and display window with the chest crop outlined in green
(A) and the abdominal crop in blue (B). Dashed lines mark the inferior bound
of the detected 3D lung region, which provides a landmark for crop placement.
Each square outlines a cross-section through a $192\times192\times192$-voxel
cubic input at 2-mm isotropic spacing ($384$~mm per side). The model processes
one regional crop at a time.}
\label{fig:nv_reason_ct_anatomy_cropping}
\end{figure}

When lung bounds cannot be recovered, the preprocessor uses a deterministic
edge-aligned fallback crop. The heuristic depends on sufficient lung visibility
and may mislocalize the intended region when lung landmarks are absent or
difficult to identify. Figure~\ref{fig:nv_reason_ct_anatomy_cropping} illustrates
chest and abdominal crop placement in the same CT volume. Additional views of
these crops and their corresponding model outputs are shown in
Appendix~\ref{app:model_output_examples}.

\subsection{Problem formulation and output representation}
\label{sec:methods_problem}

Let $x$ denote a regional CT volume, $q$ the instruction and any preceding
dialogue context, and $y$ the model response. A single conditional policy
$\pi_\theta(y\mid x,q)$ supports structured chest and abdominal reports,
free-form or constrained question answering, and multi-turn follow-up
interaction. For structured-report reinforcement learning, each example also
includes an anatomy region
$a\in\{\text{chest},\text{abdomen}\}$ and a reference finding set
$z\subseteq\mathcal{A}_a$, where $\mathcal{A}_a$ is the corresponding
30-finding chest or 29-finding abdominal ontology described in
Section~\ref{sec:data_curation}.

Reasoning-enabled SFT responses place the clinical analysis within a
\verb|<think>...</think>| span. In the structured-report format used for GRPO,
the report is followed by exactly one machine-readable
\verb|<answer>...</answer>| block containing a comma-separated finding set.
This block provides the deterministic finding representation used to compute
the abnormality-agreement reward. Other SFT tasks retain their task-specific
response formats, including free-text and Yes/No answers.

\subsection{Supervised fine-tuning}
\label{sec:methods_sft}

\begin{figure}[!htbp]
\centering
\includegraphics[width=\textwidth]{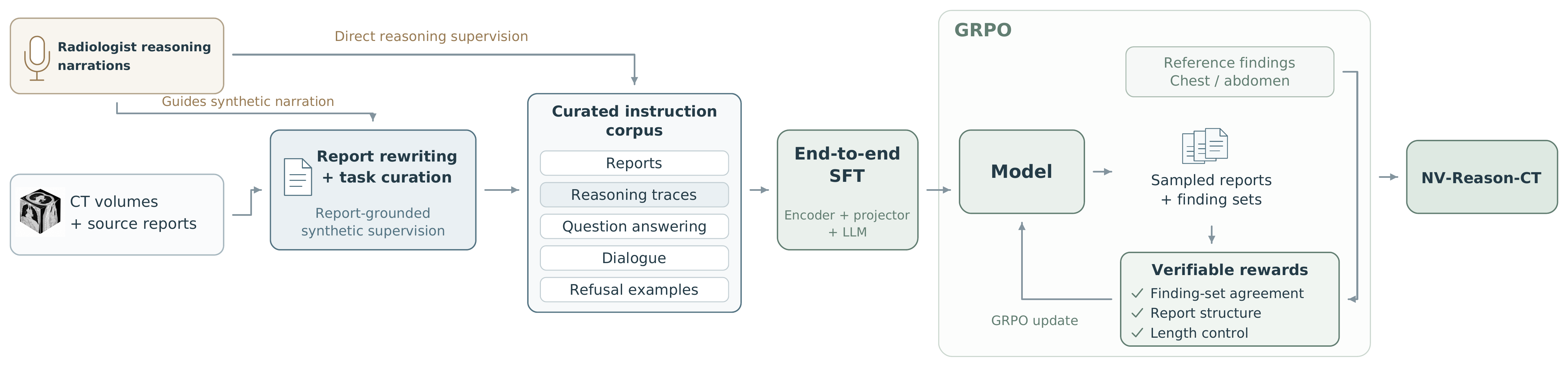}
\caption{Training workflow for NV-Reason-CT. Radiologist reasoning narrations
provide direct supervision and guide source-report rewriting into synthetic
narrations. Together with curated reports, question answering, dialogue, and
refusal examples, they form the instruction corpus for end-to-end SFT of the
visual encoder, projector, and language model. Starting from the SFT model,
GRPO samples reports and finding sets for CT-conditioned prompts and scores
each completion using region-specific reference findings, report structure,
and completion length. The resulting rewards guide policy updates.}
\label{fig:nv_reason_ct_training}
\end{figure}

We jointly fine-tune the 3D visual encoder, projector, and language model on
the curated instruction corpus. Figure~\ref{fig:nv_reason_ct_training} summarizes
the training workflow from radiologist-guided data curation through SFT and
GRPO.

\begin{samepage}
Given an SFT example $(x,q,y^\star)$, where
$y^\star$ is the target assistant response, we minimize the autoregressive
cross-entropy:
\[
\mathcal{L}_{\mathrm{SFT}}(\theta)
=-\mathbb{E}_{(x,q,y^\star)}\sum_{t=1}^{T}
\log \pi_\theta\!\left(y_t^\star\mid x,q,y_{<t}^\star\right).
\]
\end{samepage}
Here, $T$ is the target-response length. System and user tokens are excluded
from the loss. In multi-turn examples, preceding dialogue provides conditioning
context, and supervision is applied only to the final assistant response.

The training mixture includes region-specific and generic structured-report
prompts, expert and report-derived synthetic reasoning examples, concise
report-planning rationales, abnormality lists, binary and section-specific
questions, location, laterality, and severity questions, single- and multi-turn
follow-ups, and refusal examples. Data sources, regional composition, and
task-specific curation are described in
Section~\ref{sec:data}.

\subsection{Group Relative Policy Optimization}
\label{sec:methods_grpo}

Starting from the SFT policy, we apply GRPO~\cite{shao2024deepseekmath} to
optimize structured report generation using region-specific abnormality and
output-structure rewards. GRPO uses the same training split paired with
reference abnormality sets. The model is prompted to generate a complete
report followed by a finding-set answer block, and each completion is scored
using the reward function defined in Section~\ref{sec:methods_rewards}.

For each prompt, we sample $G$ completions
$y_{1:G}\sim\pi_{\theta_{\mathrm{old}}}(\cdot\mid x,q)$ and
assign each completion a scalar reward $r_i=R(y_i,z,a)$. Here,
$\pi_{\theta_{\mathrm{old}}}$ denotes the policy used to generate the
completions. The group-relative advantage is
\[
A_i=\frac{r_i-\frac{1}{G}\sum_{j=1}^{G}r_j}{\sigma_r+\epsilon}.
\]
The numerator subtracts the mean reward within the prompt group, while
$\sigma_r$ is the standard deviation of rewards across the distributed batch.
The small positive constant $\epsilon$ stabilizes the division.

The policy is updated with the clipped GRPO objective
\[
\mathcal{L}_{\mathrm{GRPO}}(\theta)=
-\mathbb{E}_{i,t}\left[
\min\!\left(\rho_{i,t}A_i,
\operatorname{clip}(\rho_{i,t},1-\epsilon_{\mathrm{low}},
1+\epsilon_{\mathrm{high}})A_i\right)
\right],
\]
where
$\rho_{i,t}=\pi_\theta(y_{i,t}\mid x,q,y_{i,<t})/
\pi_{\theta_{\mathrm{old}}}(y_{i,t}\mid x,q,y_{i,<t})$
is the token-probability ratio between the policy being optimized and the
sampling policy. The indices $i$ and $t$ denote completions and token
positions, respectively.
We use $\epsilon_{\mathrm{low}}=0.20$ and
$\epsilon_{\mathrm{high}}=0.28$. The reported run uses no explicit reference
model Kullback--Leibler (KL) penalty ($\beta=0$).

The resulting GRPO checkpoint is used in the reported evaluations.

\subsection{Region-aware verifiable rewards}
\label{sec:methods_rewards}

The anatomy region $a$ determines the finding ontology and report schema used
to score each completion. The reward combines abnormality-set correctness,
structured-report conformity, and a soft completion-length penalty:
\[
R(y,z,a)=2.0\,r_{\mathrm{set}}+0.5\,r_{\mathrm{struct}}+r_{\mathrm{len}}.
\]

\paragraph{Abnormality-set correctness.}
Let $\hat z=\Phi(y)$ be the comma-separated finding set parsed from the first
\verb|<answer>...</answer>| block. Agreement with the reference finding set
$z$ is measured using set F1:
\[
r_{\mathrm{set}}(y,z)=
\frac{2\lvert \hat z\cap z\rvert}{\lvert\hat z\rvert+\lvert z\rvert}.
\]
The correctness term is set to zero unless the completion contains exactly
one opening and one closing answer tag forming a valid answer block. A
present but empty block is mapped to the region-specific no-finding label,
whereas a missing block is invalid. A prediction combining the no-finding
label with one or more positive findings is treated as contradictory and also
receives zero correctness reward.

\paragraph{Structured-report reward.}
The format reward $r_{\mathrm{struct}}\in[0,1]$ evaluates conformity to the
report schema for the selected anatomy region. It combines common
report-block presence and ordering (0.15), region-specific organ-heading
presence and ordering (0.55), non-empty section and impression text (0.20),
and a numbered or bulleted impression (0.10). The Technique section is
optional. This term evaluates report organization and does not itself assess
the clinical correctness of the prose.

\paragraph{Length-band penalty.}
A bounded length penalty discourages completions containing only the finding
list and excessively long reports. The penalty is zero within the target
length band and decreases linearly outside it, with its magnitude capped at
one:
\[
r_{\mathrm{len}}(y)=
\begin{cases}
\max\!\left(-1,\frac{L(y)-180}{180}\right), & L(y)<180,\\
0, & 180\le L(y)\le900,\\
\max\!\left(-1,\frac{900-L(y)}{300}\right), & L(y)>900,
\end{cases}
\]
where $L(y)$ is the completion length in tokens.

\subsection{Training and implementation details}
\label{sec:methods_training}

Training uses 16 nodes, each equipped with eight NVIDIA H100 GPUs. We use fused
AdamW with a per-device batch size of one and two gradient-accumulation steps.
The training stack is implemented in PyTorch using Hugging Face Transformers,
TRL, Accelerate, and MONAI. Our training code is publicly
available\footnote{https://github.com/NVIDIA-Medtech/NV-Reason-CT}.

SFT uses a base learning rate of $2\times10^{-5}$, a cosine schedule with 3\%
warm-up, and gradient-norm clipping at 0.3. Component-specific learning rates
are $0.1\times$ the base rate for the 3D visual encoder, $5\times$ for the
projector, and $1\times$ for the language model.

For GRPO, we sample 16 completions per prompt at temperature 1.0, with a maximum
completion length of 1,300 tokens. We use batch-level reward scaling, a learning
rate of $1\times10^{-6}$, five warm-up steps, a cosine schedule decaying to 5\%
of the initial learning rate, and gradient-norm clipping at 1.0.

\section{Data}
\label{sec:data}

\subsection{Datasets}
\label{sec:data_datasets}

Our training data comprise three sources---CT-RATE, CancerVerse, and an internal
NIH dataset---each containing 3D CT volumes paired with radiology reports (see
Table~\ref{tab:dataset_overview}). Where the field of view permits, a volume
contributes separate chest and abdominal crops for region-specific tasks.
Crop overlap is generally greater in CT-RATE than in the more extensive NIH
scans. For evaluation, we use CT-RATE for chest, separate NIH internal
chest and abdomen cohorts, Merlin as an external abdominal dataset, and
RAD-ChestCT as an external chest dataset.

\begin{table}[H]
\centering
\scriptsize
\caption{NV-REASON-CT data cohorts. Chest and abdomen columns report unique
regional crops used for each task and are not mutually exclusive.}
\label{tab:dataset_overview}
\renewcommand{\arraystretch}{1.12}
\setlength{\tabcolsep}{4pt}
\begin{tabularx}{\textwidth}{@{}l>{\raggedright\arraybackslash}Xrr>{\raggedright\arraybackslash}Xrr@{}}
\toprule
\textbf{Dataset} & \multicolumn{3}{c}{\textbf{Training}} &
\multicolumn{3}{c}{\textbf{Evaluation}} \\
\cmidrule(lr){2-4}\cmidrule(l){5-7}
& \textbf{Overall} & \textbf{Chest} & \textbf{Abdomen} &
\textbf{Overall} & \textbf{Chest} & \textbf{Abdomen} \\
\midrule
CT-RATE & 47,149 cases; 24,128 studies; 20,000 patients & 46,203 & 20,243 &
3,039 public validation cases; 3,002 after correction; 1,564 studies; 1,304 patients & 3,002 & -- \\
NIH internal & 15,991 cases; 15,991 patients & 12,742 & 12,243 &
5,092 cases; 5,092 patients & 3,981 & 4,205 \\
CancerVerse & 22,720 cases; 13,778 patients & 4,668 & 4,438 & -- & -- & -- \\

\bottomrule
\end{tabularx}
\end{table}

\noindent\textbf{CT-RATE.}
CT-RATE is a public dataset for chest CT representation learning, report
generation, and multi-label abnormality classification
~\cite{hamamci2026generalist,ctratedataset2026}. The release contains 25,692
non-contrast CT studies from 21,304 patients, paired with radiology reports and
report-derived labels for 18 abnormalities. Because an acquisition can include
multiple reconstructed series, the 25,692 studies correspond to 50,188
three-dimensional cases. The patient-disjoint split contains 47,149 training
cases from 20,000 patients and 3,039 public validation cases from 1,304
patients. We use the corrected validation manifest, which removes 37 non-chest
volumes and retains 3,002 cases while preserving all 1,564 validation studies.
For training, we generated 46,203 chest crops. We additionally selected 20,243
cases with a substantial upper-abdominal field of view and included the upper
abdomen as a separate regional crop.

\noindent\textbf{NIH internal collection.}
The NIH internal data contains 15,991  training cases, each from a unique
patient. From this cohort, we generated 12,742 chest crops and 12,243 abdominal
crops. Most CTs included almost the whole body field of view,  allowing for proper chest and abdominal crops with only a small overlap. A held out set of 5,092 cases was used for validation (each from a unique patient). Of these, 3,981 contain the full chest and 4,205 contain the full abdomen. Reports in both deliveries were mapped to 30 chest and 29 abdominal findings.

\noindent\textbf{CancerVerse.}
CancerVerse is a public dataset centered on 13
abdominal and pelvic malignancy types and pairs CT volumes with radiology
reports and clinical metadata~\cite{li2026cancerverse}. We use 22,720 CT--report pairs to generate 4,668 chest crops and 4,438 abdominal crops, each targeting the full
corresponding anatomical region. The available reports contain cancer-focused
sections rather than complete radiology reports. Because most anatomical
structures are not mentioned, these reports are not suitable as targets for
structured report generation. We therefore use CancerVerse only for targeted
VQA on the available cancer-related abnormalities.

\noindent\textbf{Merlin (external).}
The Merlin Abdominal CT Dataset was used only for evaluation; it contains 25,494 abdominal or abdominopelvic
CT--report pairs from 18,317 patients treated at Stanford Hospital between
2012 and 2018~\cite{blankemeier2026merlin}. The released metadata contains
15,314 training, 5,055 validation, and 5,125 test cases. We use all 5,125 test
cases for report generation; 5,082 have the released 30-finding classification
labels. 

\noindent\textbf{RAD-ChestCT (external).}
RAD-ChestCT was used only for evaluation
~\cite{draelos2021radchestct,radchestct2022release}. The public release contains
3,630 volumes, each accompanied by an $84\times52$ abnormality-by-location
label matrix extracted from its source radiology report. We evaluate a
harmonized 16-label subset: 15 findings map directly, while the generic
calcification target combines the CT-RATE arterial-wall and coronary-artery
wall calcification outputs. The public release does not provide paired report
text, so report-generation outputs are assessed through label extraction
rather than reference-text similarity.

\subsection{Data Curation}
\label{sec:data_curation}

Our supervised fine-tuning corpus combines CT-RATE, NIH, and CancerVerse and is
organized into the complementary curation families below.
The resulting curriculum contains approximately 550,000 structured multimodal
instruction examples across chest and abdominal regions, constructed from
70,111 unique image inputs; because one CT can support multiple tasks, the
number of examples is not a count of independent examinations.

\begin{figure}[!p]
\centering
\includegraphics[width=\textwidth,height=0.85\textheight,keepaspectratio]{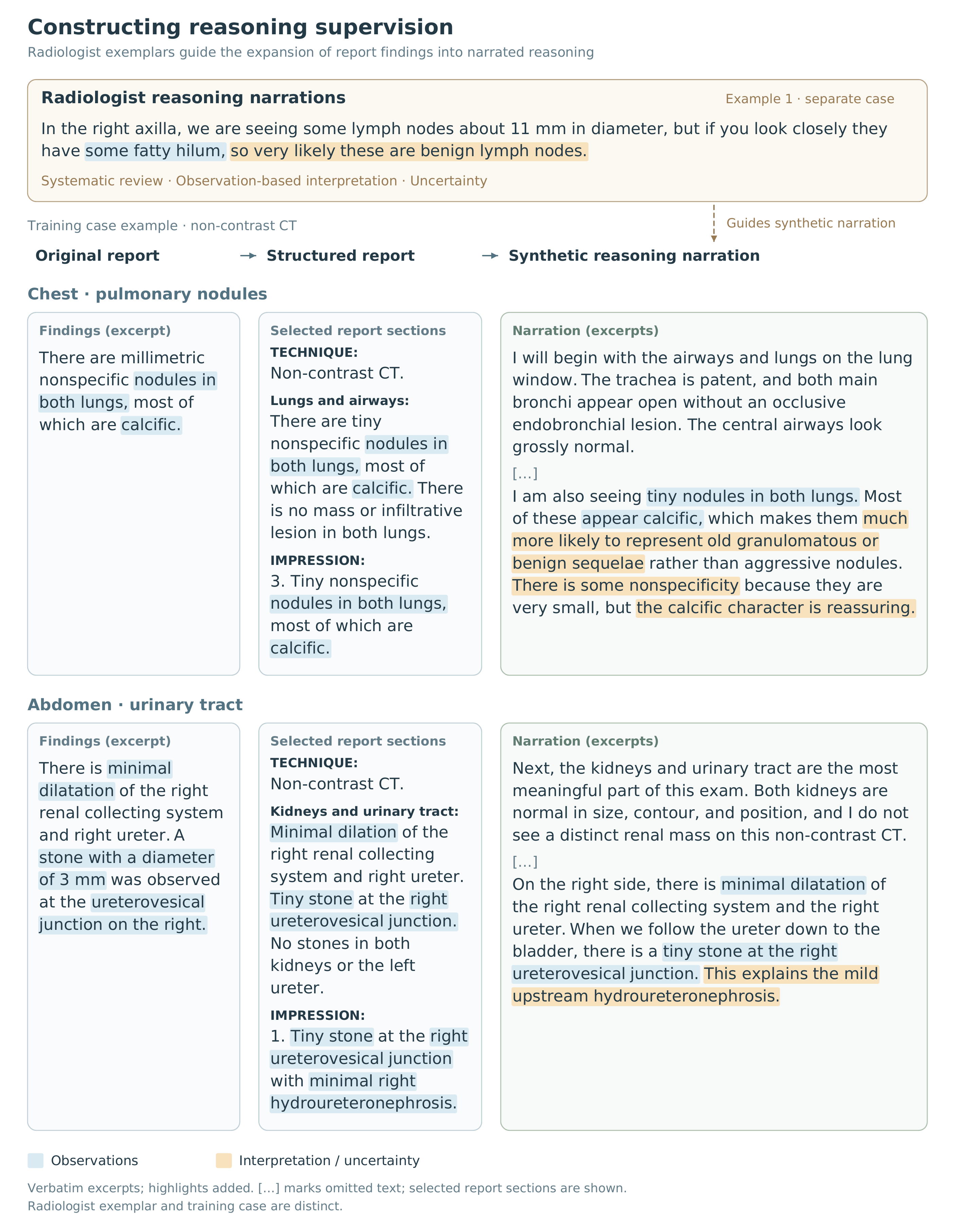}
\caption{Radiologist-guided construction of reasoning supervision. A
radiologist narration from a separate case illustrates the exemplars used to
guide synthetic rewriting. Chest and abdominal excerpts from one training
case show original report findings, selected structured-report sections, and
synthetic reasoning narrations. Blue highlights indicate observations; warm
highlights indicate interpretation or uncertainty. Excerpts retain their source
wording, with bracketed ellipses marking omitted passages and original numbering
retained for selected impression items.}
\label{fig:nv_reason_ct_reasoning_curation}
\end{figure}

\Needspace{3\baselineskip}
\noindent\textbf{Expert CT interpretations and synthetic expansion.}
Working with radiologists, we collected a limited set of recorded CT
interpretations. Radiologists were instructed to follow their usual image-review
workflow while narrating their observations and reasoning, including provisional
hypotheses, differential diagnoses, and uncertainty. They were encouraged to
articulate tentative possibilities and unresolved questions, including
considerations that might not be retained in a final radiology report, and to
conclude with a summary of their observations. The recordings were transcribed
into text, then curated and standardized into expert annotations. These
annotations provided direct SFT supervision and served as exemplars for
rewriting source reports into synthetic narrated interpretations that followed
the organization and narrative style of the expert examples.
Figure~\ref{fig:nv_reason_ct_reasoning_curation} illustrates this process using
chest and abdominal excerpts from one training case and a separate radiologist
narration exemplar.

\noindent\textbf{Ontology and coverage harmonization.}
The source reports differ in terminology, reporting style, anatomical
coverage, and available annotations. We therefore developed a
radiologist-guided ontology defining a common vocabulary and labeling rules for
the 30 chest and 29 abdominal findings evaluated in this work. For each
abnormality, the ontology specifies accepted
synonyms and aliases, inclusion and exclusion criteria, negation, uncertainty,
and temporal-status rules, anatomical and severity qualifiers, overlap and
parent--child relationships, and region-specific reporting conventions. A
language-model-assisted labeling pass mapped the heterogeneous reports to this
fixed inventory, with deterministic checks of annotation structure and
consistency between parent and child labels.
Chest, abdomen, and pelvis coverage was assessed separately as full, limited,
not imaged, or unknown. Negative labels were used only when the relevant region
was fully covered; for limited or unknown coverage, an unmentioned finding was
masked rather than treated as absent. An explicitly reported positive could
still support a targeted task when coverage was incomplete. The derived
\emph{No Chest Finding} and \emph{No Abdominal Finding} labels were assigned
only after full-coverage checks. Collaborating radiologists reviewed the
ontology definitions, standardized schemas, and representative outputs. The
complete finding inventory is listed with the per-finding evaluation in
Table~\ref{tab:internal_dataset_classification}.

\noindent\textbf{Structured-report generation.}
Eligible reports were rewritten into fixed chest- or
abdomen-specific schemas. The chest schema organizes findings by medical
devices, lungs and airways, pleura, mediastinum and hila, heart and pericardium,
great vessels, esophagus, chest wall and breast, and bones; the abdominal
schema analogously covers medical devices, hepatobiliary organs, spleen,
pancreas, adrenal glands, kidneys and urinary tract, bowel, peritoneum, lymph
nodes, vessels, abdominal wall and soft tissues, and bones. Both terminate in
a concise impression. Rewriting retained current, image-supported findings,
uncertainty, and clinically appropriate differentials while excluding history,
comparison-only statements, recommendations, and unsupported content. Outputs
were checked for the required headings and order, consistency of finding
polarity, and agreement with the extracted ontology labels. The family
contains complete explicit-region prompts, region-specific prompt variants,
and generic structured-report prompts; prompt augmentation changes the
instruction while preserving the same target report.

\noindent\textbf{Reasoning walkthroughs.}
For the same eligible chest and abdominal studies, the
reports were converted into continuous, first-person radiologist
search-pattern walkthroughs. These examples move systematically through the
relevant anatomy, connect important positive and negative observations, and
retain only supported uncertainty or differential reasoning. They exclude
patient-history leakage, prior-comparison claims, recommendations, dataset
references, and unsupported precision. Each walkthrough ends with the
canonical positive-label set, providing both free-form reasoning supervision
and an auditable structured answer. Additional prompt variants change the
wording of the request without altering the target reasoning.

\noindent\textbf{Single-turn follow-up QA.}
We generated clinically natural follow-up questions covering finding presence or absence,
location, differential diagnosis, management implications, clinical context,
unsupported requests, terminology, and deterministically derived scan coverage
or contrast questions. Each answer was constrained to the image-grounded
report context and paired with concise reasoning. For the single-turn family,
we retained only questions that remain unambiguous when separated from their
source conversation and rejected dangling references that require an earlier
turn.

\noindent\textbf{Multi-turn follow-up dialogue.}
The complementary multi-turn family preserves the initial structured-report
request and four subsequent follow-up exchanges, allowing later questions to
refer to findings established earlier in the conversation. Responses remain concise and report-grounded, and
unsupported questions are answered by stating the evidentiary limitation
rather than inventing a finding.

\noindent\textbf{Binary abnormality QA.}
For each chest or abdominal ontology finding, we formed explicit positive and
negative question pools from CT-RATE, NIH, and CancerVerse, applied per-label
caps, and deterministically sampled from each source pool.
Targets are direct yes/no answers with brief supporting reasoning.

\noindent\textbf{Section-specific reporting.}
Structured reports were decomposed into their canonical anatomical sections,
and ontology findings were mapped to the section in which they should appear.

\noindent\textbf{Anatomical localization QA.}
This family comprises left/right/bilateral questions, fine anatomical-location questions,
and contrastive location yes/no questions. The finding and
its location had to co-occur affirmatively within the same canonical report
section. Negated, resolved, ambiguous, or cross-section associations were
discarded. Sampling balanced laterality within findings, capped bilateral
dominance, limited each study to three questions.

\noindent\textbf{Severity and attribute QA.}
Qualitative size, burden, multiplicity, severity, distribution, fluid amount,
and extent were extracted from 
reports. As in localization curation, the abnormality and its
descriptor had to be unambiguously bound within the same report section; a
positive label alone did not license an attribute. 

\noindent\textbf{Refusal and input-validity supervision.}
Two small safety-oriented training sets were added. The first pairs valid CT
volumes with unrelated or nonmedical requests and teaches the model to decline
the request while redirecting toward image interpretation. The second pairs CT
interpretation prompts with invalid, non-CT, or anatomically mismatched 3D
inputs and teaches the model not to hallucinate a CT reading. These sets were
audited to remove accidentally valid prompt--image pairs. 
A small set of external invalid-image examples was added for refusal supervision.

\section{Results}
\label{results}

We evaluate NV-Reason-CT on CT-RATE, Merlin, and RAD-ChestCT for abnormality
classification and report generation across chest and abdominal CT. Evaluation
on a held-out NIH cohort extends the assessment to a broader inventory of chest
and abdominal findings. A preliminary study with expert radiologists further
examines the perceived quality and usefulness of the generated reports and
clinical explanations.

\begin{figure}[!hbp]
\centering
\includegraphics[width=\textwidth]{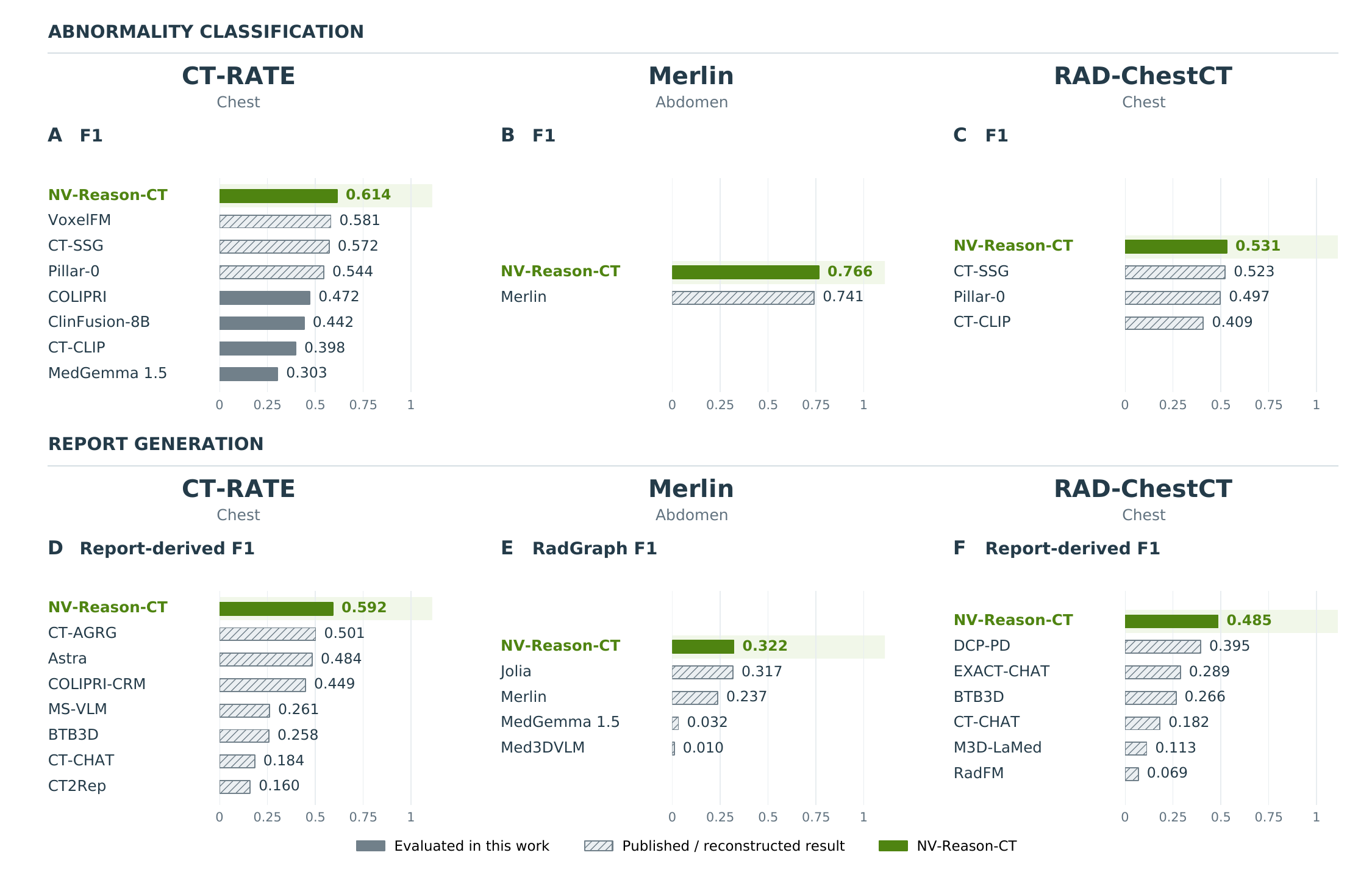}
\caption{Performance across CT benchmarks. Selected abnormality-classification
(A--C) and report-generation (D--F) results on CT-RATE, Merlin, and RAD-ChestCT.
F1 in panels A--D and F denotes positive-class macro-F1 across abnormalities;
panel E reports RadGraph F1 for clinical-entity and relation overlap, using
complete matching for NV-Reason-CT. Green highlights NV-Reason-CT. Solid bars
denote evaluations performed in this work; hatched bars denote published or
reconstructed results. 
Differences in cohorts, training, inference, label mapping, and evaluation are
detailed in Tables~\ref{tab:ctrate_classification},
\ref{tab:ctrate_report_generation}, \ref{tab:merlin_classification},
\ref{tab:merlin_report_generation}, \ref{tab:radchestct_classification_published},
and~\ref{tab:radchestct_report_generation}. All scores are on a $[0,1]$ scale;
higher is better.}
\label{fig:nv_reason_ct_benchmark_summary}
\end{figure}

Figure~\ref{fig:nv_reason_ct_benchmark_summary} summarizes selected classification
and report-generation results across the three public benchmarks. The following
subsections provide the full comparisons and their evaluation protocols.

\subsection{CT-RATE evaluation}
\label{sec:results_ctrate}

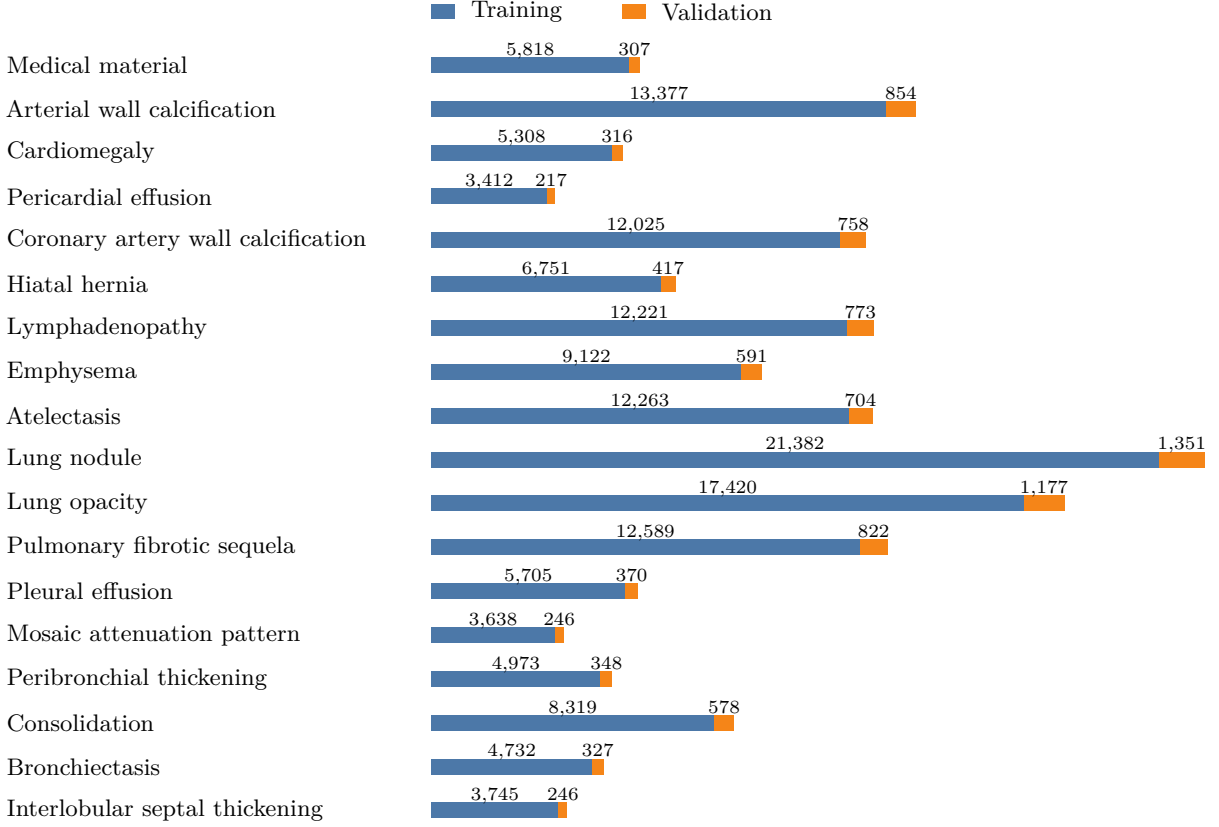
\begin{figure}[t]
\centering
\begingroup
\definecolor{ctrateTrain}{HTML}{4C78A8}
\definecolor{ctrateValidation}{HTML}{F58518}
\newcommand{\ctratebar}[6]{%
  \node[anchor=west, align=left, text width=5.0cm, font=\small]
    at (-12.75,#1) {#2};
  \path[fill=ctrateTrain]
    (0,{#1-0.18}) rectangle (#3,{#1+0.18});
  \path[fill=ctrateValidation]
    (#3,{#1-0.18}) rectangle ({#3+#4},{#1+0.18});
  \node[anchor=base, inner sep=0pt, font=\scriptsize]
    at ({#3/2},{#1+0.23}) {#5};
  \node[anchor=base, inner sep=0pt, font=\scriptsize]
    at ({#3+#4/2},{#1+0.23}) {#6};
}
\begin{adjustbox}{max width=\textwidth}
\begin{tikzpicture}[x=0.45cm,y=0.58cm]
  \path[fill=ctrateTrain] (0,1.05) rectangle (0.7,1.39);
  \node[anchor=west, font=\small] at (0.9,1.22) {Training};
  \path[fill=ctrateValidation] (5.6,1.05) rectangle (6.3,1.39);
  \node[anchor=west, font=\small] at (6.5,1.22) {Validation};

  \ctratebar{0}{Medical material}{5.818}{0.307}{5,818}{307}
  \ctratebar{-1}{Arterial wall calcification}{13.377}{0.854}{13,377}{854}
  \ctratebar{-2}{Cardiomegaly}{5.308}{0.316}{5,308}{316}
  \ctratebar{-3}{Pericardial effusion}{3.412}{0.217}{3,412}{217}
  \ctratebar{-4}{Coronary artery wall calcification}{12.025}{0.758}{12,025}{758}
  \ctratebar{-5}{Hiatal hernia}{6.751}{0.417}{6,751}{417}
  \ctratebar{-6}{Lymphadenopathy}{12.221}{0.773}{12,221}{773}
  \ctratebar{-7}{Emphysema}{9.122}{0.591}{9,122}{591}
  \ctratebar{-8}{Atelectasis}{12.263}{0.704}{12,263}{704}
  \ctratebar{-9}{Lung nodule}{21.382}{1.351}{21,382}{1,351}
  \ctratebar{-10}{Lung opacity}{17.420}{1.177}{17,420}{1,177}
  \ctratebar{-11}{Pulmonary fibrotic sequela}{12.589}{0.822}{12,589}{822}
  \ctratebar{-12}{Pleural effusion}{5.705}{0.370}{5,705}{370}
  \ctratebar{-13}{Mosaic attenuation pattern}{3.638}{0.246}{3,638}{246}
  \ctratebar{-14}{Peribronchial thickening}{4.973}{0.348}{4,973}{348}
  \ctratebar{-15}{Consolidation}{8.319}{0.578}{8,319}{578}
  \ctratebar{-16}{Bronchiectasis}{4.732}{0.327}{4,732}{327}
  \ctratebar{-17}{Interlobular septal thickening}{3.745}{0.246}{3,745}{246}
\end{tikzpicture}
\end{adjustbox}
\endgroup
\caption{CT-RATE class (abnormality) composition in the training and v2 validation
cohorts. Horizontal bars show the number of positive cases for each of the 18
abnormality labels; Labels are not mutually exclusive, several abnormalities can occur in the same CT case.}
\label{fig:ctrate_abnormality_composition}
\end{figure}

Dataset composition and split construction are described in
Section~\ref{sec:data_datasets}. Unless otherwise stated, our CT-RATE
experiments use the corrected v2 validation cohort of 3,002 chest CT
reconstructions, corresponding to 1,564 studies from 1,304 patients. Several
published comparators instead use the historical 3,039-reconstruction manifest
or evaluate one image--report pair for each of the 1,564 studies. We therefore
report the cohort and evaluation unit for each result rather than treating
these protocols as interchangeable. Table~\ref{tab:ctrate_cohort} summarizes the training split and the historical
v1 and corrected v2 validation protocols. A study denotes one CT acquisition,
whereas a case denotes one reconstructed 3D volume; our v2 evaluation treats
each reconstruction as one example~\cite{ctratedataset2026}.
Figure~\ref{fig:ctrate_abnormality_composition} shows positive-case counts for
the 18 abnormalities.

\subsubsection{CT-RATE abnormality classification}
\label{sec:results_ctrate_classification}

\begin{table}[t]
\centering
\small
\caption{CT-RATE cohort composition. A study denotes one CT acquisition and a
case denotes one reconstruction-level 3D volume. Cases from the same study
share a radiology report and its extracted labels. The original 3,039-case
validation manifest was revised to 3,002 cases after 37 non-chest volumes,
predominantly head CTs, were excluded.}
\label{tab:ctrate_cohort}
\renewcommand{\arraystretch}{1.12}
\begin{adjustbox}{max width=\columnwidth}
\begin{tabular}{@{}lrrr@{}}
\toprule
\textbf{Split / manifest} & \textbf{Cases} & \textbf{Studies} &
\textbf{Patients} \\
\midrule
Training & 47,149 & 24,128 & 20,000 \\
Validation (v1 / v2) & 3,039 / 3,002 & 1,564 & 1,304 \\
\midrule
Total (original release) & 50,188 & 25,692 & 21,304 \\
\bottomrule
\end{tabular}
\end{adjustbox}
\end{table}

\paragraph{Unified evaluation strategy.}

Our primary comparison target is image-only classification of all 18 original
CT-RATE abnormalities for all 3,002 cases. No validation report text is supplied
at inference. Our key metrics are macro-F1 score and macro-AUROC. For
abnormality $c$, we compute the positive-class F1 and then average equally
across labels,
\begin{equation}
\operatorname{MacroF1}_{18}=\frac{1}{18}\sum_{c=1}^{18}
\frac{2\operatorname{TP}_{c}}
{2\operatorname{TP}_{c}+\operatorname{FP}_{c}+\operatorname{FN}_{c}}.
\end{equation}
For our continuous-score evaluations, binary predictions use a fixed threshold
of 0.5, $\hat{y}_{ic}=\mathbf{1}\{p_{ic}>0.5\}$. Macro-AUROC is the unweighted
mean of the 18 per-label AUROCs computed from the continuous scores. For
evaluations based on generated Yes/No answers, F1 is computed from the parsed
answers; AUROC requires a separate continuous score.

We emphasize macro-F1 because it measures the precision--recall tradeoff at
the evaluated operating point, including the effect of false-positive
predictions for uncommon abnormalities. AUROC provides complementary
information about ranking across thresholds but does not directly describe
precision and recall at the chosen operating point.

Published CT-RATE results differ in F1 averaging, operating thresholds,
evaluation cohorts, and classification interfaces. Some average
prevalence-weighted positive- and negative-class F1, whereas others report
positive-class F1. Thresholds may be fixed, calibrated on training-derived
data, or optimized on public validation labels. Native prompted inference,
supervised classifier probes, ensembling, and case filtering introduce further
differences. Table~\ref{tab:ctrate_classification} is therefore a
protocol-aware comparison rather than an undifferentiated leaderboard.

Where released checkpoints and inference code were available, we re-evaluated
models on the same 3,002-case v2 manifest using the authors' checkpoint,
model-specific image preprocessing, and the closest documented native
classification interface. For contrastive models, positive and negative text
similarities were normalized as a two-way probability and thresholded at 0.5.
No threshold tuning or calibration was applied in these re-evaluations;
model-specific prompts and additional prompt comparisons are described below.
Retaining each model's native preprocessing and text encoder is necessary for
a faithful reproduction, although it prevents the comparison from isolating
architecture alone. Table~\ref{tab:ctrate_classification} distinguishes our
re-evaluations from published results, including results obtained with
additional supervised classifier training. The per-abnormality F1 and AUROC
results are reported in Appendix Tables~\ref{tab:ctrate_per_abnormality_f1}
and \ref{tab:ctrate_per_abnormality_auroc}, respectively.

\begin{table*}[!htbp]
\centering
\small
\caption{Multi-label abnormality classification evaluation on the official
public validation split. Macro-F1 is the unweighted mean of the 18
positive-class F1 scores (at 0.5 unified threshold when applicable); all metrics are reported on a $[0,1]$ scale. }
\label{tab:ctrate_classification}
\renewcommand{\arraystretch}{1.06}
\begin{adjustbox}{max width=\textwidth}
\setlength{\tabcolsep}{3pt}
\begin{tabular}{@{}>{\raggedright\arraybackslash}p{2cm}
>{\raggedright\arraybackslash}p{3cm}
>{\raggedright\arraybackslash}p{4.7cm}
>{\raggedright\arraybackslash}p{4.4cm}
>{\centering\arraybackslash}p{1.5cm}
>{\centering\arraybackslash}p{1.8cm}@{}}
\toprule
\textbf{Method} & \textbf{Pretraining paradigm} & \textbf{Vision Encoder} &
\textbf{Classification adaptation / Source} & \textbf{Macro-F1 $\uparrow$} &
\textbf{Macro-AUROC $\uparrow$} \\
\midrule
CT-CLIP~\cite{hamamci2026generalist}
& CLIP based image-text contrastive alignment
& Factorized ViT (3D patches 20x20x10, 2D in-plane transformer, followed by out-of-plane, resample 0.75x0.75x1.5mm, crop 480x480x240)
& \multicolumn{3}{@{}l@{}}{%
  \begin{tabular}[t]{@{}>{\raggedright\arraybackslash}p{4.4cm}
  >{\centering\arraybackslash}p{1.5cm}
  >{\centering\arraybackslash}p{1.8cm}@{}}
  Our eval using \emph{Zero-shot} official checkpoint. Pos+Neg prompts. & 0.3976 & 0.7328 \\
  \cmidrule(l){1-3}
  Our eval using \emph{VocabFine} official checkpoint. Pos+Neg prompts. & 0.2662 & 0.7583 \\
  \cmidrule(l){1-3}
  VoxelFM~\cite{morenoaguado2026voxelfm} eval: retrained extra head classifier.
  & 0.2760 & 0.5740 \\
  \end{tabular}%
} \\
\cmidrule(l){4-6}
Merlin~\cite{blankemeier2026merlin}
& CLIP based image-text contrastive alignment
& 3D ConvNet, inflated from ResNet152, (224x224x160 crop, 1.5x1.5x3mm resampling)
& \multicolumn{3}{@{}l@{}}{%
  \begin{tabular}[t]{@{}>{\raggedright\arraybackslash}p{4.4cm}
  >{\centering\arraybackslash}p{1.5cm}
  >{\centering\arraybackslash}p{1.8cm}@{}}
  Our eval using official checkpoint. Pos+Neg prompts. & 0.3579 & 0.6617 \\
  \cmidrule(l){1-3}
  VoxelFM~\cite{morenoaguado2026voxelfm} eval: retrained extra head classifier.
  & 0.4955 & 0.7983 \\
  \end{tabular}%
} \\
\cmidrule(l){4-6}
COLIPRI~\cite{wald2026colipri}
& Multi-objective: CLIP based, VQA, image-only SSL pretraining
& 3D ViT (8x8x8 patch, 192x192x192 voxels input, 2x2x2mm resampling)
& Authors' supervised-probe AUROC; our F1 eval using the released checkpoint and Pos+Neg prompts.
& 0.4717 & 0.8615 \\
\cmidrule(l){4-6}
Jolia~\cite{khlaut2026jolia}
& CLIP based image-text contrastive alignment
& 3D ConvNet stem + transformer (192x192x192 crop, 1.5x1.5x1.5mm resampling, 11 CT-window channels)
& Our eval using official checkpoint.  Contrastive [CLS]-only zero-shot using 8 Pos+Neg prompt pairs.
& 0.3589 & 0.6986 \\
\cmidrule(l){4-6}
Pillar-0 (Chest)~\cite{agrawal2025pillar0}
& CLIP based image-text contrastive alignment
& 3D ConvNet stem+transformer (256x256x256 crop, 1.25x1.25x1.25mm  resampling, 11 CT-window channels)
& DCP-1 eval~\cite{wang2026dcppd}: retrained extra head classifier.
& 0.5440 & 0.8610 \\
\cmidrule(l){4-6}
VoxelFM~\cite{morenoaguado2026voxelfm}
& DINO based image only pretraining & 3D ViT (14x14x14 patch, 112x112x122 voxels input, isotropic resample to smallest spacing)
& Authors' eval: single-layer Q-Former classifier on top of the vision encoder.
& 0.5807 & 0.8701 \\
\cmidrule(l){4-6}
CT-SSG~\cite{dipiazza2026ctssg}
& ResNet18 init & 2.5D ConvNet + graph. crop 480×480x240, 1.5x1.5x0.75 resampling.
& Authors' eval
& 0.5718 & 0.8364 \\
\cmidrule(l){4-6}
RadFM~\cite{wu2023radfm}
& VLM instruction tuning & 3D ViT + Perceiver (32×32×4 patch size, 64-slice input)
& \multicolumn{3}{@{}l@{}}{%
  \begin{tabular}[t]{@{}>{\raggedright\arraybackslash}p{4.4cm}
  >{\centering\arraybackslash}p{1.5cm}
  >{\centering\arraybackslash}p{1.8cm}@{}}
  Our eval using the official checkpoint. Native generated Yes/No answers.
  & 0.2153 & -- \\
  \cmidrule(l){1-3}
  VoxelFM~\cite{morenoaguado2026voxelfm} eval: retrained extra head classifier.
  & 0.4325 & 0.7592 \\
  \end{tabular}%
} \\
\cmidrule(l){4-6}
M3D-LaMed~\cite{bai2024m3dlamed}
& Multi-objective: CLIP based, VQA & 3D ViT (16×16×4 patch size, resize input to 256×256x32)
& \multicolumn{3}{@{}l@{}}{%
  \begin{tabular}[t]{@{}>{\raggedright\arraybackslash}p{4.4cm}
  >{\centering\arraybackslash}p{1.5cm}
  >{\centering\arraybackslash}p{1.8cm}@{}}
  Our eval using the official checkpoint. One-token Yes/No forced choice.
  & 0.3046 & -- \\
  \cmidrule(l){1-3}
  VoxelFM~\cite{morenoaguado2026voxelfm} eval: retrained extra head classifier.
  & 0.4257 & 0.7403 \\
  \end{tabular}%
} \\
\cmidrule(l){4-6}
MedGemma 1.5~\cite{sellergren2026medgemma15}
& Frozen MedSigLIP vision encoder; VQA
& 2D ViT (sequence of up to 85 axial 480x480 slices, 0.75×0.75×1.5 mm resampling)
& Our eval using official checkpoint. Yes/No prompt per abnormality.
& 0.3033 & -- \\
\cmidrule(l){4-6}
ClinFusion~\cite{yuan2026clinfusion}
& 2D/3D VLM instruction tuning
& 2D and 3D ViT fusion (256x256x32 crop, 8x8x8 patch)
& \multicolumn{3}{@{}l@{}}{%
  \begin{tabular}[t]{@{}>{\raggedright\arraybackslash}p{4.4cm}
  >{\centering\arraybackslash}p{1.5cm}
  >{\centering\arraybackslash}p{1.8cm}@{}}
  ClinFusion-8B checkpoint. Single prompt for Yes/No.
  & 0.4419 & -- \\
  \cmidrule(l){1-3}
  ClinFusion-32B checkpoint. Single prompt for Yes/No.
  & 0.3741 & -- \\
  \end{tabular}%
} \\
\cmidrule(l){4-6}
NV-REASON-CT (ours)
& VLM instruction tuning: VQA, reports, reasoning. SFT + GRPO/RL & 3D ViT (8x8x8 patch, 192x192x192 voxels input, 2x2x2mm resampling)
& Single prompt for Yes/No abnormality presence (uniform template) for F1 score. Yes/No token probability for AUROC.
& \textbf{0.6140} & \textbf{0.8710} \\
\bottomrule
\end{tabular}
\end{adjustbox}

\end{table*}

\FloatBarrier

\noindent\textbf{NV-Reason-CT.}
NV-Reason-CT achieved macro-F1 0.6140 using generated binary answers and
macro-AUROC 0.8710 using a separate token-probability evaluation. For macro-F1
evaluation, we used the uniform binary prompt ``\{finding\}: present or absent
in this CT? Answer Yes (for present) or No (for absent).'' We parsed the Yes or
No text output after the generated thinking.

For AUROC, we disabled thinking and applied a two-way softmax to the Yes and No
logits at the first generated position, using the same binary prompt. The
resulting Yes probability provided the continuous score. Thus, the reported
F1 measures generated-answer behavior, whereas AUROC measures ranking in the
separate probability-scoring mode.

Additional uniform binary prompts produced similar results. We also evaluated
a single prompt requesting all present abnormalities from the allowed label
inventory, with ``No Chest Finding'' returned when none was present. This
yielded macro-F1 0.6100, compared with 0.6140 for individual abnormality
questions, while requiring only one inference call per CT.

\medskip
\noindent\textbf{CT-CLIP.}
CT-CLIP~\cite{hamamci2026generalist} was released jointly with CT-RATE as a
baseline for the dataset. CT-CLIP aligns volumetric CT and report
representations using a factorized ViT vision encoder and a CXR-BERT-family
text encoder; its VocabFine variant additionally adapts the text vocabulary to
CT terminology. The original publication reports mean F1 values of 0.7069 and
0.7384 and macro-AUROCs of 0.7308 and 0.7561 for the zero-shot and VocabFine
checkpoints, respectively. These F1 values average prevalence-weighted binary
F1 and use per-label operating points, so they are not directly comparable to
the fixed-threshold positive-class macro-F1 used here.

We re-evaluated both released checkpoints on the 3,002-case v2 cohort with the
authors' volume preprocessing. Each label was scored with
``\{abnormality\} is present.'' and ``\{abnormality\} is not present.''; a
two-way softmax over image--text similarities defined $P(\mathrm{present})$,
with no calibration or prompt variation. We used the prompt formulation
selected and released by the CT-CLIP authors. This yielded macro-F1/AUROC of
0.3976/0.7328 for zero-shot and 0.2662/0.7583 for VocabFine. Separately,
VoxelFM~\cite{morenoaguado2026voxelfm} trained a classifier on frozen CT-CLIP
features and reported macro-F1/AUROC of 0.2760/0.5740 on the historical
3,039-case cohort. Table~\ref{tab:ctrate_classification} includes this
supervised-probe result as a distinct evaluation.

\medskip
\noindent\textbf{Merlin.}
We evaluated the official Merlin checkpoint~\cite{blankemeier2026merlin}, with
its released preprocessing, on all 3,002 CT-RATE v2 cases. L2-normalized image
and text embeddings were compared using the uniform pair
``\{finding\} is present.'' and ``No \{finding\}.'', and the learned logit
scale and a two-way softmax produced the positive probability. At the fixed
0.5 threshold, this native contrastive evaluation obtained
macro-F1/macro-AUROC of 0.3579/0.6617.

Using the CT-CLIP prompt pair, ``\{abnormality\} is present.'' and
``\{abnormality\} is not present.'', yielded lower macro-F1/macro-AUROC of
0.3081/0.5931. Independently, the VoxelFM study trained a supervised classifier
on the frozen Merlin visual encoder, obtaining macro-F1/macro-AUROC of
0.4955/0.7983 on 3,039 reconstructions at a fixed 0.5
threshold~\cite{morenoaguado2026voxelfm}.

\medskip
\noindent\textbf{COLIPRI.}
The COLIPRI publication~\cite{wald2026colipri} reports macro-AUROC 0.8615 on
CT-RATE using a supervised classification probe, but does not report macro-F1.
We re-evaluated the released COLIPRI-CRM checkpoint in its native contrastive
mode on all 3,002 v2 cases. The official processor resampled volumes to 2-mm
isotropic spacing and cropped or padded them to $192\times192\times192$
voxels. For each label, we used ``No \{abnormality\} present'' and
``\{abnormality\} present'', divided the two image--text similarities by the
released temperature, and applied a two-way softmax. We used the prompt
formulation selected by the COLIPRI authors. No threshold tuning or
calibration was performed. This evaluation produced macro-F1 0.4717 and
macro-AUROC 0.8161 and required no additional CT-RATE-supervised training.

Table~\ref{tab:ctrate_classification} reports our contrastive macro-F1
alongside the authors' supervised-probe macro-AUROC; these values come from
different evaluation configurations.

\medskip
\noindent\textbf{Pillar-0.}
The Pillar-0 Chest result~\cite{agrawal2025pillar0} comes from the separate
DCP-PD study, which freezes the Atlas visual encoder and trains a linear
multi-label classifier on CT-RATE training data. This classifier is denoted
DCP-1~\cite{wang2026dcppd}. The reported all-label averages are macro-F1 0.544
and macro-AUROC 0.861 on the historical 3,039-case validation manifest. This
entry therefore evaluates supervised classification on frozen Pillar-0
features.

\medskip
\noindent\textbf{Jolia.}
Jolia~\cite{khlaut2026jolia} reports a short-prompt, [CLS]-only macro-AUROC of
0.7346 using 1,564 acquisition-level examples; a directly comparable
positive-class macro-F1 is not reported. Our evaluation used the released
checkpoint, but scored all 3,002 images. We used the global [CLS] embedding and
the eight positive/negative prompt pairs listed in the publication, averaged
cosine similarity within the positive and negative prompt sets, and applied a
two-way softmax with a fixed 0.5 threshold. We did not use organ-query
features, calibration, or a trained linear probe. The resulting
reconstruction-level macro-F1 was 0.3589 and macro-AUROC was 0.6986. The
difference from the reported AUROC should not be attributed to model variation
alone because the evaluation unit and manifest also differ.

\medskip
\noindent\textbf{VoxelFM.}
VoxelFM~\cite{morenoaguado2026voxelfm} is an image-only volumetric model
pretrained with a DINO-style self-supervised objective. For CT-RATE, the
authors train a single-layer Q-Former classifier using the official training
labels, reserving 20\% of the training split for internal validation, and
evaluate all 18 abnormalities on the historical 3,039-case validation set.
The reported fixed-0.5 result is macro-F1 0.5807 and macro-AUROC 0.8701. The
study provides a complete per-label breakdown and applies the same head and
scoring protocol to several comparator encoders, making it a useful
supervised-probe reference. We use the authors' reported result without
modification.

\medskip
\noindent\textbf{CT-SSG.}
CT-SSG uses an ImageNet-initialized ResNet-18 to encode adjacent-slice
triplets, followed by graph-based spectral aggregation for multilabel CT
classification~\cite{dipiazza2026ctssg}. Its five-fold evaluation reports
macro-F1 0.5718 and macro-AUROC 0.8364 on the historical public validation
cohort, with all 18 per-label values available.

\medskip
\noindent\textbf{RadFM.}
We evaluated the public RadFM checkpoint~\cite{wu2023radfm} following the
released inference example, querying each abnormality with ``Can you identify
any visible signs of \{abnormality\} in the image?''. The released volume
loader was adapted to CT-RATE's array orientation before resizing axial
slices to $512\times512$, resizing depth to 64, replicating the grayscale
volume across three channels, and applying per-volume min--max normalization;
no HU windowing or physical-spacing resampling was used. Native generation
yielded positive-class macro-F1 0.2153.

Separately, VoxelFM discarded the language-model component and trained a
supervised classifier on the frozen RadFM visual encoder, obtaining macro-F1
0.4325 and macro-AUROC 0.7592 on the historical 3,039-case
cohort~\cite{morenoaguado2026voxelfm}. The two rows in
Table~\ref{tab:ctrate_classification} therefore measure native prompted
behavior and supervised representation transfer, respectively.

\medskip
\noindent\textbf{M3D-LaMed.}
We evaluated the public M3D-LaMed Phi-3-4B checkpoint~\cite{bai2024m3dlamed}.
Each CT image was transposed to depth-first order, min--max normalized,
foreground-cropped, and resized to $32\times256\times256$. This is an
independent raw-volume adaptation because M3D-LaMed was trained with rendered
8-bit slice stacks and its release does not define raw-HU CT-RATE windowing.
We used ``Is \{abnormality\} present in this CT scan? Answer only Yes or No.''
This evaluation yielded macro-F1 0.3046. Separately, VoxelFM trained a
supervised classifier on the frozen M3D visual encoder, without the language
model, and reported macro-F1 0.4257 and macro-AUROC 0.7403 on the historical
3,039-case cohort~\cite{morenoaguado2026voxelfm}.

\medskip
\noindent\textbf{MedGemma 1.5.}
MedGemma 1.5 is a 4B medical 2D VLM with a MedSigLIP visual encoder that
accepts a sequence of sampled 2D CT slices rather than a native 3D
image~\cite{sellergren2026medgemma15}. The technical report evaluates up to
85 axial slices per scan and reports CT-RATE macro-F1 values from 0.235 to
0.269 on 1,558 acquisition-level examples.

Our evaluation uses the official checkpoint, the report's three-window slice
construction, $0.75\times0.75\times1.5$-mm resampling, and the published
per-abnormality prompt: ``You are an expert radiologist for chest CT. Looking
at these CT slices, is there \{abnormality\}? Answer with `Final Answer: yes'
or `Final Answer: no'.'' This generated-answer evaluation yielded macro-F1
0.3033. AUROC was not computed from the discrete answers.

\medskip
\noindent\textbf{ClinFusion-8B and ClinFusion-32B.}
We independently evaluated both official checkpoints: ClinFusion-8B built on
Qwen3-VL-8B-Instruct, and ClinFusion-32B built on
Qwen3-VL-32B-Instruct~\cite{yuan2026clinfusion}. Both evaluations covered all
18 labels and all 3,002 CT-RATE validation cases. We used 1-mm isotropic
resampling, resizing to $32\times256\times256$, and deterministic selection
of four 2D anchor slices; both the 3D and 2D visual streams were retained. Each
abnormality was evaluated independently with ``Is \{label\} present in this CT
scan? Answer only Yes or No.'' Macro-F1 was 0.4419 for ClinFusion-8B and 0.3741
for ClinFusion-32B. These results reflect our independently implemented
evaluation and prompt selection; alternative protocols or prompt formulations
may yield different results.

\paragraph{Notable results excluded from the direct comparison.}
AnyMC3D~\cite{liu2026anymc3d} reports macro-F1 0.646 with class-specific
calibration for a ten-model ensemble learned from a separate training-derived
holdout; the same ensemble obtained 0.426 without calibration. We do not
include the calibrated ensemble in Table~\ref{tab:ctrate_classification}
because it changes both model multiplicity and class-specific operating
points relative to the single-model evaluations emphasized here. Other
non-matched results in the table are retained only with the explicit protocol
qualifications above.

fVLM~\cite{shui2025fvlm} is also not directly included because its released
evaluator requires corrected-HU volumes and TotalSegmentator-derived masks
for four anatomical regions. It provides organ-conditioned zero-shot scores
for only 16 of the 18 CT-RATE labels.

\newcommand{\ctratePerAbnormalityFOneTables}{%
\begin{table*}[!htbp]
\centering
\small
\caption{Per-abnormality positive-class F1 on the CT-RATE public validation
set (Part 1 of 2). }
\label{tab:ctrate_per_abnormality_f1}
\renewcommand{\arraystretch}{1.06}
\begin{adjustbox}{max width=\textwidth}
\begin{tabular}{@{}lrrrrrrr@{}}
\toprule
& \multicolumn{1}{c}{\textbf{Proposed}} &
\multicolumn{3}{c}{\textbf{CT-CLIP}} &
\multicolumn{2}{c}{\textbf{Merlin}} &
\multicolumn{1}{c}{\textbf{COLIPRI}} \\
\cmidrule(lr){2-2}\cmidrule(lr){3-5}\cmidrule(lr){6-7}\cmidrule(l){8-8}
\textbf{Abnormality} &
\shortstack{\textbf{NV-Reason-CT}\\\textbf{(ours)}} &
\shortstack{\textbf{Ours}\\\textbf{(Zero-shot)}} &
\shortstack{\textbf{Ours}\\\textbf{(VocabFine)}} &
\shortstack{\textbf{VoxelFM}\\\textbf{eval.}} &
\shortstack{\textbf{Ours}\\\textbf{(official ckpt)}} &
\shortstack{\textbf{VoxelFM}\\\textbf{eval.}} &
\shortstack{\textbf{Ours}\\\textbf{(released ckpt)}} \\
\midrule
Medical material                   & 0.7270 & 0.2672 & 0.0364 & 0.1763 & 0.1956 & 0.5607 & 0.2677 \\
Arterial wall calcification        & 0.7750 & 0.6546 & 0.6475 & 0.4452 & 0.5436 & 0.7278 & 0.7337 \\
Cardiomegaly                       & 0.6390 & 0.3330 & 0.5303 & 0.2300 & 0.2733 & 0.4817 & 0.5689 \\
Pericardial effusion               & 0.5780 & 0.2136 & 0.0000 & 0.1532 & 0.2202 & 0.3028 & 0.4263 \\
Coronary artery wall calcification & 0.7770 & 0.6139 & 0.6029 & 0.4419 & 0.5294 & 0.6890 & 0.6550 \\
Hiatal hernia                      & 0.5410 & 0.3369 & 0.0828 & 0.2444 & 0.2644 & 0.3574 & 0.3565 \\
Lymphadenopathy                    & 0.5750 & 0.5082 & 0.3003 & 0.3728 & 0.3881 & 0.4734 & 0.4181 \\
Emphysema                          & 0.5270 & 0.4435 & 0.2828 & 0.3413 & 0.3395 & 0.4676 & 0.3706 \\
Atelectasis                        & 0.6040 & 0.4362 & 0.3473 & 0.3482 & 0.3012 & 0.4880 & 0.5069 \\
Lung nodule                        & 0.6590 & 0.5229 & 0.5781 & 0.4738 & 0.5845 & 0.6181 & 0.6332 \\
Lung opacity                       & 0.7450 & 0.5504 & 0.3262 & 0.4501 & 0.5668 & 0.6168 & 0.6796 \\
Pulmonary fibrotic sequela         & 0.5410 & 0.4104 & 0.2130 & 0.0000 & 0.3420 & 0.4839 & 0.4954 \\
Pleural effusion                   & 0.8590 & 0.3863 & 0.6110 & 0.2644 & 0.5674 & 0.7135 & 0.8302 \\
Mosaic attenuation pattern         & 0.4520 & 0.2545 & 0.0472 & 0.1653 & 0.1554 & 0.3252 & 0.2176 \\
Peribronchial thickening           & 0.4410 & 0.2977 & 0.0000 & 0.2262 & 0.2383 & 0.3766 & 0.2591 \\
Consolidation                      & 0.7120 & 0.4398 & 0.1696 & 0.2835 & 0.4551 & 0.5395 & 0.5807 \\
Bronchiectasis                     & 0.4680 & 0.2397 & 0.0000 & 0.1879 & 0.1850 & 0.2929 & 0.2226 \\
Interlobular septal thickening     & 0.4290 & 0.2476 & 0.0160 & 0.1627 & 0.2920 & 0.4043 & 0.2679 \\
\midrule
\textbf{Macro-F1}                  & \textbf{0.6140} & \textbf{0.3976} &
\textbf{0.2662} & \textbf{0.2760} & \textbf{0.3579} & \textbf{0.4955} &
\textbf{0.4717} \\
\bottomrule
\end{tabular}
\end{adjustbox}
\vspace{3pt}
\end{table*}

\begin{table*}[!htbp]
\ContinuedFloat
\centering
\small
\caption{Per-abnormality positive-class F1 on the CT-RATE public validation
set (Part 2 of 2).}
\label{tab:ctrate_per_abnormality_f1_part2}
\renewcommand{\arraystretch}{1.06}
\begin{adjustbox}{max width=\textwidth}
\begin{tabular}{@{}lrrrrrrrrr@{}}
\toprule
\textbf{Abnormality} & \textbf{VoxelFM} & \textbf{CT-SSG} &
\textbf{RadFM} & \textbf{M3D-LaMed} & \textbf{Jolia} &
\shortstack{\textbf{Pillar-0}\\\textbf{(DCP-1)}} &
\shortstack{\textbf{MedGemma}\\\textbf{1.5}} &
\shortstack{\textbf{ClinFusion}\\\textbf{8B}} &
\shortstack{\textbf{ClinFusion}\\\textbf{32B}} \\
\midrule
Medical material                   & 0.6269 & 0.6310 & 0.3402 & 0.3579 & 0.2525 & 0.5940 & 0.2605 & 0.3731 & 0.1829 \\
Arterial wall calcification        & 0.7849 & 0.7780 & 0.6474 & 0.6441 & 0.5248 & 0.7430 & 0.5392 & 0.6579 & 0.6386 \\
Cardiomegaly                       & 0.5410 & 0.6240 & 0.5279 & 0.5216 & 0.2222 & 0.5640 & 0.4415 & 0.4441 & 0.3909 \\
Pericardial effusion               & 0.4916 & 0.5540 & 0.2557 & 0.3090 & 0.2228 & 0.3800 & 0.0090 & 0.3189 & 0.1172 \\
Coronary artery wall calcification & 0.7663 & 0.7670 & 0.6373 & 0.5030 & 0.4817 & 0.7180 & 0.4344 & 0.5827 & 0.5747 \\
Hiatal hernia                      & 0.4125 & 0.4210 & 0.3197 & 0.3147 & 0.2733 & 0.3860 & 0.2439 & 0.2439 & 0.3294 \\
Lymphadenopathy                    & 0.5554 & 0.5200 & 0.4607 & 0.4330 & 0.4112 & 0.3680 & 0.3976 & 0.4825 & 0.0743 \\
Emphysema                          & 0.5356 & 0.5090 & 0.4322 & 0.3559 & 0.3595 & 0.5040 & 0.3350 & 0.4007 & 0.3668 \\
Atelectasis                        & 0.5537 & 0.5210 & 0.4096 & 0.4123 & 0.4046 & 0.6060 & 0.3802 & 0.4491 & 0.3579 \\
Lung nodule                        & 0.6661 & 0.6370 & 0.5915 & 0.6180 & 0.4802 & 0.6960 & 0.1659 & 0.6348 & 0.6214 \\
Lung opacity                       & 0.7194 & 0.7380 & 0.5873 & 0.5723 & 0.6193 & 0.7910 & 0.6037 & 0.6252 & 0.6345 \\
Pulmonary fibrotic sequela         & 0.5405 & 0.4670 & 0.3183 & 0.4407 & 0.4208 & 0.3510 & 0.1103 & 0.4349 & 0.4437 \\
Pleural effusion                   & 0.8400 & 0.8250 & 0.6158 & 0.6257 & 0.3823 & 0.8470 & 0.4884 & 0.5195 & 0.4100 \\
Mosaic attenuation pattern         & 0.5090 & 0.4560 & 0.3114 & 0.2970 & 0.2509 & 0.3980 & 0.1631 & 0.3174 & 0.0394 \\
Peribronchial thickening           & 0.4017 & 0.3940 & 0.2755 & 0.2850 & 0.2292 & 0.4020 & 0.2078 & 0.3398 & 0.3556 \\
Consolidation                      & 0.6700 & 0.6410 & 0.4636 & 0.4387 & 0.5275 & 0.6690 & 0.2468 & 0.5009 & 0.5088 \\
Bronchiectasis                     & 0.4013 & 0.3910 & 0.2492 & 0.2603 & 0.2044 & 0.3830 & 0.2028 & 0.2740 & 0.3017 \\
Interlobular septal thickening     & 0.4374 & 0.4180 & 0.3409 & 0.2732 & 0.1925 & 0.3860 & 0.2285 & 0.3551 & 0.3870 \\
\midrule
\textbf{Macro-F1}                  & \textbf{0.5807} & \textbf{0.5718} &
\textbf{0.4325} & \textbf{0.4257} & \textbf{0.3589} &
\textbf{0.5440} & \textbf{0.3033} & \textbf{0.4419} & \textbf{0.3741} \\
\bottomrule
\end{tabular}
\end{adjustbox}
\vspace{3pt}
\end{table*}
}%

\newcommand{\ctratePerAbnormalityAurocTables}{%
\begin{table*}[!htbp]
\centering
\small
\caption{Per-abnormality AUROC on the CT-RATE public validation set (Part 1
of 2). }
\label{tab:ctrate_per_abnormality_auroc}
\renewcommand{\arraystretch}{1.06}
\begin{adjustbox}{max width=\textwidth}
\begin{tabular}{@{}lrrrrrrr@{}}
\toprule
& \multicolumn{1}{c}{\textbf{Proposed}} &
\multicolumn{3}{c}{\textbf{CT-CLIP}} &
\multicolumn{2}{c}{\textbf{Merlin}} &
\multicolumn{1}{c}{\textbf{COLIPRI}} \\
\cmidrule(lr){2-2}\cmidrule(lr){3-5}\cmidrule(lr){6-7}\cmidrule(l){8-8}
\textbf{Abnormality} &
\shortstack{\textbf{NV-Reason-CT}\\\textbf{(ours)}} &
\shortstack{\textbf{Ours}\\\textbf{(Zero-shot)}} &
\shortstack{\textbf{Ours}\\\textbf{(VocabFine)}} &
\shortstack{\textbf{VoxelFM}\\\textbf{eval.}} &
\shortstack{\textbf{Ours}\\\textbf{(official ckpt)}} &
\shortstack{\textbf{VoxelFM}\\\textbf{eval.}} &
\shortstack{\textbf{Ours}\\\textbf{(released ckpt)}} \\
\midrule
Medical material                   & 0.9520 & 0.7172 & 0.7532 & 0.5453 & 0.6136 & 0.8815 & 0.7357 \\
Arterial wall calcification        & 0.9280 & 0.8557 & 0.8758 & 0.6973 & 0.7458 & 0.9016 & 0.9317 \\
Cardiomegaly                       & 0.9420 & 0.8582 & 0.9200 & 0.6037 & 0.7901 & 0.9172 & 0.9297 \\
Pericardial effusion               & 0.9200 & 0.7728 & 0.7818 & 0.5620 & 0.7526 & 0.8206 & 0.8479 \\
Coronary artery wall calcification & 0.9350 & 0.8533 & 0.8600 & 0.6786 & 0.7433 & 0.8976 & 0.9220 \\
Hiatal hernia                      & 0.8430 & 0.7055 & 0.7208 & 0.5722 & 0.5713 & 0.7318 & 0.7108 \\
Lymphadenopathy                    & 0.7810 & 0.7001 & 0.7058 & 0.5569 & 0.6013 & 0.6993 & 0.7552 \\
Emphysema                          & 0.8130 & 0.7426 & 0.7518 & 0.5835 & 0.6437 & 0.7545 & 0.7690 \\
Atelectasis                        & 0.8240 & 0.6727 & 0.6904 & 0.5812 & 0.5949 & 0.7342 & 0.7744 \\
Lung nodule                        & 0.7800 & 0.5716 & 0.6525 & 0.5144 & 0.4944 & 0.6487 & 0.7065 \\
Lung opacity                       & 0.8730 & 0.6401 & 0.6948 & 0.5547 & 0.6720 & 0.7451 & 0.8056 \\
Pulmonary fibrotic sequela         & 0.7650 & 0.5778 & 0.6479 & 0.5071 & 0.5389 & 0.6732 & 0.7055 \\
Pleural effusion                   & 0.9750 & 0.9074 & 0.9344 & 0.6146 & 0.9136 & 0.9370 & 0.9668 \\
Mosaic attenuation pattern         & 0.8960 & 0.7745 & 0.7781 & 0.5476 & 0.5913 & 0.8331 & 0.7777 \\
Peribronchial thickening           & 0.8130 & 0.6913 & 0.7094 & 0.5898 & 0.6150 & 0.7829 & 0.8210 \\
Consolidation                      & 0.9280 & 0.7307 & 0.7508 & 0.5253 & 0.7241 & 0.8133 & 0.9007 \\
Bronchiectasis                     & 0.8180 & 0.6536 & 0.6138 & 0.5378 & 0.5451 & 0.7436 & 0.7976 \\
Interlobular septal thickening     & 0.8870 & 0.7655 & 0.8077 & 0.5593 & 0.7593 & 0.8550 & 0.8319 \\
\midrule
\textbf{Macro-AUROC}               & \textbf{0.8710} & \textbf{0.7328} &
\textbf{0.7583} & \textbf{0.5740} & \textbf{0.6617} & \textbf{0.7983} &
\textbf{0.8161} \\
\bottomrule
\end{tabular}
\end{adjustbox}
\vspace{3pt}
\end{table*}

\begin{table*}[!htbp]
\ContinuedFloat
\centering
\small
\caption{Per-abnormality AUROC on the CT-RATE public validation set (Part 2
of 2). }
\label{tab:ctrate_per_abnormality_auroc_part2}
\renewcommand{\arraystretch}{1.06}
\setlength{\tabcolsep}{8pt}
\begin{adjustbox}{max width=\textwidth}
\begin{tabular}{@{}lrrrrrr@{}}
\toprule
\textbf{Abnormality} & \textbf{VoxelFM} & \textbf{CT-SSG} &
\textbf{RadFM} & \textbf{M3D-LaMed} & \textbf{Jolia} &
\shortstack{\textbf{Pillar-0}\\\textbf{(DCP-1)}} \\
\midrule
Medical material                   & 0.9423 & 0.9050 & 0.7402 & 0.7749 & 0.7319 & 0.9340 \\
Arterial wall calcification        & 0.9436 & 0.9280 & 0.8730 & 0.8351 & 0.7444 & 0.9140 \\
Cardiomegaly                       & 0.9358 & 0.9350 & 0.9096 & 0.9009 & 0.8011 & 0.9370 \\
Pericardial effusion               & 0.8963 & 0.9020 & 0.7953 & 0.7987 & 0.7939 & 0.8580 \\
Coronary artery wall calcification & 0.9420 & 0.9310 & 0.8562 & 0.8202 & 0.7504 & 0.9300 \\
Hiatal hernia                      & 0.8140 & 0.7730 & 0.7313 & 0.7071 & 0.6032 & 0.7590 \\
Lymphadenopathy                    & 0.7700 & 0.7430 & 0.6960 & 0.6659 & 0.5843 & 0.8830 \\
Emphysema                          & 0.8367 & 0.7890 & 0.7310 & 0.6971 & 0.6896 & 0.8110 \\
Atelectasis                        & 0.8057 & 0.7760 & 0.6837 & 0.6763 & 0.6404 & 0.8050 \\
Lung nodule                        & 0.7863 & 0.6260 & 0.6322 & 0.5936 & 0.5096 & 0.7940 \\
Lung opacity                       & 0.8849 & 0.8660 & 0.7079 & 0.6335 & 0.7302 & 0.8850 \\
Pulmonary fibrotic sequela         & 0.7500 & 0.6830 & 0.6227 & 0.6188 & 0.5497 & 0.7060 \\
Pleural effusion                   & 0.9738 & 0.9690 & 0.9323 & 0.9115 & 0.8991 & 0.9820 \\
Mosaic attenuation pattern         & 0.9057 & 0.8840 & 0.8009 & 0.8190 & 0.7849 & 0.8730 \\
Peribronchial thickening           & 0.8043 & 0.7980 & 0.7218 & 0.6959 & 0.5865 & 0.8140 \\
Consolidation                      & 0.9210 & 0.9010 & 0.7410 & 0.7267 & 0.8228 & 0.9260 \\
Bronchiectasis                     & 0.8495 & 0.7870 & 0.6786 & 0.6832 & 0.5604 & 0.7890 \\
Interlobular septal thickening     & 0.9007 & 0.8600 & 0.8113 & 0.7673 & 0.7914 & 0.8920 \\
\midrule
\textbf{Macro-AUROC}               & \textbf{0.8701} & \textbf{0.8364} &
\textbf{0.7592} & \textbf{0.7403} & \textbf{0.6986} &
\textbf{0.8610} \\
\bottomrule
\end{tabular}
\end{adjustbox}
\vspace{3pt}
\end{table*}
}%

\subsubsection{CT-RATE report generation}
\label{sec:results_ctrate_reports}

We evaluate chest CT report generation on the public CT-RATE validation cohort
described in Section~\ref{sec:results_ctrate}. Our unified evaluation uses all
3,002 chest reconstructions in the corrected v2 cohort. NV-REASON-CT generates the Findings and Impression
sections without access to the reference report. Clinical-content metrics are
computed from the 18 CT-RATE abnormalities, whereas text metrics compare the
generated sections with the corresponding reference sections. Because several
reconstructions can originate from the same study, they share one acquisition-
level reference report. Published comparators variously evaluate the historical
3,039 reconstructions or 1,564 study--report pairs; we therefore state the
evaluation unit and cohort for every result in
Table~\ref{tab:ctrate_report_generation}.

We organize the reported metrics into clinical-content and report-similarity
measures. For report-derived Macro-F1 metric, the official CT-RATE RadBERT labeler
maps each generated report to 18 abnormality probabilities, which are
thresholded at 0.5 and compared with the reference labels. Positive-class F1
is computed independently for each abnormality and averaged with equal weight
across all 18 labels. CRG metric uses the resulting true-positive, false-negative,
and false-positive counts but excludes true negatives; it applies
prevalence-dependent rewards and penalties before normalization, reducing the
inflation produced by class imbalance and trivial predominantly normal
reports~\cite{hamamci2025crg}. GREEN uses a learned evaluator to identify
clinically relevant errors relative to a reference report. BLEU-4 measures
four-gram precision with a brevity penalty, ROUGE-L measures longest-common-
subsequence overlap, and METEOR aligns unigrams while allowing stemming and
synonym matches. All metrics in Table~\ref{tab:ctrate_report_generation} are
reported on a $[0,1]$ scale, with higher values indicating better performance.

We treat report-derived Macro-F1 and CRG as the primary clinically oriented
metrics in this evaluation. Both directly test whether abnormalities are
correctly communicated and penalize missed findings and unsupported findings;
Macro-F1 gives rare and common abnormalities equal weight, while CRG further
accounts for label prevalence and is not increased by abundant true negatives.
The natural language metrics BLEU-4, ROUGE-L, and METEOR primarily reward n-gram or
sequence agreement with a single report and are therefore sensitive to
clinically inconsequential differences in terminology, sentence order, section
organization, and report length. This consideration is particularly relevant
because NV-REASON-CT produces standardized Findings and Impression sections,
whereas the CT-RATE references retain heterogeneous narrative styles. Prior CT
report-generation studies have similarly noted that natural-language metrics
may fail to recognize semantically equivalent phrasing or fully capture
diagnostic fidelity~\cite{lee2024msvlm,hamamci2026generalist,hamamci2025crg,sharma2026ctestmetric};
GREEN improves on purely lexical
comparison but remains dependent on a reference report and a learned evaluator.
Its official checkpoint was fine-tuned primarily on chest X-ray reports, making
its application to 3D CT reports out of domain and potentially less
reliable~\cite{hamamci2025crg}. We report all metrics for comparability while
placing greater interpretive weight on clinically grounded abnormality metrics.

\begin{table*}[!htbp]
\centering
\scriptsize
\caption{CT-RATE radiology report-generation results.
Unless otherwise noted, F1 denotes positive-class macro-F1 over the 18
CT-RATE abnormalities extracted from generated reports. All scores are reported
on a $[0,1]$ scale.}
\label{tab:ctrate_report_generation}
\renewcommand{\arraystretch}{1.13}
\setlength{\tabcolsep}{2.4pt}
\begin{adjustbox}{max width=\textwidth}
\begin{tabular}{@{}p{2.05cm}p{6.25cm}cccccc@{}}
\toprule
\textbf{Method} & \textbf{Evaluation details} &
\textbf{F1} $\uparrow$ &
\textbf{CRG} $\uparrow$ &
\textbf{GREEN} $\uparrow$ &
\shortstack{\textbf{BLEU-4}\\$\uparrow$} &
\shortstack{\textbf{ROUGE-L}\\$\uparrow$} &
\textbf{METEOR} $\uparrow$ \\
\midrule
CT2Rep~\cite{hamamci2024ct2rep,hamamci2025crg} &
Authors' evaluation on 3,039 reconstructions &
0.160 & 0.359 & 0.487 & 0.213 & 0.362 & 0.197 \\
CT-CHAT~\cite{hamamci2026generalist,hamamci2025crg} &
Authors' evaluation on 3,039 reconstructions &
0.184 & 0.368 & 0.436 & 0.198 & 0.326 & 0.215 \\
BTB3D~\cite{hamamci2025btb3d} &
Authors' evaluation on 3,039 reconstructions &
0.258 & 0.370 & -- & 0.215 & -- & 0.223 \\
RadFM~\cite{wu2023radfm,hamamci2025crg} &
CRG evaluation~\cite{hamamci2025crg}; official pretrained checkpoint &
0.059 & 0.335 & 0.018 & 0.000 & 0.042 & 0.020 \\
CT-SSG~\cite{dipiazza2026ctssg} &
Authors' evaluation on the public CT-RATE validation partition; five-fold mean; F1 averaging not specified &
0.3872 & 0.4366 & -- & 0.107 & 0.246 & 0.164 \\
MS-VLM~\cite{lee2024msvlm} &
Authors' evaluation on 3,039 reconstructions &
0.261 & -- & -- & 0.232 & 0.438 & 0.396 \\
M3D-LaMed~\cite{bai2024m3dlamed} &
CT-Agent~\cite{mao2025ctagent} evaluation; micro-F1 &
0.148 & -- & -- & 0.245 & 0.400 & 0.326 \\
CT-AGRG~\cite{dipiazza_2025_ctagrg} &
Authors' evaluation on 3,039 reconstructions; CT-Net variant &
0.5011 & -- & -- & 0.172 & 0.280 & 0.196 \\
RMR~\cite{suprijadi2026resolution} &
1,564 test reports; reconstruction selection or aggregation not specified &
0.495 & -- & -- & 0.203 & 0.280 & 0.235 \\
MonteRET~\cite{lin2026monteret} &
Authors' evaluation on 1,564 scans; macro-F1 derived from reported per-finding F1 values; reconstruction mapping not specified &
0.365 & -- & -- & 0.252 & 0.379 & 0.454 \\
Astra~\cite{wang2026astra} &
Authors' evaluation on 1,564 acquisition-level cases with harmonized references; macro-F1 approximated from label-wise F1 bars in Fig.~7e~\cite{wang2026astra} &
0.484 & -- & -- & 0.2501 & 0.4409 & 0.2397 \\
COLIPRI-CRM~\cite{wald2026colipri} &
Evaluation cohort size not specified &
0.449 & -- & -- & -- & -- & -- \\
VoxelFM~\cite{morenoaguado2026voxelfm} &
3,039 reconstructions; reformatted reference reports and GPT-OSS-120B abnormality extraction &
0.432 & -- & -- & -- & -- & -- \\
ClinFusion-8B~\cite{yuan2026clinfusion} &
3,039 reconstructions; reference-derived clinical context; GPT-4.1 claim-level F1 &
0.216 & -- & -- & -- & -- & -- \\
ClinFusion-32B~\cite{yuan2026clinfusion} &
3,039 reconstructions; reference-derived clinical context; GPT-4.1 claim-level F1 &
0.239 & -- & -- & -- & -- & -- \\
\midrule
\textbf{NV-REASON-CT (ours)} &
Our evaluation on 3,002 v2 reconstructions; RadBERT extraction from generated Findings and Impression &
\textbf{0.5923} & \textbf{0.5138} & 0.40722 & 0.1145 & 0.2140 & 0.1921 \\
\bottomrule
\end{tabular}
\end{adjustbox}
\end{table*}

\FloatBarrier

\noindent\textbf{NV-REASON-CT.}
NV-REASON-CT is our native 3D generative VLM, trained with supervised
fine-tuning and reinforcement learning on VQA, structured-report, and
reasoning tasks.

The report evaluation compares generated
Findings and Impression with the corresponding reference sections for all
3,002 reconstructions (1,564 studies
from 1,304 patients). Generated thinking text, technique, and the auxiliary structured
label list are excluded from report scoring. We apply the official CT-RATE
RadBERT label extractor with a fixed 0.5 sigmoid threshold to obtain the
report-derived macro-F1 of 0.5923 and use the same extracted labels for CRG.
BLEU-4, ROUGE-L, and METEOR are computed with the COCO-caption implementations,
and GREEN is computed with its official evaluator. The two F1 values therefore
measure distinct outputs and should not be treated as interchangeable.

Our generated reports end with a summary list of detected abnormalities, including findings from the 18-label CT-RATE inventory. This list allows direct computation of F1 without the RadBERT extractor.  Using such direct computation, our model achieves 0.6010 macro-F1 score from the report.  In Table~\ref{tab:ctrate_report_generation} we decided to report the RadBERT report derived macro-F1 of 0.5923 for consistency with other methods.

Compared to binary QA per-abnormality prompt F1 score of 0.6140 in Table~\ref{tab:ctrate_classification} the report based F1 score of 0.6010 is slightly lower, but includes reacher information (location and severity of abnormalities, or abnormalities beyond 18 ct-rate labels).

\medskip

\noindent\textbf{CT2Rep.}
CT2Rep~\cite{hamamci2024ct2rep} is an early 3D sequence-to-sequence report generator trained from scratch on CT-RATE findings. Its visual encoder operates on
volumetric patches, and a Transformer decoder combines relational memory with
memory-conditioned normalization to generate report text. It comes from the authors of CT-RATE dataset, which they later evaluated on report generation in CRG study of the released model on the historical 3,039-reconstruction
validation cohort, using an 18-label report classifier and standardized text
metrics~\cite{hamamci2025crg}.

\medskip
\noindent\textbf{CT-CHAT.}
CT-CHAT~\cite{hamamci2025crg} connects the frozen CT-CLIP volumetric encoder to Llama~3.1 through a learned projector~\cite{hamamci2026generalist}. It first aligns CT features to
report text and then applies visual instruction tuning with low-rank language
model adaptation;
The report benchmark uses the 8B language-model
configuration and the historical 3,039 reconstruction-level validation
examples.

\medskip
\noindent\textbf{BTB3D.}
BTB3D learns discrete volumetric tokens with a causal 3D convolutional
encoder--decoder, Haar-wavelet compression, and lookup-free quantization, then
maps those tokens to a Llama~3.1-8B report decoder~\cite{hamamci2025btb3d}.
We use the best reported $16\times16\times8$
tokenization result. Its test set is the historical public cohort of 3,039
images.

\medskip
\noindent\textbf{RadFM.}
RadFM is a generalist medical VLM comprising a 3D ViT, a Perceiver-style token
resampler, and a MedLLaMA-13B LLM~\cite{wu2023radfm}. CRG evaluates the official pretrained model directly on
the historical 3,039 CT-RATE reconstructions, without CT-RATE-specific report
retraining~\cite{hamamci2025crg}.

\medskip
\noindent\textbf{MS-VLM.}
MS-VLM combines a self-supervised 2D DINO slice encoder with a Z-former for
inter-slice modeling and a learned bridge to a largely frozen language
model~\cite{lee2024msvlm}. After representation alignment, it is jointly tuned
for report generation and VQA. It's evaluated on 3,039 images.

\medskip
\noindent\textbf{M3D-LaMed.}
M3D-LaMed pairs a native 3D ViT with a Llama-family language model and is
trained through image--text contrastive alignment followed by multimodal
instruction tuning on the M3D-Data corpus~\cite{bai2024m3dlamed}. Its original
publication evaluates report generation on M3D-Cap rather than CT-RATE. The
CT-RATE values in Table~\ref{tab:ctrate_report_generation} come only from the
CT-Agent comparison on the stated 3,039-case cohort and use CT-Agent's
micro-averaged clinical-efficacy protocol~\cite{mao2025ctagent}.

\medskip
\noindent\textbf{CT-AGRG.}
CT-AGRG first trains a CT-Net or CT-ViT multilabel classifier, then uses a
PubMed-pretrained GPT-2 decoder to generate one sentence for each predicted
positive abnormality and concatenates the sentences into a
report~\cite{dipiazza_2025_ctagrg}. The classifier uses abnormality-specific
thresholds selected on an internal validation split, so report content is
explicitly gated by supervised label decisions rather than generated as a
single open-ended narrative. The reported CT-Net result is the mean of five
runs on 3,039 images.

\medskip
\noindent\textbf{COLIPRI-CRM.}
COLIPRI pretrains a 3D ViT using image--report contrastive learning, report
generation, opposite-sentence discrimination, and image-only self-supervision~\cite{wald2026colipri}. For its CRM report experiment, the visual encoder
is frozen and a two-layer projector and Qwen2.5-1B decoder are trained on
CT-RATE findings with next-token supervision. The authors report RadBERT
macro-F1 0.449 on a withheld CT-RATE test set, but do not state its number of
images. The cohort equivalence to either 3,039
reconstructions or 1,564 report-level studies remains unknown.

\medskip
\noindent\textbf{Resolution Meets Reduction (RMR).}
RMR~\cite{suprijadi2026resolution} uses a Curia-64S slice encoder, a Perceiver resampler, and Gemma3-4B; the
vision encoder is frozen while the projector and language-model LoRA weights
are optimized with next-token report supervision. The evaluation is on 1,564 official
validation reports. Evaluation is therefore acquisition/report-level, whereas
the 3,039 protocol is reconstruction-level. The exact image selection (or aggregation) is not specified.

\medskip
\noindent\textbf{MonteRET.}
MonteRET combines report generator with condition- and
region-aware retrieval, reranking, and LLM refinement~\cite{lin2026monteret}.
It is tested on 1,564 scans. The authors report RadBERT micro-F1 0.420; the 0.365
entry in Table~\ref{tab:ctrate_report_generation} is our unweighted mean of
the 18 per-abnormality F1 values printed to three decimals. It is  a
derived approximation, and neither the scan-to-reconstruction mapping nor the
retrieval database available to each test case is specified.

\medskip
\noindent\textbf{Astra.}
Astra connects a Merlin visual encoder through a Perceiver to Qwen2.5-VL and
is trained using supervised fine-tuning followed by
FORTE-guided reinforcement learning~\cite{wang2026astra}. Its CT-RATE
evaluation uses 1,564 acquisition-level cases and harmonized reference
findings, rather than the native reports paired independently with every
released reconstruction. The authors report RadBERT micro-F1 0.5502; the
approximately 0.484 macro-F1 in our table is an unweighted mean digitized from
the 18 label-wise bars in their Fig.~7e. Both digitization uncertainty and the
harmonized references preclude treating this as an exact, protocol-matched
macro-F1.

\medskip
\noindent\textbf{VoxelFM.}
VoxelFM is a 3D ViT pretrained with DINO-style self-supervision~\cite{morenoaguado2026voxelfm}. For report generation, the encoder is
frozen and a Q-Former interface to Qwen3-8B is trained on CT-RATE with
low-rank language-model adaptation. The authors reformat reference reports
with GPT-OSS-120B and use the same model to extract the 18 findings from
generated reports. They evaluate the full historical public validation
cohort, reported as 3,039 reconstructions.

\medskip
\noindent\textbf{CT-SSG.}
CT-SSG is a 2.5D system that combines an ImageNet-initialized ResNet-18 over
adjacent slices with graph-based spectral aggregation~\cite{dipiazza2026ctssg}.
For reporting, it freezes the classification-pretrained visual encoder and
trains a lightweight CT2Rep-style autoregressive decoder with next-token
supervision. Its results are reported as a five-fold CT-RATE evaluation with
RadBERT label extraction. The paper states the evaluation is on the CT-RATE test cohort, but the number of report-generation test examples is not specified.

\medskip
\noindent\textbf{ClinFusion.}
ClinFusion augments Qwen3-VL with Qwen-ViT, DINOv2-Large, ConvNeXt-Large, and a
native 3D ViT fused through a cascade spatial-aware locality
module~\cite{yuan2026clinfusion}. The 8B and 32B variants are trained through
five alignment and instruction-tuning stages. Both variants are evaluated on 3,039 historical
CT-RATE cases, but generation is conditioned on clinical indication and
anatomical regions extracted from the reference report by GPT-4.1. GPT-4.1
then labels generated claims as matched, missed, or hallucinated, giving
claim-level F1 of 0.216 and 0.239.

\paragraph{Notable method not evaluated.}
Merlin is a 3D vision--language foundation model pretrained on abdominal and
abdominopelvic CT examinations and their corresponding radiology
reports~\cite{blankemeier2026merlin}. Its adapted report-generation module was
likewise trained for section-by-section abdominal reporting and evaluated on
the authors' internal abdominal CT test set, not on chest CT. Although
the released generator includes a lower-thorax section, reflecting the lung
bases visible on abdominal examinations, it does not provide a native complete
chest-report protocol. We therefore exclude the official Merlin checkpoint
from Table~\ref{tab:ctrate_report_generation}.

\subsection{Merlin dataset evaluation}
\label{sec:results_merlin_dataset}

The Merlin Abdominal CT Dataset comprises 25,494 abdominal and abdominopelvic
CT scan--report pairs from 18,317 patients treated at Stanford Hospital between
2012 and 2018~\cite{blankemeier2026merlin}. Each case corresponds to one CT
examination, for which the DICOM series with the largest number of slices was
retained and converted into a 3D NIfTI volume. The original
study describes more than six million constituent two-dimensional images in
its 15,331-scan training cohort. It reports patient-disjoint internal splits of
15,331 training, 5,060 validation, and 5,137 test scans. The subsequently
released metadata differs slightly, containing 15,314 training, 5,055
validation, and 5,125 test scans. We use the released test split throughout.
For abnormality classification, 5,082 of the 5,125 test scans have released
30-finding labels; the remaining 43 scans have reports but no row in the
classification-label file. For report generation, we use the paired
reports for all 5,125 released test scans. Classification uses our general
checkpoint without Merlin-specific fine-tuning, whereas report generation uses
the adapted checkpoint described in Section~\ref{sec:results_merlin_reports}.

\subsubsection{Merlin abnormality classification}
\label{sec:results_merlin_classification}

We evaluate abnormality classification on the held-out Merlin abdominal CT
cohort~\cite{blankemeier2026merlin}. This benchmark complements CT-RATE by
testing the model on a different anatomical domain and label ontology. The
released labels cover 30 findings and are report-derived and tri-state:
positive, explicitly negative, or unmentioned. Following the original
zero-shot protocol, unmentioned findings are excluded and positive and negative
examples are balanced independently for each finding before computing
positive-class F1. Table~\ref{tab:merlin_classification} presents preliminary
results. The comparison is not strictly paired because the original paper used
an internal 5,137-CT test split and did not release its exact balanced case IDs,
whereas our evaluation reconstructs balanced cohorts from the released test
labels. In addition, our reported prompt was selected after prompt exploration
on this same released test cohort and should therefore be regarded as a
test-prompt-tuned estimate. For our probability-based evaluation, thinking is
disabled and the normalized next-token probability of \textit{Yes} is retained
as the continuous score. F1 uses a fixed threshold of 0.5, without threshold
tuning, whereas AUROC is threshold-free. For six lower-thoracic findings---
aortic valve calcification, coronary calcification, cardiomegaly, pleural
effusion, atelectasis, and hiatal hernia---we used the chest anatomical region;
the remaining 24 findings used the abdomen region. The same prompt with
thinking enabled and generated hard Yes/No answers yielded macro-F1 0.7531.
Per-finding results for the probability-based evaluation are reported in
Appendix~\ref{app:merlin_classification_table}.

\begin{table}[!htbp]
\centering
\small
\caption{Merlin 30-finding abnormality-classification results.
Macro-F1 is the unweighted mean of the positive-class F1 scores across the 30
findings after balancing explicit positive and negative examples separately for
each finding. Merlin produces hard decisions and does not report AUROC for this
task. For NV-REASON-CT, F1 is computed from next-token probabilities at a fixed
0.5 threshold and AUROC from the corresponding continuous scores. All scores
are reported on a $[0,1]$ scale.}
\label{tab:merlin_classification}
\renewcommand{\arraystretch}{1.12}
\setlength{\tabcolsep}{5pt}
\begin{adjustbox}{max width=\textwidth}
\begin{tabular}{@{}p{3.0cm}p{8.4cm}cc@{}}
\toprule
\textbf{Method} & \textbf{Evaluation details} &
\shortstack{\textbf{Macro}\\\textbf{F1} $\uparrow$} &
\shortstack{\textbf{Macro}\\\textbf{AUROC} $\uparrow$} \\
\midrule
Merlin~\cite{blankemeier2026merlin} &
Authors' internal 5,137-CT test split; per-finding balanced cohorts; hard
comparison of mean image--text cosine similarity across radiologist-developed
positive and negative prompt banks. &
0.7410 & -- \\
VoxelFM~\cite{morenoaguado2026voxelfm} &
VoxelFM-study evaluation on 5,082 labeled studies from the public Merlin test
split; all 30 findings, with unmentioned labels treated as negative; frozen
VoxelFM encoder with a supervised Q-Former probe trained on the Merlin training
split and selected by validation AUROC. &
-- & 0.7970 \\
\textbf{NV-REASON-CT (ours)} &
Released test labels; balanced cohort of 11,348 study--finding
pairs from 4,670 CTs across 30 findings; one uniform prompt; normalized
next-token Yes/No probabilities with thinking disabled; fixed 0.5 F1 threshold
and no threshold tuning; chest anatomical region for six lower-thoracic
findings and abdomen region for the remaining 24. &
\textbf{0.7655} & \textbf{0.8329} \\
\bottomrule
\end{tabular}
\end{adjustbox}
\end{table}

\subsubsection{Merlin report generation}
\label{sec:results_merlin_reports}

We evaluate abdominal CT report generation on all 5,125 scans in the released
Merlin test split using the Findings section. We
compute BLEU, ROUGE-L, BERTScore, RadGraph-F1, and GREEN using the RadEval
framework~\cite{xu2025radeval}. BLEU measures lexical
$n$-gram precision with a brevity penalty, ROUGE-L measures longest common
subsequence overlap, and BERTScore measures contextual token-level semantic
similarity. RadGraph-F1 first maps each candidate and reference report to a
graph of clinical entities and their relations, then measures graph overlap
with an F1 score~\cite{radgraph1,radgraph2}. We report RadEval's stricter
complete-match RadGraph-F1. GREEN uses a learned evaluator to identify
clinically significant errors relative to the reference
report~\cite{ostmeier2024green}. We follow the evaluation setup used by
Jolia~\cite{khlaut2026jolia}.

The released Merlin reports use a specific structure comprising short,
organ-specific subsections~\cite{blankemeier2026merlin}.
By contrast, our native NV-REASON-CT reporting format is different: it is longer with different subsections and organizes
Findings and Impression differently. Lexical and semantic similarity metrics are sensitive to report
length, wording, section order, and headings; evaluating the native output
directly would therefore conflate clinical content with output-format mismatch.
To reduce this mismatch, we performed supervised in-domain output-format
adaptation starting from our general checkpoint. We fine-tuned on the
15,314 released Merlin training rows for one epoch, without mixing other
datasets. A dedicated prompt
requested only the Findings section and specified the fixed order of the 15
Merlin anatomical headings. The same prompt was used to generate Findings for
all 5,125 test scans; no test report was used for parameter updates.

Table~\ref{tab:merlin_report_generation} compares our adapted model with the
author-reported results compiled by Jolia~\cite{khlaut2026jolia}. The nominal
cohort and report scope are matched, although exact evaluator versions and
reference preprocessing may differ. 

\begin{table}[!htbp]
\centering
\scriptsize
\caption{Report generation evaluation on the Merlin dataset test split
($N=5{,}125$).}
\label{tab:merlin_report_generation}
\renewcommand{\arraystretch}{1.13}
\setlength{\tabcolsep}{10.0pt}
\begin{adjustbox}{max width=\textwidth}
\begin{tabular}{@{}lccccc@{}}
\toprule
\textbf{Method} &
\textbf{BLEU} $\uparrow$ &
\textbf{ROUGE-L} $\uparrow$ &
\textbf{BERTScore} $\uparrow$ &
\textbf{RadGraph-F1}$\uparrow$ &
\textbf{GREEN} $\uparrow$ \\
\midrule
Med3DVLM                  & 0.004 & 0.079 & 0.256 & 0.010 & 0.007 \\
MedGemma 1.5              & \textbf{0.124} & 0.106 & 0.361 & 0.032 & 0.038 \\
Merlin                    & 0.063 & 0.285 & 0.551 & 0.237 & 0.316 \\
Jolia                     & 0.119 & 0.323 & 0.567 & 0.317 & 0.324 \\
\textbf{NV-REASON-CT (ours)} & 0.0769 & \textbf{0.3360} & \textbf{0.5788} &
\textbf{0.3216} & \textbf{0.3658} \\
\bottomrule
\end{tabular}
\end{adjustbox}
\end{table}

\FloatBarrier

Following adaptation to Merlin's Findings-section format, NV-REASON-CT achieves
the highest reported ROUGE-L, BERTScore, RadGraph-F1, and GREEN scores among the
models in Table~\ref{tab:merlin_report_generation}. Its BLEU score is lower than
those of MedGemma 1.5 and Jolia, showing that the relative ranking depends on
the evaluation metric.

\subsection{RAD-ChestCT dataset evaluation}
\label{sec:results_radchestct_dataset}

RAD-ChestCT~\cite{draelos2021radchestct,radchestct2022release} is an external, out-of-distribution chest CT benchmark that we use only for evaluation. The public image release
contains 3,630 volumes. Each publicly released volume is accompanied by an
$84\times52$ abnormality-by-location label matrix extracted from its source
radiology report.  Following prior work, we evaluate a subset of 16 harmonized abnormality labels
for consistency of comparisons~\cite{hamamci2026generalist, buess2026alo}.
This mapping starts from the 18-label CT-RATE abnormality ontology, excludes
Mosaic attenuation, and merges the Arterial wall calcification and Coronary
artery wall calcification outputs into a single RAD-ChestCT Calcification target. The remaining 15 abnormalities are matched directly, yielding 16 scan-level targets in total.  Some studies use alternative 16-label mappings, which we explicitly identify below.
Table~\ref{tab:radchestct_classification_published} summarizes published direct
classification results.

\subsubsection{RAD-ChestCT abnormality classification}
\label{sec:results_radchestct_classification}

\begin{table*}[!htbp]
\centering
\scriptsize
\caption{Published RAD-ChestCT abnormality-classification results. Macro F1 is the unweighted mean of the per-abnormality
positive-class F1 scores, whereas weighted F1 averages the positive- and
negative-class F1 within each abnormality according to class support before
averaging across abnormalities.  }
\label{tab:radchestct_classification_published}
\renewcommand{\arraystretch}{1.12}
\setlength{\tabcolsep}{3.5pt}
\begin{adjustbox}{max width=\textwidth}
\begin{tabular}{@{}>{\raggedright\arraybackslash}p{2.6cm}>{\raggedright\arraybackslash}p{8.2cm}rrrrr@{}}
\toprule
\textbf{Method} & \textbf{Evaluation details} &
\textbf{Macro F1} $\uparrow$ &
\textbf{Weighted F1} $\uparrow$ &
\textbf{AUROC} $\uparrow$ &
\textbf{Precision} $\uparrow$ & \textbf{Recall} $\uparrow$ \\
\midrule
CT-CLIP~\cite{hamamci2026generalist} &
3,630 volumes; 16 classes. VocabFine version. macro-F1 was approximated from the authors' normalized per-class confusion matrices. &
0.4087 & 0.677 & 0.650 & 0.346 & 0.5996 \\
Pillar-0~\cite{agrawal2025pillar0} &
3,630 volumes; 16 classes. DCP-PD~\cite{wang2026dcppd} evaluation over frozen Pillar-0 features. Weighted binary F1 was recomputed from the authors' per-class precision and recall values. &
0.497 & 0.6942 & 0.796 & 0.399 & 0.811 \\
CT-SSG~\cite{dipiazza2026ctssg} &
1,334 patients (but not stated if all 3,630 volumes used); 16 classes. Five-fold mean. &
0.5225 & -- & 0.7458 & -- & -- \\
fVLM~\cite{shui2025fvlm} &
3,630 volumes; 16 classes (alternative definition). Zero-shot, organ-conditioned paired-prompt inference.  &
-- & 0.688 & 0.680 & 0.374 & 0.646 \\
Merlin~\cite{blankemeier2026merlin} &
3,630 volumes; 16 classes (alternative definition).  fVLM-paper evaluation~\cite{shui2025fvlm} under the organ-mapped set. &
-- & 0.663 & 0.644 & 0.348 & 0.610 \\

COLIPRI-CRM\cite{wald2026colipri} &
3,630 volumes; 16 classes (mapping not stated);  best of five pooling schemes. &
-- & -- & 0.7657 & -- & -- \\

\textbf{NV-REASON-CT (ours)} &
3,630 volumes; 16 classes . Direct next-token Yes/No evaluation with a fixed $P(\mathrm{Yes})\geq0.5$ rule, no re-training, no calibration.  &
0.5311 & 0.7379 & 0.7958 & 0.4503 & 0.7647 \\
\bottomrule
\end{tabular}
\end{adjustbox}
\vspace{3pt}
\end{table*}
\FloatBarrier

\medskip
\noindent\textbf{NV-REASON-CT.}
We evaluated all 3,630 public RAD-ChestCT volumes using the 16-label mapping
described above, a single binary prompt, and a fixed 0.5 decision threshold,
with no RAD-ChestCT-specific prompt or threshold tuning. Direct classification
achieved positive-class macro-F1 0.5311, weighted binary F1 0.7379, and
macro-AUROC 0.7958.

\medskip
\noindent\textbf{CT-CLIP.}
CT-CLIP was contrastively pretrained on the same CT-RATE training split of
47,149 reconstructed volumes and their reports. 
It was evaluated on 3,630 RAD-ChestCT volumes using the 16-label harmonization
described above.

\medskip
\noindent\textbf{fVLM.}
fVLM learns organ-conditioned image--text representations using anatomical
masks and one learned query per organ~\cite{shui2025fvlm}. 
The authors evaluate the whole 3,630-volume
public RAD-ChestCT release with paired positive/negative prompts. Its released
evaluator implements an alternative 16-label mapping and requires
TotalSegmentator-derived masks.

\medskip
\noindent\textbf{Merlin.}
The official Merlin encoder was pretrained on
15,331 abdominal CT scans with reports and structured clinical data
~\cite{blankemeier2026merlin}. The fVLM paper~\cite{shui2025fvlm} evaluates Merlin on all 3,630
RAD-ChestCT volumes using this alternative 16-label mapping; we report these
results in Table~\ref{tab:radchestct_classification_published}.

\medskip
\noindent\textbf{COLIPRI.}
COLIPRI~\cite{wald2026colipri} combines image--text contrastive learning, report generation,
opposite-sentence learning, and image-only self-supervision. 
The authors' frozen
probe is trained and selected only with CT-RATE data, sweeping five pooling
schemes and four learning rates.
It uses all 3,630 RAD-ChestCT volumes; the 16-class mapping is not explicitly stated.

\medskip
\noindent\textbf{CT-SSG.}
CT-SSG is a supervised 2.5D ResNet-18 and spectral-graph classifier trained in
five folds on CT-RATE~\cite{dipiazza2026ctssg}. Evaluation uses the 16-label harmonization described above.
The source reports 1,334 RAD-ChestCT patients but not the exact number of evaluated volumes; all
values are five-fold means.

\medskip
\noindent\textbf{Pillar-0.}
Pillar-0's Atlas chest encoder was contrastively pretrained on 86,411 private
chest CT scans~\cite{agrawal2025pillar0}. The later DCP-PD study freezes Atlas,
pools its multiscale representation, and trains the DCP-1 linear multilabel
classifier on CT-RATE~\cite{wang2026dcppd}. DCP-1 is evaluated on all
3,630 RAD-ChestCT volumes using this 16-label harmonization.

\subsubsection{RAD-ChestCT report generation}
\label{sec:results_radchestct_reports}

The public RAD-ChestCT release does not include paired reference report text.
Consequently, generated reports cannot be evaluated with reference-dependent
text or semantic metrics such as BLEU, GREEN, RadGraph F1, or RadFact-CT F1.
They can nevertheless be evaluated for clinical content by extracting the
harmonized abnormalities from each generated report and comparing them with
the released binary labels. Table~\ref{tab:radchestct_report_generation}
separates this report-derived clinical evaluation from direct image
classification. 

\begin{table*}[!htbp]
\centering
\scriptsize
\caption{Published RAD-ChestCT report-derived evaluation. Generated
reports are converted to abnormality predictions and compared with the public
binary labels. Macro F1, precision, and recall are unweighted
means of the corresponding positive-class per-label metrics. Because reference reports are unavailable,
BLEU, ROUGE, GREEN, and other reference-dependent report metrics cannot be
computed. }
\label{tab:radchestct_report_generation}
\renewcommand{\arraystretch}{1.12}
\setlength{\tabcolsep}{3pt}
\begin{adjustbox}{max width=\textwidth}
\begin{tabular}{@{}>{\raggedright\arraybackslash}p{2.6cm}>{\raggedright\arraybackslash}p{9.0cm}rrr@{}}
\toprule
\textbf{Method} & \textbf{Evaluation details} &
\textbf{Macro F1} $\uparrow$ &
\textbf{Precision} $\uparrow$ &
\textbf{Recall} $\uparrow$ \\
\midrule
CT-CHAT~\cite{hamamci2026generalist} & &
0.182 & 0.382 & 0.171 \\
BTB3D~\cite{hamamci2025btb3d} &
BTB3D-16 variant. &
0.266 & 0.272 & 0.329 \\
RadFM~\cite{wu2023radfm} &
CT-CHAT~\cite{hamamci2026generalist} evaluation using the official pretrained checkpoint. &
0.069 & 0.283 & 0.044 \\
M3D-LaMed~\cite{bai2024m3dlamed,bai2026exact} &
EXACT evaluation~\cite{bai2026exact} using official weights; modified 16 labels harmonization. &
0.113 & 0.269 & 0.080 \\
DCP-PD~\cite{wang2026dcppd} &
DCP-PD base VLM. &
0.395 & 0.443 & 0.418 \\
EXACT-CHAT~\cite{bai2026exact} &
modified 16 label harmonization &
0.289 & 0.469 & 0.298 \\
\multirow{2}{2.6cm}{\raggedright\textbf{NV-REASON-CT (ours)}} &
Default structured-report prompt; no RAD-ChestCT tuning. Findings and Impression were classified with the frozen official CT-RATE RadBERT at 0.5; 288 reports exceeded its 512-token limit and were truncated. &
0.4854 & 0.5101 & 0.4943 \\
& Default structured-report prompt, but the labels were parsed from the answer block directly (without RadBERT) &
0.5105 & 0.5158 & 0.5524 \\
\bottomrule
\end{tabular}
\end{adjustbox}
\end{table*}
\FloatBarrier

\medskip
\noindent\textbf{NV-REASON-CT.}
We generate reports for all 3,630 public volumes with the
default structured-report prompt and without RAD-ChestCT-specific tuning. A
frozen CT-RATE RadBERT applied to Findings and Impression yields macro F1
0.4854, precision 0.5101, and recall 0.4943; 288 reports exceed its 512-token
limit and are truncated. Parsing the final abnormality list directly
from the same generated reports yields macro F1 0.5105, precision 0.5158, and recall
0.5524. See Table~\ref{tab:radchestct_report_generation}.

\medskip
\noindent\textbf{CT-CHAT.}
CT-CHAT combines the CT-CLIP volumetric encoder with Llama~3.1 through a
learned multimodal projector; the projector is first aligned to CT-RATE report
text and is subsequently optimized together with low-rank language-model
adaptation on CT-RATE instruction data~\cite{hamamci2026generalist}. The
reported RAD-ChestCT experiment uses the Llama~3.1-70B configuration on all
3,630 public volumes. Abnormalities extracted from the generated reports are
compared with the 16 harmonized binary targets, yielding macro F1 0.182,
precision 0.382, and recall 0.171.

\medskip
\noindent\textbf{BTB3D.}
BTB3D learns discrete volumetric tokens with a causal 3D convolutional
encoder--decoder, Haar-wavelet compression, and lookup-free quantization, and
projects these tokens to a Llama~3.1-8B report decoder~\cite{hamamci2025btb3d}.
We report its best $16\times16\times8$ tokenization variant under the
CT-CHAT-16 evaluation protocol. The paper presents RAD-ChestCT as the external
test set but does not state the exact evaluated number of volumes; it reports
macro F1 0.266, precision 0.272, and recall 0.329.

\medskip
\noindent\textbf{RadFM.}
RadFM is a generalist medical VLM comprising a 3D ViT, a Perceiver-style token
resampler, and a MedLLaMA-13B language model~\cite{wu2023radfm}. The
RAD-ChestCT result is not a native RadFM evaluation: the CT-CHAT study applies
the official pretrained checkpoint without retraining to all 3,630 public
volumes and evaluates its generated reports with the CT-CHAT-16 label
protocol~\cite{hamamci2026generalist}. This later evaluation gives macro F1
0.069, precision 0.283, and recall 0.044.

\medskip
\noindent\textbf{M3D-LaMed.}
M3D-LaMed pairs a native 3D ViT with a Llama-family language model and is
trained by image--text contrastive alignment followed by multimodal
instruction tuning on M3D-Data~\cite{bai2024m3dlamed}. Its original
publication does not evaluate RAD-ChestCT. The values in Table~\ref{tab:radchestct_report_generation}
instead come from the later EXACT study, which runs the official weights on
all 3,630 volumes and uses its modified 16-label harmonization and RadBERT
report labeling~\cite{bai2026exact}. It reports macro F1 0.113 (95\% CI,
0.091--0.137), precision 0.269, and recall 0.080.

\medskip
\noindent\textbf{DCP-PD base VLM.}
DCP-PD builds its report generator from the frozen Pillar-0 Atlas visual
backbone, an attention-pooling projector, and Llama~3-Instruct-8B, and trains
the projector and language-model adapters on structured CT-RATE
reports~\cite{wang2026dcppd}. 
On all 3,630 RAD-ChestCT volumes under the
16-label protocol, this base model achieves macro F1 0.395, precision 0.443,
and recall 0.418.

\medskip
\noindent\textbf{EXACT-CHAT.}
EXACT learns 18 voxel-level anomaly-aware maps with a Y-Mamba image backbone;
EXACT-CHAT connects the frozen encoder to Llama~3.1-8B-Instruct through an
attentional pooling projector and also supplies its map-derived diagnostic
predictions as text~\cite{bai2026exact}. It is trained on CT-RATE and evaluated
without RAD-ChestCT adaptation on all 3,630 external volumes. Under the
modified 16-label protocol, the unrefined model listed here achieves RadBERT
macro F1 0.289, precision 0.469, and recall 0.298.

\FloatBarrier

\subsection{Internal Dataset Evaluation}
\label{sec:results_internal_dataset}

\begin{figure}[t]
\centering
\begin{minipage}[t]{0.49\textwidth}
\centering
\includegraphics[width=\linewidth]{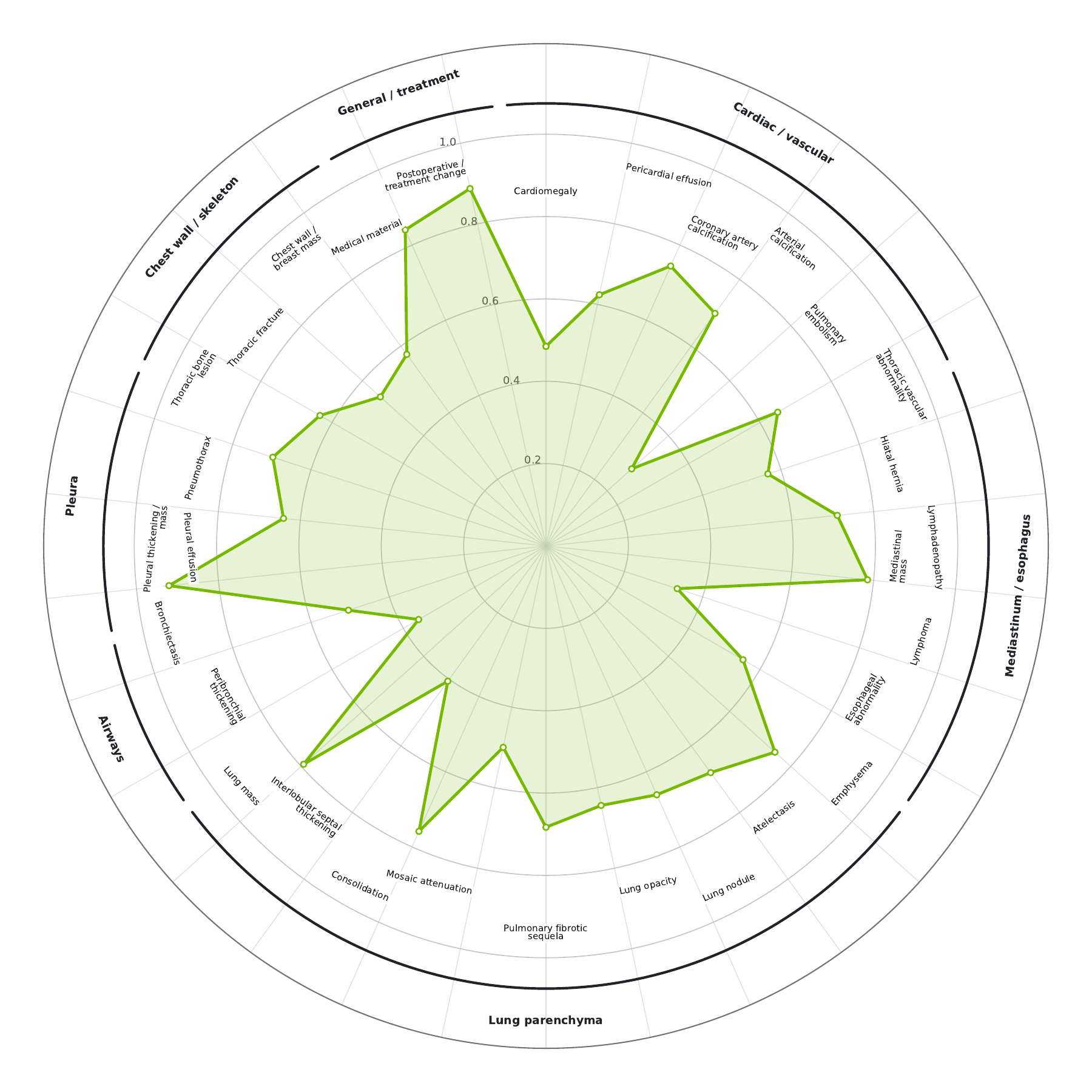}
\textbf{(a) Chest}
\end{minipage}\hfill
\begin{minipage}[t]{0.49\textwidth}
\centering
\includegraphics[width=\linewidth]{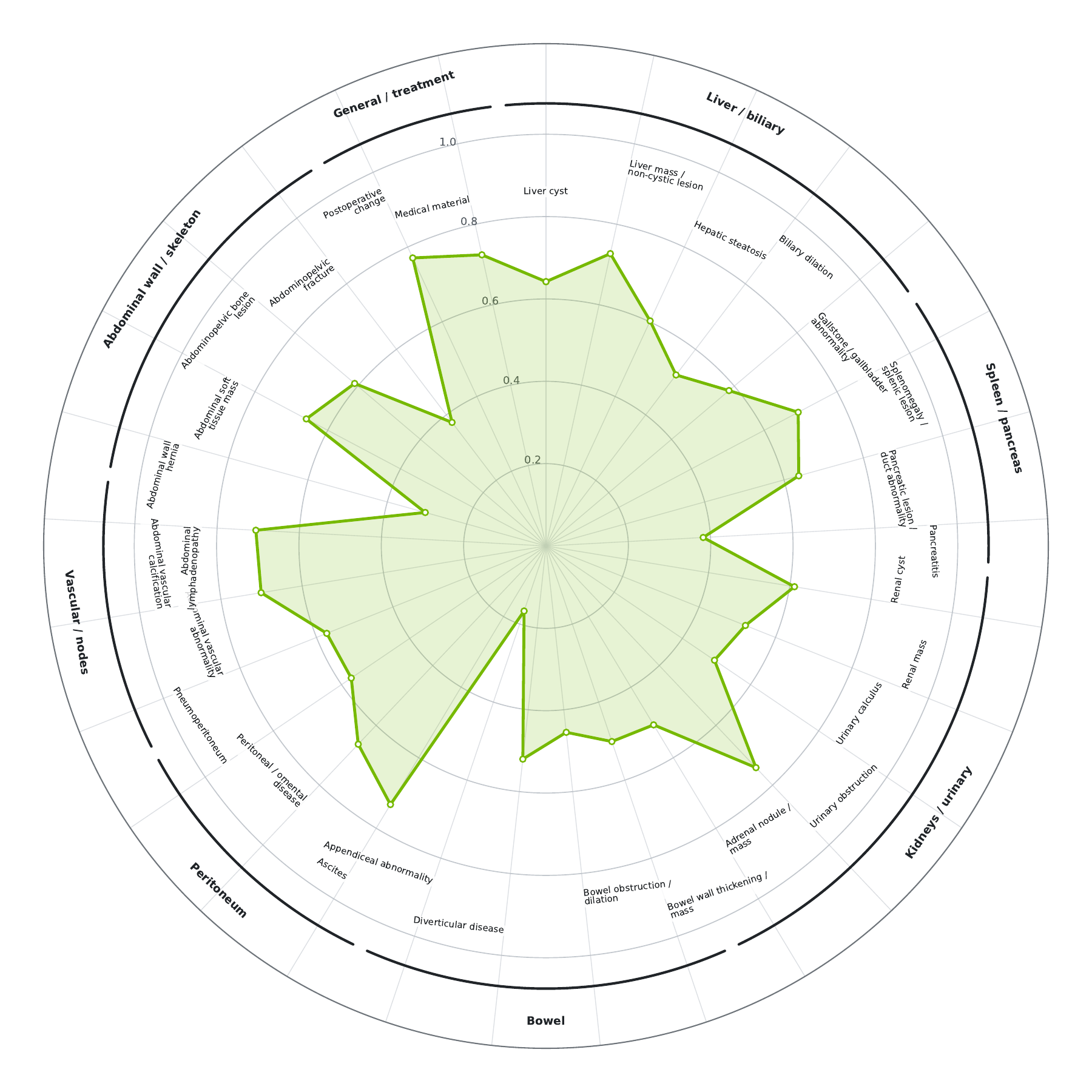}
\textbf{(b) Abdomen}
\end{minipage}
\caption{Per-abnormality positive-class F1 on the balanced NIH validation set cohorts: (a) 30 chest findings and (b) 29 abdominal findings. Findings
are grouped by anatomical system, and each spoke reports the F1 value from the
independently balanced positive--negative cohort used in
Table~\ref{tab:internal_dataset_classification}.}
\label{fig:nih_regional_f1}
\end{figure}

We additionally evaluate NV-REASON-CT on a held-out internal NIH collection. The dataset
contains 5,092 cases from a unique
patients. Evaluation is coverage-aware: 3,981 cases contain the full chest and
4,205 contain the full abdomen, with 3,132 cases contributing to both regional
cohorts. Reports were labeled for 30 chest and 29 abdominal findings. We balance positive vs negative findings via subsampling. This produced 35,992 balanced chest case--finding examples and 37,740
balanced abdomen case--finding examples. The class-specific cohorts vary
substantially in size: for example, the balanced evaluation includes only 22
examples for appendiceal abnormality, 56 for pneumothorax, 58 for
pneumoperitoneum, and 86 for pulmonary embolism. Macro-F1 is the unweighted
mean of the positive-class F1 scores over the findings in each region.
Table~\ref{tab:internal_dataset_classification} reports the per-abnormality and
macro results, while Figure~\ref{fig:nih_regional_f1} visualizes the
corresponding balanced F1 profiles by anatomical system. For AUROC, thinking
was disabled and the normalized next-token
probability of \textit{Yes} was retained as a continuous score. The same balanced selections were used for both metrics; no threshold tuning was performed.

\begin{table}[H]
\centering
\scriptsize
\caption{Per-abnormality NV-REASON-CT classification results on the internal
NIH validation cohort. Positive and negative cases are balanced
independently for each finding. Macro denotes
the unweighted mean over findings in each anatomical cohort. Pos. gives the
number of positive cases retained after balancing.}
\label{tab:internal_dataset_classification}
\renewcommand{\arraystretch}{1.04}
\setlength{\tabcolsep}{3pt}
\begin{adjustbox}{max width=\textwidth}
\begin{tabular}{@{}lccc@{\hspace{6pt}}lccc@{}}
\toprule
\multicolumn{4}{c}{\textbf{Chest cohort (30 findings)}} &
\multicolumn{4}{c}{\textbf{Abdomen cohort (29 findings)}} \\
\cmidrule(lr){1-4}\cmidrule(lr){5-8}
\textbf{Finding} & \textbf{Pos.} & \textbf{F1} $\uparrow$ & \textbf{AUROC} $\uparrow$ &
\textbf{Finding} & \textbf{Pos.} & \textbf{F1} $\uparrow$ & \textbf{AUROC} $\uparrow$ \\
\midrule
\textbf{Macro} & -- & \textbf{0.6302} & \textbf{0.8113} &
\textbf{Macro} & -- & \textbf{0.5759} & \textbf{0.7615} \\
\midrule
Medical material & 1847 & 0.8405 & 0.8831 & Liver cyst & 398 & 0.6420 & 0.7657 \\
Arterial wall calcification & 911 & 0.6984 & 0.8353 & Liver mass or non-cystic lesion & 1853 & 0.7271 & 0.7963 \\
Cardiomegaly & 339 & 0.4848 & 0.7586 & Hepatic steatosis & 342 & 0.6024 & 0.8058 \\
Pericardial effusion & 343 & 0.6238 & 0.8099 & Biliary dilation & 367 & 0.5219 & 0.7644 \\
Coronary artery wall calcification & 827 & 0.7443 & 0.8587 & Gallstone or gallbladder abnormality & 519 & 0.5831 & 0.7578 \\
Hiatal hernia & 243 & 0.5666 & 0.7819 & Splenomegaly or splenic lesion & 755 & 0.6933 & 0.7976 \\
Lymphadenopathy & 1735 & 0.7114 & 0.7690 & Pancreatic lesion or duct abnormality & 454 & 0.6371 & 0.7570 \\
Emphysema & 169 & 0.7483 & 0.8643 & Pancreatitis & 52 & 0.3824 & 0.8023 \\
Atelectasis & 1202 & 0.6804 & 0.7758 & Renal cyst & 1134 & 0.6115 & 0.7474 \\
Lung nodule & 1648 & 0.6614 & 0.7136 & Renal mass & 743 & 0.5214 & 0.6590 \\
Lung opacity & 655 & 0.6442 & 0.7978 & Urinary calculus & 384 & 0.4944 & 0.7111 \\
Pulmonary fibrotic sequela & 1096 & 0.6831 & 0.7872 & Urinary obstruction & 211 & 0.7416 & 0.8716 \\
Pleural effusion & 481 & 0.9207 & 0.9656 & Adrenal nodule or mass & 940 & 0.5070 & 0.6587 \\
Mosaic attenuation pattern & 89 & 0.5000 & 0.8514 & Bowel wall thickening or mass & 534 & 0.5012 & 0.6433 \\
Peribronchial thickening & 178 & 0.3571 & 0.7656 & Bowel obstruction or dilation & 107 & 0.4552 & 0.8080 \\
Consolidation & 257 & 0.7586 & 0.9040 & Diverticular disease & 554 & 0.5207 & 0.6993 \\
Bronchiectasis & 383 & 0.5046 & 0.7473 & Appendiceal abnormality & 11 & 0.1667 & 0.6322 \\
Interlobular septal thickening & 107 & 0.4058 & 0.8207 & Ascites & 670 & 0.7328 & 0.8637 \\
Lung mass & 333 & 0.7923 & 0.8867 & Peritoneal/omental disease & 467 & 0.6631 & 0.8279 \\
Pleural thickening or mass & 544 & 0.6411 & 0.7708 & Pneumoperitoneum & 29 & 0.5714 & 0.8966 \\
Mediastinal mass & 188 & 0.7853 & 0.8891 & Abdominal vascular abnormality & 510 & 0.5732 & 0.7266 \\
Lymphoma & 458 & 0.3351 & 0.7190 & Abdominal vascular calcification & 1272 & 0.7011 & 0.8193 \\
Pneumothorax & 28 & 0.6977 & 0.9700 & Abdominal lymphadenopathy & 1365 & 0.7057 & 0.7823 \\
Thoracic bone lesion & 531 & 0.6339 & 0.7399 & Abdominal wall hernia & 379 & 0.3044 & 0.6411 \\
Thoracic fracture & 208 & 0.5413 & 0.8005 & Abdominal soft tissue mass & 941 & 0.6590 & 0.7552 \\
Chest wall or breast mass & 404 & 0.5752 & 0.7673 & Abdominopelvic bone lesion & 892 & 0.6098 & 0.7032 \\
Pulmonary embolism & 43 & 0.2800 & 0.6723 & Abdominopelvic fracture & 190 & 0.3770 & 0.7305 \\
Thoracic vascular abnormality & 742 & 0.6499 & 0.7614 & Abdominal postoperative change & 1913 & 0.7709 & 0.8361 \\
Esophageal abnormality & 387 & 0.5524 & 0.7471 & Abdominal medical material & 884 & 0.7241 & 0.8233 \\
Postoperative or treatment-related change & 1620 & 0.8874 & 0.9250 & & & & \\
\bottomrule
\end{tabular}
\end{adjustbox}
\end{table}

Performance on the balanced NIH cohorts varied substantially across findings.
In the chest, the highest F1 scores were observed for pleural effusion,
postoperative or treatment-related change, and medical material. In the abdomen,
postoperative change, urinary obstruction, ascites, and liver masses or
non-cystic lesions yielded the highest F1 scores. Lower F1 scores were observed
for pulmonary embolism and lymphoma in the chest, and for appendiceal
abnormality and abdominal wall hernia in the abdomen. These findings highlight
variation within each anatomical region; estimates for findings with small
evaluation cohorts should be interpreted cautiously.

\FloatBarrier

\subsection{Subjective Evaluation with Expert Radiologists}

We conducted a preliminary within-subject reader study with US board-certified
radiologists to assess perceived output quality, usefulness, time savings, and
influence on clinical judgment.
For each case, readers first interpreted the CT without AI assistance, following
their usual clinical workflow. After a short distraction, they reviewed the
same case with the model's full reasoning text and structured report. Readers
could copy the structured report, verify its contents, and edit it as needed.

\paragraph{Participants and Materials.}
\begin{itemize}
    \item \textbf{Participants:} 2 radiologists with 10+ years of clinical practice.
    \item \textbf{Cases:} 10 cases comprising 2 normal and 8 abnormal examinations, with 5 chest and 5 abdominal CT cases.
    \item \textbf{Platform:} Web interface of the trained NV-REASON-CT model.
\end{itemize}

\paragraph{Questionnaire Items}

The ten agreement items were rated using a 5-point \textbf{Likert scale}: \emph{1 = Strongly disagree, 2 = Disagree, 3 = Neither agree nor disagree, 4 = Agree, 5 = Strongly agree}.

\paragraph{A. Accuracy and Reasoning Quality}
\begin{itemize}
    \item The reasoning trace includes correct statements about key observations.
    \item The reasoning does not include speculative or unsupported statements.
    \item The AI identified a relevant finding I initially missed.
    \item Uncertainty and limitations are explicitly stated when appropriate.
\end{itemize}

\paragraph{B. Time \& Efficiency}
\begin{itemize}
    \item The AI output reduced the time required to compose my report.
    \item The AI output helped me reach a decision faster.
    \item Approximate percentage of time saved on this case (percentage calculated)
\end{itemize}

\paragraph{C. Trust and Confidence}
\begin{itemize}
    \item After reading the AI output, I feel more confident (less uncertain) in my decision.
    \item If the AI had disagreed with me on this case, I would re-evaluate and re-examine the study.
    \item The AI’s suggestions improved my final decision.
    \item The AI did not bias me toward incorrect conclusions. 
\end{itemize}

\subsubsection{Reader-assessment results}

\definecolor{NVGreenColor}{RGB}{118,185,0} 
\definecolor{CadmiumColor}{RGB}{210,43,43}   

\begin{figure*}[t]
\begin{tikzpicture}
\begin{axis}[
  xbar,
  width=0.85\linewidth,
  height=0.35\textheight,          
  xmin=1, xmax=5.6,                         
  xlabel={Likert score (1--5)},
  ytick={1,2,3,4},
  y dir=reverse,                           
  yticklabels={
    {The reasoning trace includes correct statements about key observations.},
    {The reasoning does not include speculative or unsupported statements.},
    { The AI identified a relevant finding I initially missed.},
    {Uncertainty and limitations are explicitly stated when appropriate.},
  },
  yticklabel style={text width=.22\linewidth, align=right, font=\footnotesize},
  enlarge y limits=0.1,
  bar width=10pt,
  xmajorgrids=true,
  legend style={at={(0.5,1.02)}, anchor=south, legend columns=-1, font=\footnotesize},
   ytick pos=left, 
]

\addplot+[area legend, fill=NVGreenColor, error bars/.cd, x dir=both, x explicit] coordinates {
  (4.75,1) +- (0.55,0)
  (4.75,2) +- (0.55,0)
  (2.60,3) +- (1.23,0)
  (4.60,4) +- (0.82,0)

};
\legend{AI-assisted}
\end{axis}
\end{tikzpicture}%

\caption{The results (mean/std) of Accuracy and Reasoning Quality survey. Expert radiologists were tasked to write a report given the AI output. Likert scale: 1 - Strongly Disagree, 2 - Disagree, 3 - Neither Agree Nor Disagree, 4 - Agree, 5 - Strongly Agree. Ratings were high for correctness, avoidance of unsupported statements, and acknowledgment of uncertainty, and lower for identifying initially missed findings.}
\label{fig:results_survey_acc}
\end{figure*}
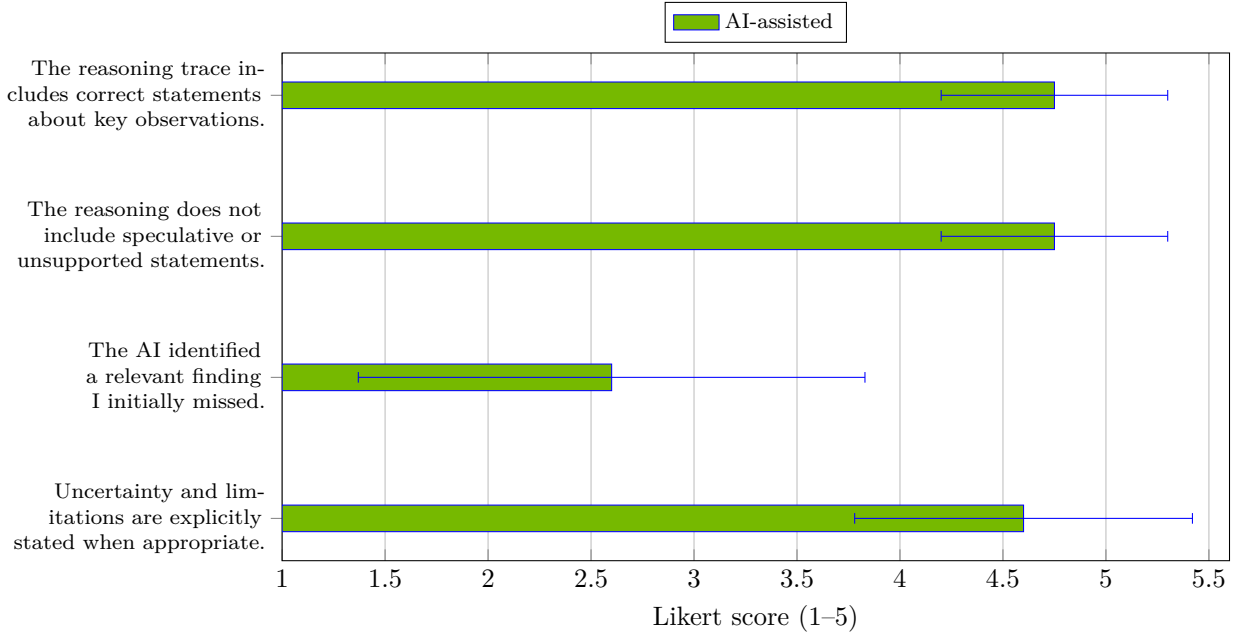

\paragraph{Accuracy and Reasoning Quality.}
Accuracy and Reasoning Quality ratings are shown in Figure~\ref{fig:results_survey_acc}.
\begin{itemize}
    \item \textbf{The reasoning trace includes correct statements about key observations}. Ratings were predominantly \emph{Agree} or \emph{Strongly Agree} ($4.75 \pm 0.55$).
    \item \textbf{The reasoning does not include speculative or unsupported statements}. Ratings were predominantly \emph{Agree} or \emph{Strongly Agree} ($4.75 \pm 0.55$).
    \item \textbf{The AI identified a relevant finding I initially missed}. Ratings were lower for this item ($2.60 \pm 1.23$). The radiologists attributed this to their careful initial review of each case, reporting that they did not miss any findings during their manual review. Whether this differs under routine clinical workloads of many daily cases remains to be evaluated.
    \item \textbf{Uncertainty and limitations are explicitly stated when appropriate}. Most responses were \emph{Agree} or \emph{Strongly Agree} ($4.60 \pm 0.82$).
\end{itemize}

\paragraph{Time \& Efficiency.}

\begin{figure*}[t]
\centering
\begin{tikzpicture}
\begin{axis}[
  ybar,
  width=0.55\linewidth,
  height=0.25\textheight,
  bar width=36pt,
  ylabel={Average time (min)},
  xmin=0.4, xmax=2.6,
  xtick={1,2},
  xticklabels={Without AI, With AI},
  ymin=0, ymax=30,
  ytick={0,5,10,15,20,25,30},
  ymajorgrids=true,
  nodes near coords,
  nodes near coords style={font=\small, text=black, /pgf/number format/fixed, /pgf/number format/precision=2, /pgf/number format/fixed zerofill},
  nodes near coords align={vertical},
]
\addplot+[fill=CadmiumColor, draw=none, bar shift=0pt] coordinates {(1,26.25)};
\addplot+[fill=NVGreenColor, draw=none, bar shift=0pt] coordinates {(2,13.125)};
\end{axis}
\end{tikzpicture}
\caption{Average reported time spent interpreting a CT and preparing a report without and with AI assistance.}
\label{fig:results_time}
\end{figure*}
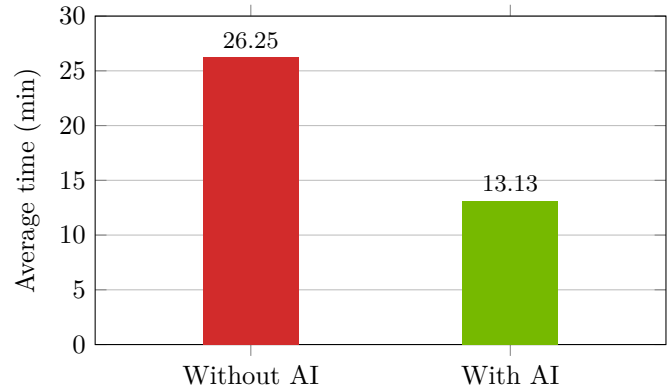

Figure~\ref{fig:results_time} compares the average reported time without and with AI assistance. Readers could copy and edit the AI-generated structured reports as needed. The mean time was 26.25\,min without AI and 13.13\,min with AI, corresponding to a 50\% reduction.

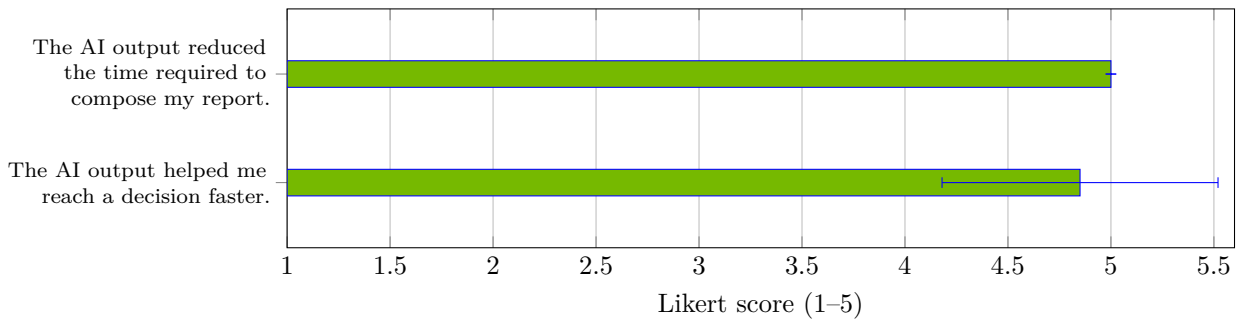
\begin{figure*}
\begin{tikzpicture}
\begin{axis}[
  xbar,
  width=0.85\linewidth,
  height=0.2\textheight,          
  xmin=1, xmax=5.6,                         
  xlabel={Likert score (1--5)},
  ytick={1,2},
  y dir=reverse,                           
  yticklabels={
    {The AI output reduced the time required to compose my report.},
    {The AI output helped me reach a decision faster.}
  },
  yticklabel style={text width=.22\linewidth, align=right, font=\footnotesize},
  enlarge y limits=0.6,
  bar width=10pt,
  xmajorgrids=true,
  legend style={at={(0.5,1.02)}, anchor=south, legend columns=-1, font=\footnotesize},
   ytick pos=left, 
]

\addplot+[area legend, fill=NVGreenColor, error bars/.cd, x dir=both, x explicit] coordinates {
  (5.00,1) +- (0.00,0)
  (4.85,2) +- (0.67,0)
};

\end{axis}
\end{tikzpicture}%

\caption{The Time \& Efficiency survey (mean/std). Likert scale: 1 - Strongly Disagree, 2 - Disagree, 3 - Neither Agree Nor Disagree, 4 - Agree, 5 - Strongly Agree. }
\label{fig:results_survey_time}
\end{figure*}

The Time \& Efficiency survey ratings are shown in Figure~\ref{fig:results_survey_time}.
\begin{itemize}
    \item \textbf{The AI output reduced the time required to compose my report.} All responses were \emph{Strongly Agree} ($5.00 \pm 0.00$).
    \item \textbf{The AI output helped me reach a decision faster.} Nineteen responses were \emph{Strongly Agree} and one was \emph{Disagree} ($4.85 \pm 0.67$).
\end{itemize}

\paragraph{Trust \& Confidence.}

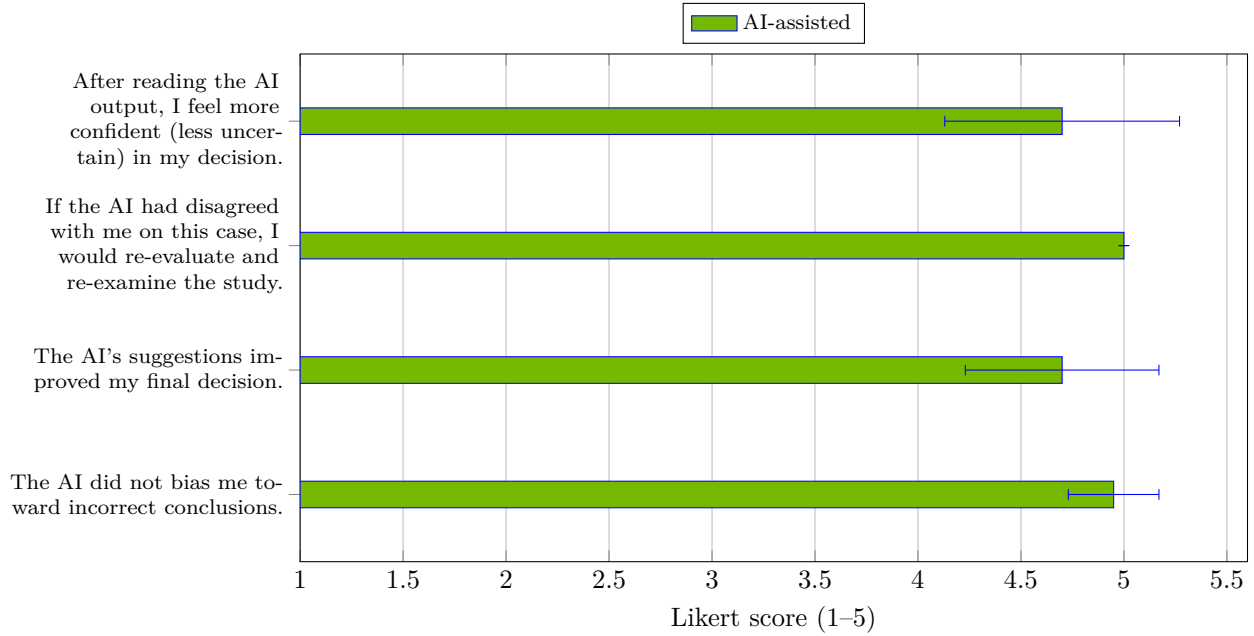
\begin{figure*}
\begin{tikzpicture}
\begin{axis}[
  xbar,
  width=0.85\linewidth,
  height=0.35\textheight,          
  xmin=1, xmax=5.6,                         
  xlabel={Likert score (1--5)},
  ytick={1,2,3,4},
  y dir=reverse,                           
  yticklabels={
    {After reading the AI output, I feel more confident (less uncertain) in my decision.},
    {If the AI had disagreed with me on this case, I would re-evaluate and re-examine
the study.},
    {The AI’s suggestions improved my final decision.},
    {The AI did not bias me toward incorrect conclusions.}
  },
  yticklabel style={text width=.22\linewidth, align=right, font=\footnotesize},
  enlarge y limits=0.18,
  bar width=10pt,
  xmajorgrids=true,
  legend style={at={(0.5,1.02)}, anchor=south, legend columns=-1, font=\footnotesize},
   ytick pos=left, 
]

\addplot+[area legend, fill=NVGreenColor, error bars/.cd, x dir=both, x explicit] coordinates {
  (4.70,1) +- (0.57,0)
  (5.00,2) +- (0.00,0)
  (4.70,3) +- (0.47,0)
  (4.95,4) +- (0.22,0)
};

\legend{AI-assisted}
\end{axis}
\end{tikzpicture}%

\caption{The Trust \& Confidence survey (mean/std). Likert scale: 1 - Strongly Disagree, 2 - Disagree, 3 - Neither Agree Nor Disagree, 4 - Agree, 5 - Strongly Agree. }
\label{fig:results_survey_trust}
\end{figure*}

The Trust \& Confidence survey ratings are shown in Figure~\ref{fig:results_survey_trust}.
\begin{itemize}
    \item \textbf{After reading the AI output, I feel more confident (less uncertain) in my decision.} Ratings were predominantly \emph{Agree} or \emph{Strongly Agree} ($4.70 \pm 0.57$).
    \item \textbf{If the AI had disagreed with me on this case, I would re-evaluate and re-examine the study.} All responses were \emph{Strongly Agree} ($5.00 \pm 0.00$).
    \item \textbf{The AI’s suggestions improved my final decision.} All responses were \emph{Agree} or \emph{Strongly Agree} ($4.70 \pm 0.47$).
    \item \textbf{The AI did not bias me toward incorrect conclusions.} Most responses were \emph{Strongly Agree} ($4.95 \pm 0.22$).
\end{itemize}

Overall, radiologists reported favorable assessments of output quality,
report-writing efficiency, and confidence in their decisions. These findings
represent preliminary reader perceptions; diagnostic accuracy and clinical
safety require separate evaluation.

\section{Limitations}
\label{sec:limitations}

We acknowledge the following limitations:

\begin{itemize}
    \item \textbf{Hallucinations and reasoning reliability.} As with other
    generative LLM and VLM systems, NV-REASON-CT can produce findings or
    explanations that are not supported by the image. Generated reasoning
    provides a reviewable explanation, but does not establish that the text
    faithfully represents the model's internal computation.

    \item \textbf{Anatomical scope and cropping.} The model is designed for
    chest and abdominal CT and processes one fixed-size regional crop at a time.
    The automatic cropping heuristic relies on detecting a sufficient portion of
    the lungs and can select a suboptimal field of view in scans with limited or
    unusual coverage. Findings outside the selected crop cannot be recovered by
    the language model. More robust segmentation, learned localization, or
    multi-region processing may be needed for whole-body examinations and other
    anatomical regions.

    \item \textbf{Data and reasoning supervision.} Although training combines
    several CT sources and task types, it does not capture the full diversity of
    scanners, protocols, institutions, patient populations, and disease
    presentations encountered in clinical practice. Rare findings remain
    underrepresented, and report-derived labels and targets inherit ambiguity
    from their source reports. Direct expert reasoning supervision comes from a
    limited set of recorded and transcribed CT interpretations; the expanded
    corpus includes synthetic narrated interpretations derived from source
    reports using these expert examples. Larger multi-institutional datasets
    with more directly authored and reviewed expert reasoning may improve
    robustness.

    \item \textbf{Clinical context and prior examinations.} The reported
    evaluations primarily condition the model on the CT volume and task
    instruction. Clinical indications, laboratory results, patient history, and
    prior examinations are not incorporated systematically. Radiologists use
    this information to resolve uncertainty and assess interval change, and its
    principled integration is an important direction for future work.

    \item \textbf{Evaluation and clinical validation.} Automated results depend
    on dataset-specific label ontologies, report formats, and evaluator
    definitions, as well as differences in evaluation cohorts, prompting, and
    model adaptation, limiting direct comparison across studies. The reported
    evaluations assess the complete model; controlled ablations are needed to
    isolate the contributions of the spatial interface, reasoning supervision,
    and GRPO. The reader study is preliminary and reflects reader perceptions.
    Because unaided review always preceded AI-assisted review of the same case,
    recall and reading-order effects may influence the reported benefits. It is
    not a prospective assessment of clinical safety or effectiveness. Larger,
    multi-center studies are required to evaluate calibration, failure modes,
    workflow impact, and generalization in routine practice.
\end{itemize}

\section{Conclusion}
\label{conclusion}

We introduced NV-REASON-CT, a generative vision--language model for native 3D
interpretation of chest and abdominal CT. The model uses a native 3D vision
encoder and passes all visual tokens, together with their explicit 3D positions,
to the language model for joint processing with text. This architecture
preserves volumetric spatial context throughout the vision--language interface
and supports abnormality classification, report generation, and interactive
questions with reviewable clinical explanations.

A central contribution of this work is the curated, diverse multimodal training
corpus spanning chest and abdominal CT. It combines structured reports,
abnormality-focused and anatomy-specific questions, multi-turn interactions,
and refusal examples. A limited set of recorded and transcribed expert CT
interpretations provides direct supervision and serves as exemplars for
generating additional report-grounded synthetic narrations. These annotations
are designed to teach the model to communicate supporting observations,
diagnostic considerations, and uncertainty. Training follows two stages:
end-to-end supervised fine-tuning on the complete task mixture, followed by GRPO
using region-aware, verifiable rewards for finding agreement and report
structure.

We evaluated NV-REASON-CT across CT-RATE, Merlin, RAD-ChestCT, and a held-out
internal NIH cohort, covering classification and report generation across
different datasets, anatomical regions, and label ontologies. On CT-RATE, the
model achieved strong classification and report-generation performance without
a task-specific classification head. In a preliminary study, expert radiologists
reported increased confidence with AI-assisted review, alongside a 50\%
reduction in average reported interpretation and reporting time. By releasing
NV-REASON-CT together with its SFT and GRPO training code, we aim to support
further research and independent evaluation of expert-guided 3D CT
interpretation.

\appendix
\numberwithin{table}{section}
\renewcommand{\thetable}{\thesection\arabic{table}}
\section{Per-abnormality CT-RATE classification results}
\label{app:ctrate_classification_tables}

Tables~\ref{tab:ctrate_per_abnormality_f1} and
\ref{tab:ctrate_per_abnormality_auroc} expand the macro-level comparison in
Table~\ref{tab:ctrate_classification} by reporting F1 and AUROC, respectively,
for each of the 18 CT-RATE abnormalities. Each table is divided into two parts
to retain the complete comparator set at a readable scale. They are included
to make label-specific performance and differences in checkpoints, evaluation
units, and adaptation protocols auditable without interrupting the main Results
narrative.

\ctratePerAbnormalityFOneTables
\ctratePerAbnormalityAurocTables
\FloatBarrier

\section{Per-abnormality CT-RATE report-derived results}
\label{app:ctrate_report_per_abnormality}

Table~\ref{tab:ctrate_report_per_abnormality_f1} reports the positive-class F1
of each CT-RATE abnormality after labels are extracted from generated reports.
These values quantify finding coverage in report text and are distinct from
the direct image-classification results in
Tables~\ref{tab:ctrate_per_abnormality_f1} and
\ref{tab:ctrate_per_abnormality_auroc}. The CT2Rep, CT-CHAT, and BTB3D-16
columns reproduce BTB3D Table 5~\cite{hamamci2025btb3d}; MS-VLM reproduces
Table E.1 of that study~\cite{lee2024msvlm}; and MonteRET reproduces
Supplementary Table S4~\cite{lin2026monteret}. COLIPRI-CRM and VoxelFM do not
provide numerical per-abnormality tables, so their entries were digitized from
the authors' vector and raster figures, respectively, and should be regarded as
approximate~\cite{wald2026colipri,morenoaguado2026voxelfm}. Because these
studies use non-identical cohorts and report-extraction implementations, the
columns document available evidence rather than a controlled ranking.

\begin{table*}[!htbp]
\centering
\small
\caption{Per-abnormality positive-class F1 derived from generated reports on
CT-RATE. Columns marked
$\dagger$ were digitized from author-provided figures and are approximate; the
corresponding Macro-F1 values are the authors' reported aggregate results.}
\label{tab:ctrate_report_per_abnormality_f1}
\renewcommand{\arraystretch}{1.07}
\setlength{\tabcolsep}{4pt}
\begin{adjustbox}{max width=\textwidth}
\begin{tabular}{@{}lrrrrrrrr@{}}
\toprule
\textbf{Abnormality} & \textbf{CT2Rep} & \textbf{CT-CHAT} &
\shortstack{\textbf{BTB3D}\\\textbf{16$\times$16$\times$8}} &
\textbf{MS-VLM} &
\shortstack{\textbf{COLIPRI}\\\textbf{CRM}$^\dagger$} &
\textbf{VoxelFM}$^\dagger$ & \textbf{MonteRET} &
\shortstack{\textbf{NV-REASON-CT}\\\textbf{(ours)}} \\
\midrule
Medical material                   & 0.000 & 0.006 & 0.142 & 0.157 & 0.414 & 0.234 & 0.103 & 0.679 \\
Arterial wall calcification        & 0.322 & 0.451 & 0.414 & 0.334 & 0.682 & 0.636 & 0.700 & 0.766 \\
Cardiomegaly                       & 0.013 & 0.123 & 0.305 & 0.271 & 0.481 & 0.488 & 0.508 & 0.585 \\
Pericardial effusion               & 0.000 & 0.009 & 0.095 & 0.162 & 0.223 & 0.244 & 0.380 & 0.561 \\
Coronary artery wall calcification & 0.335 & 0.412 & 0.403 & 0.321 & 0.642 & 0.588 & 0.630 & 0.756 \\
Hiatal hernia                      & 0.074 & 0.207 & 0.164 & 0.179 & 0.199 & 0.282 & 0.480 & 0.531 \\
Lymphadenopathy                    & 0.013 & 0.069 & 0.358 & 0.319 & 0.451 & 0.311 & 0.198 & 0.545 \\
Emphysema                          & 0.198 & 0.391 & 0.196 & 0.282 & 0.422 & 0.464 & 0.420 & 0.521 \\
Atelectasis                        & 0.323 & 0.341 & 0.269 & 0.279 & 0.489 & 0.481 & 0.365 & 0.597 \\
Lung nodule                        & 0.029 & 0.443 & 0.427 & 0.499 & 0.538 & 0.605 & 0.536 & 0.687 \\
Lung opacity                       & 0.557 & 0.266 & 0.408 & 0.505 & 0.680 & 0.665 & 0.331 & 0.734 \\
Pulmonary fibrotic sequela         & 0.104 & 0.069 & 0.318 & 0.327 & 0.453 & 0.286 & 0.404 & 0.491 \\
Pleural effusion                   & 0.341 & 0.173 & 0.308 & 0.187 & 0.724 & 0.676 & 0.272 & 0.847 \\
Mosaic attenuation pattern         & 0.198 & 0.064 & 0.183 & 0.164 & 0.296 & 0.425 & 0.278 & 0.457 \\
Peribronchial thickening           & 0.099 & 0.000 & 0.125 & 0.194 & 0.267 & 0.273 & 0.239 & 0.370 \\
Consolidation                      & 0.236 & 0.120 & 0.259 & 0.287 & 0.521 & 0.569 & 0.278 & 0.682 \\
Bronchiectasis                     & 0.013 & 0.091 & 0.126 & 0.111 & 0.334 & 0.208 & 0.260 & 0.461 \\
Interlobular septal thickening     & 0.032 & 0.075 & 0.135 & 0.125 & 0.258 & 0.353 & 0.190 & 0.392 \\
\midrule
\textbf{Macro-F1}                  & \textbf{0.160} & \textbf{0.184} &
\textbf{0.258} & \textbf{0.261} & \textbf{0.449} &
\textbf{0.432} & \textbf{0.365} & \textbf{0.592} \\
\bottomrule
\end{tabular}
\end{adjustbox}
\end{table*}
\FloatBarrier

\section{Per-abnormality Merlin classification results}
\label{app:merlin_classification_table}

Table~\ref{tab:merlin_per_abnormality_f1_auroc} expands the NV-REASON-CT
summary in Table~\ref{tab:merlin_classification} by reporting the
positive-class F1 and AUROC for each of the 30 Merlin findings. Results use the
same balanced cohort and uniform prompt as the main-text result.
F1 is computed at the fixed 0.5 next-token probability threshold, and AUROC is
computed from the continuous normalized \textit{Yes} probabilities. The six
lower-thoracic findings specified in
Section~\ref{sec:results_merlin_classification} use the chest anatomical region;
the other 24 use the abdomen region. Prompt selection and this region policy
used the released test cohort, as noted in
Section~\ref{sec:results_merlin_classification}.

\begin{table*}[!htbp]
\centering
\small
\caption{Per-abnormality positive-class F1 and AUROC for NV-REASON-CT on the
balanced Merlin test cohort. F1 uses a fixed 0.5 threshold on the normalized
next-token probability of \textit{Yes}; AUROC uses the continuous probability.
The macro average is the unweighted mean across the 30 findings.}
\label{tab:merlin_per_abnormality_f1_auroc}
\renewcommand{\arraystretch}{1.06}
\setlength{\tabcolsep}{12pt}
\begin{adjustbox}{max width=\textwidth}
\begin{tabular}{@{}lrr@{}}
\toprule
\textbf{Finding} &
\shortstack{\textbf{F1 at 0.5}\\$\uparrow$} &
\textbf{AUROC} $\uparrow$ \\
\midrule
Submucosal edema                  & 0.7092 & 0.7608 \\
Renal hypodensities               & 0.6901 & 0.7307 \\
Aortic valve calcification        & 0.9078 & 0.9523 \\
Coronary calcification            & 0.8000 & 0.8447 \\
Thrombosis                        & 0.6750 & 0.6743 \\
Metastatic disease                & 0.7931 & 0.8705 \\
Pancreatic atrophy                & 0.7258 & 0.7641 \\
Renal cyst                        & 0.6779 & 0.7877 \\
Osteopenia                        & 0.8102 & 0.9164 \\
Surgically absent gallbladder     & 0.7423 & 0.8363 \\
Atelectasis                       & 0.7698 & 0.8349 \\
Abdominal aortic aneurysm         & 0.9697 & 0.9931 \\
Anasarca                          & 0.8702 & 0.9193 \\
Hiatal hernia                     & 0.7917 & 0.8582 \\
Lymphadenopathy                   & 0.6435 & 0.6591 \\
Prostatomegaly                    & 0.7273 & 0.6999 \\
Biliary ductal dilation           & 0.7010 & 0.8060 \\
Cardiomegaly                      & 0.8118 & 0.8871 \\
Splenomegaly                      & 0.8140 & 0.8982 \\
Hepatomegaly                      & 0.7347 & 0.8546 \\
Atherosclerosis                   & 0.8197 & 0.8468 \\
Ascites                           & 0.8859 & 0.9520 \\
Pleural effusion                  & 0.9236 & 0.9642 \\
Hepatic steatosis                 & 0.6800 & 0.7407 \\
Appendicitis                      & 0.5484 & 0.7331 \\
Gallstones                        & 0.7234 & 0.8220 \\
Hydronephrosis                    & 0.7292 & 0.8316 \\
Bowel obstruction                & 0.7794 & 0.9006 \\
Free air                          & 0.8132 & 0.8741 \\
Fracture                          & 0.6978 & 0.7749 \\
\midrule
\textbf{Macro average}            & \textbf{0.7655} & \textbf{0.8329} \\
\bottomrule
\end{tabular}
\end{adjustbox}
\end{table*}
\FloatBarrier

\section{Model Output Examples}
\label{app:model_output_examples}

This section presents representative outputs from NV-REASON-CT. The first
example shows the full CT volume in axial, coronal, and sagittal cross-sections,
with the square projection of the anatomy-aware
$384\times384\times384$-mm chest crop outlined in green. This cubic crop is the
image input to the model. The visualization is followed by a structured chest
CT report and a chest CT reasoning output generated from this input. User
prompts are shown in green and NV-REASON-CT outputs are shown in gray.

\begin{center}
\includegraphics[width=\textwidth]{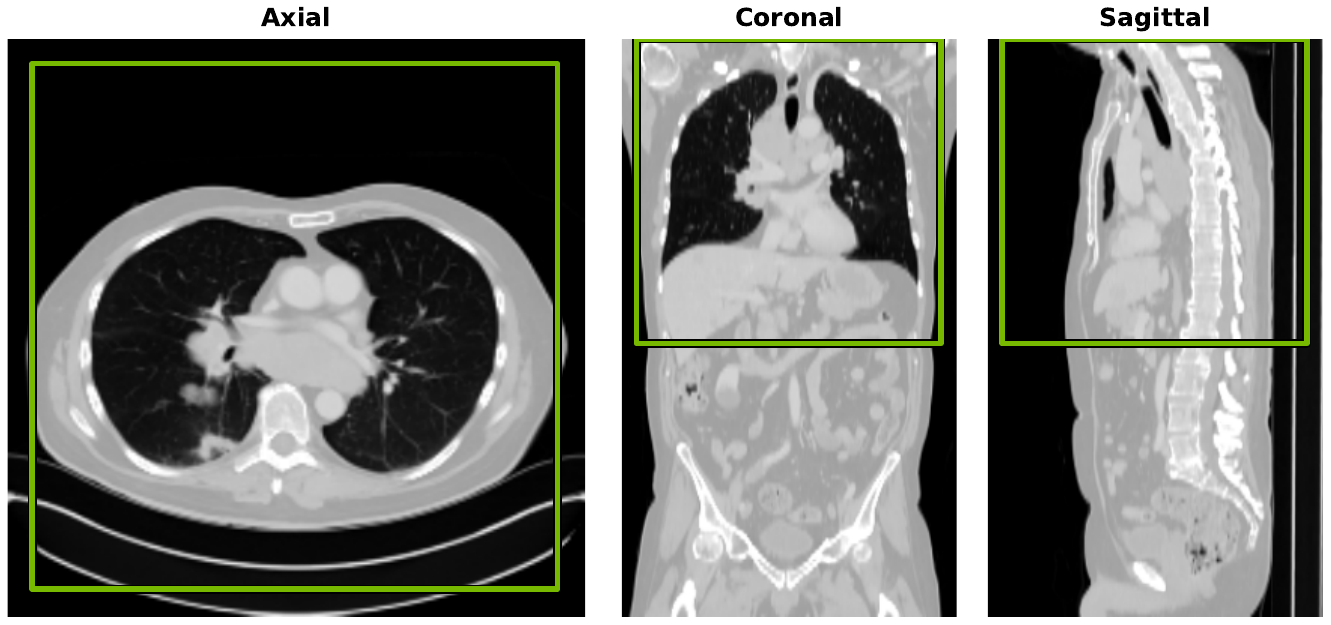}
\end{center}

\Needspace{10\baselineskip}
\subsection{Structured chest CT report}
\label{app:structured_chest_output}

\medskip
\begin{userbubble}
\small\sffamily
\textbf{User:} write a structured chest CT report
\end{userbubble}

\begin{aibubble}
\small\sffamily
\textbf{NV-REASON-CT:}

\smallskip
\texttt{\textless think\textgreater} I will produce a structured chest CT
report with a concise impression. \texttt{\textless/think\textgreater}

\smallskip
\textbf{TECHNIQUE:} IV contrast CT.

\smallskip
\textbf{FINDINGS:}

\textbf{Medical devices:} Left-sided port catheter with the tip in the superior
vena cava.

\smallskip
\textbf{CHEST:}

\textbf{Lungs and airways:} Right lung mass with multiple satellite nodules in
the right lower lobe. Peripheral ground-glass opacities in the right lower lobe
superior segment and few small focal ground-glass opacities in the right upper
lobe. Patchy ground-glass opacities also noted in the left upper lobe and
superior segment left lower lobe. These may relate to infection or inflammation
or tumor spread. Central airways are patent.

\textbf{Pleura:} No pleural effusion or pneumothorax.

\textbf{Mediastinum and hila:} Multiple heterogeneously enhancing right
supraclavicular, mediastinal, right hilar, posterior mediastinal, paravertebral,
para-aortic, and retrocrural nodes, including bulky right supraclavicular
adenopathy and bulky subcarinal adenopathy.

\textbf{Heart and pericardium:} Normal heart size. No pericardial effusion.

\textbf{Great vessels:} Normal caliber thoracic aorta and main pulmonary artery.

\textbf{Esophagus:} No esophageal abnormality.

\textbf{Chest wall and breast:} No chest wall or breast mass.

\textbf{Bones:} Enhancing paraspinal soft tissue mass with destruction of the
T10 vertebral body.

\smallskip
\textbf{IMPRESSION:}

1. Right lung mass with multiple satellite nodules in the right lower lobe.

2. Multifocal ground-glass opacities in both lungs, which may relate to
infection or inflammation or tumor spread.

3. Extensive right supraclavicular, mediastinal, right hilar, posterior
mediastinal, paravertebral, para-aortic, and retrocrural adenopathy.

4. Enhancing paraspinal soft tissue mass with destruction of the T10 vertebral
body.
\end{aibubble}

\Needspace{12\baselineskip}
\begin{userbubble}
\small\sffamily
\textbf{User:} Could the bilateral ground-glass opacities represent infection
rather than tumor spread? Explain the evidence and uncertainty.
\end{userbubble}

\begin{aibubble}
\small\sffamily
\textbf{NV-REASON-CT:}

\smallskip
\texttt{\textless think\textgreater} The question asks whether infection is a
plausible explanation for the ground-glass opacities. Ground-glass opacity can
reflect inflammation, infection, edema, hemorrhage, or tumor-related change
depending on context. I inspect the CT image and see multifocal bilateral
ground-glass opacities with a right lung mass and extensive adenopathy. The
imaging pattern is not specific, so infection cannot be confirmed from CT
alone. \texttt{\textless/think\textgreater}

\smallskip
Yes, infection is a plausible possibility, especially for the peripheral
ground-glass opacities. However, the CT appearance is nonspecific, and
tumor-related inflammation or spread cannot be excluded given the lung mass
and extensive adenopathy.
\end{aibubble}

\Needspace{12\baselineskip}
\begin{userbubble}
\small\sffamily
\textbf{User:} The patient has fever and cough. How does this new clinical
information change your assessment?
\end{userbubble}

\begin{aibubble}
\small\sffamily
\textbf{NV-REASON-CT:}

\smallskip
\texttt{\textless think\textgreater} The question asks how symptoms affect the
differential. Fever and cough can be seen with infection, inflammatory tumor
spread, or treatment-related change. I review the CT and see bilateral
ground-glass opacities along with a right lung mass and extensive adenopathy.
Imaging alone cannot determine the exact cause of the symptoms. The new
clinical information makes infection a real consideration but does not exclude
malignancy. \texttt{\textless/think\textgreater}

\smallskip
It raises infection as a plausible explanation for the bilateral ground-glass
opacities, especially if clinically apparent. However, the CT also shows a
right lung mass and extensive adenopathy, so tumor-related spread or
treatment-related change remains possible.
\end{aibubble}

\Needspace{12\baselineskip}
\begin{userbubble}
\small\sffamily
\textbf{User:} Where is the dominant lung mass?
\end{userbubble}

\begin{aibubble}
\small\sffamily
\textbf{NV-REASON-CT:}

\smallskip
\texttt{\textless think\textgreater} The question asks for the location of the
main lung mass. I review the lung windows and focus on the dominant mass-like
abnormality. The CT shows the main mass centered in the right lower lobe. There
are also satellite nodules in the right lower lobe and additional small focal
opacities elsewhere. \texttt{\textless/think\textgreater}

\smallskip
The dominant lung mass is in the right lower lobe, with several satellite
nodules in the same lobe.
\end{aibubble}

\Needspace{12\baselineskip}
\begin{userbubble}
\small\sffamily
\textbf{User:} Which nodal regions contain lymphadenopathy?
\end{userbubble}

\begin{aibubble}
\small\sffamily
\textbf{NV-REASON-CT:}

\smallskip
\texttt{\textless think\textgreater} The question asks for the distribution of
enlarged lymph nodes. I inspect the mediastinum, hila, supraclavicular regions,
and posterior mediastinum on the CT. The CT shows extensive adenopathy in
multiple thoracic and lower neck stations. The most prominent nodal disease is
in the right supraclavicular and subcarinal regions.
\texttt{\textless/think\textgreater}

\smallskip
Lymphadenopathy is present in the right supraclavicular, mediastinal, right
hilar, posterior mediastinal, paravertebral, para-aortic, and retrocrural
regions, with bulky right supraclavicular and subcarinal adenopathy.
\end{aibubble}

\Needspace{12\baselineskip}
\begin{userbubble}
\small\sffamily
\textbf{User:} What is the most likely differential for the right lung mass?
\end{userbubble}

\begin{aibubble}
\small\sffamily
\textbf{NV-REASON-CT:}

\smallskip
\texttt{\textless think\textgreater} The question asks for likely causes of the
right lung mass. I review the CT pattern, including the right lower lobe mass,
satellite nodules, multifocal ground-glass opacities, extensive adenopathy, and
destructive paraspinal soft tissue. This combination is most concerning for
malignancy with nodal and possible metastatic disease. Infection or
inflammatory disease could contribute to the ground-glass opacities, but would
be less likely to explain the mass and extensive adenopathy together.
\texttt{\textless/think\textgreater}

\smallskip
The leading consideration is primary lung malignancy with nodal and metastatic
disease. Metastatic disease from another primary is another possibility.
Infection or inflammatory disease may contribute to the multifocal ground-glass
opacities, but the mass pattern and extensive adenopathy are more concerning for
malignancy.
\end{aibubble}

\Needspace{10\baselineskip}
\subsection{Chest CT reasoning output}
\label{app:chest_reasoning_output}

\medskip
\begin{userbubble}
\small\sffamily
\textbf{User:} full chest CT reasoning analysis
\end{userbubble}

\begin{aibubble}
\footnotesize\sffamily
\textbf{NV-REASON-CT:}

\smallskip
\texttt{\textless think\textgreater} I will review the CT volume systematically,
use the valid technique context, identify the key imaging findings, and then
provide the final labels. \texttt{\textless/think\textgreater}

\smallskip
We are going to review this IV contrast-enhanced CT of the chest in a systematic
way. First, one important technical and support-device detail: there is a
left-sided port catheter, and the catheter tip terminates in the superior vena
cava, which is an expected central venous position.

I will begin with the lungs and airways. The central airways are patent, so there
is no major tracheal or main bronchial occlusion. However, when we come down into
the right lung, the right lower lobe is the dominant abnormality. There is a
large mass-like process in the right lower lobe, and it is associated with
partial collapse of the right lower lobe. This is a meaningful finding because a
mass with postobstructive or associated volume loss is concerning for a
malignant process.

Looking more carefully at the right lower lobe abnormality, it is not just simple
atelectatic linear change. There is a moderate-sized mass component, and there
are adjacent patchy opacities around it. Some of this could represent surrounding
inflammatory change, postobstructive change, or adjacent small nodular disease.
It is not easy to separate all of these components perfectly, but the overall
morphology is clearly abnormal and concerning.

In the same abnormal right lower lobe region, there are a few small
calcifications associated with the mass. Calcification can sometimes be seen in
benign processes, but in this setting, because there is a large mass and
extensive thoracic adenopathy, these calcifications do not make the finding
reassuring. They are probably best considered part of the same pathologic
process rather than a separate benign feature.

Now we move to the left lung. There are several patchy ground-glass opacities in
the left lung. These are not dense mass-like consolidations. The appearance is
more nonspecific and could be inflammatory or infectious in nature. I would
favor an inflammatory-type process based on the patchy ground-glass morphology,
although in the context of a suspected malignancy, it is important to
acknowledge that imaging alone cannot be completely definitive.

At the apices, there is biapical scarring. This has a chronic-appearing pattern
and is a common incidental-type finding. I do not see this as the main clinically
significant thoracic issue. There is no pneumothorax, and there is no pleural
effusion, which is important because despite the large right lower lobe process,
there is no associated pleural fluid collection or pleural air.

Next, we should evaluate the mediastinum and hilar regions. There is bulky
mediastinal adenopathy. This includes enlarged lymph nodes in the right
paratracheal region and subcarinal region. These are not just tiny incidental
lymph nodes; they are bulky and clinically significant, and in the setting of
the right lower lobe mass, they are concerning for nodal metastatic disease.

The right hilum is also abnormal, with bulky right hilar adenopathy. This is
important because the dominant right lower lobe mass and right hilar nodal
disease together create a pattern that is suspicious for thoracic malignancy
with nodal involvement. The left lung has patchy ground-glass opacities, but the
dominant nodal disease is on the right side.

Now looking at the heart and pericardium, the heart size is within normal limits.
There is no pericardial effusion. The thoracic aorta and main pulmonary artery
are normal in caliber. There is a seven-vessel aortic arch, and the aortic root
has a normal three-vessel branching pattern. I do not see an acute great-vessel
abnormality.

The esophagus does not show a gross abnormality on this exam. The chest wall and
breast soft tissues also do not show a discrete mass. So the major soft tissue
abnormality is really centered in the right lower lobe lung and the mediastinal
and right hilar lymph node stations rather than the superficial chest wall.

Finally, we need to check the bones carefully. There are multiple sclerotic
lesions involving the thoracic spine, including lesions at the upper thoracic
vertebral body region and lower thoracic vertebral body region. In this imaging
context, these are concerning for osseous metastatic disease. It is also
important to stress that there is no visible paravertebral soft tissue component,
so there is no obvious extending extraosseous mass along those lesions.

Overall, the most significant thoracic findings are a large right lower lobe
mass with partial right lower lobe collapse, bulky right hilar and mediastinal
adenopathy, and sclerotic thoracic vertebral lesions concerning for metastatic
involvement. There are also several patchy left lung ground-glass opacities,
favored to be inflammatory or infectious in appearance. The overall thoracic
pattern is suspicious for malignancy with nodal disease and osseous metastases.
\end{aibubble}

Next, we show the abdominal crop example from the same full CT volume. The
green box marks the square projection of the anatomy-aware
$384\times384\times384$-mm abdominal crop used as model input. The structured
abdominal CT report and abdominal CT reasoning output follow.

\begin{center}
\includegraphics[width=\textwidth]{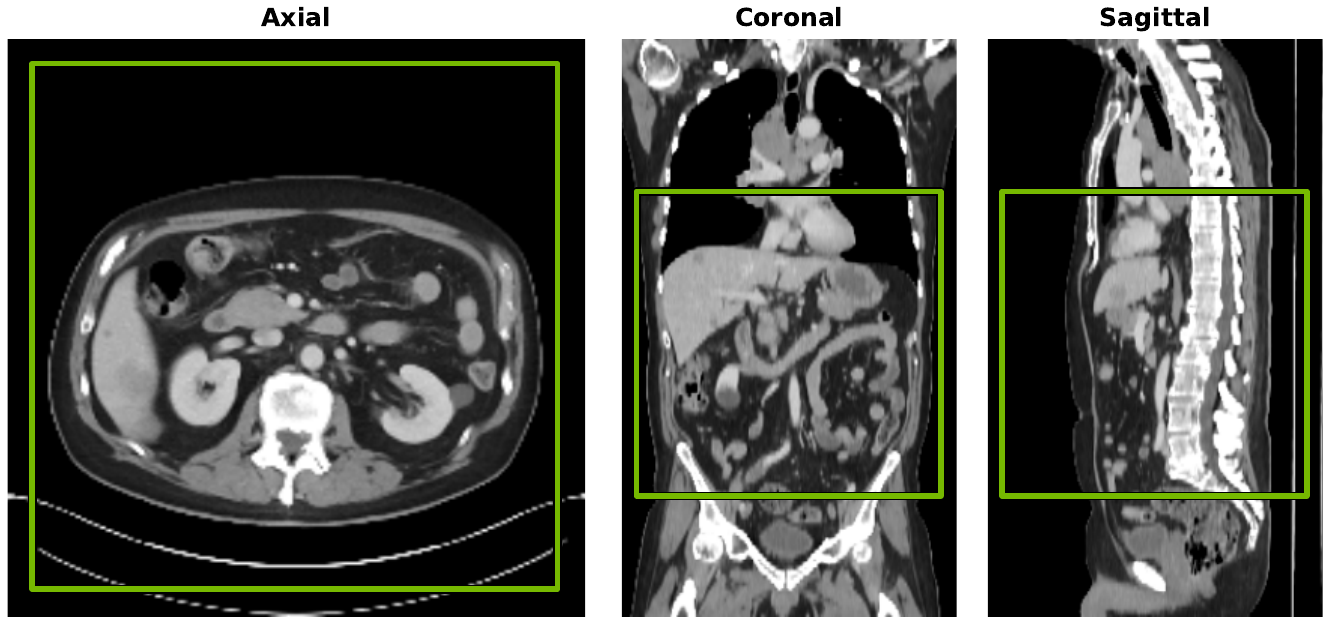}
\end{center}

\Needspace{10\baselineskip}
\subsection{Structured abdominal CT report}
\label{app:structured_abdomen_output}

\medskip
\begin{userbubble}
\small\sffamily
\textbf{User:} write a structured abdominal CT report
\end{userbubble}

\begin{aibubble}
\small\sffamily
\textbf{NV-REASON-CT:}

\smallskip
\texttt{\textless think\textgreater} I will produce a structured abdominal CT
report with a concise impression. \texttt{\textless/think\textgreater}

\smallskip
\textbf{TECHNIQUE:} IV contrast CT.

\smallskip
\textbf{FINDINGS:}

\textbf{Medical devices:} None present.

\smallskip
\textbf{ABDOMEN:}

\textbf{Liver and biliary system:} Multiple hypoattenuating foci within the
liver, including a small lesion within the left hepatic lobe and larger lesions
within the right hepatic lobe, compatible with metastasis. No biliary ductal
dilation. Gallbladder is unremarkable.

\textbf{Spleen:} Normal spleen.

\textbf{Pancreas:} Normal pancreas. No pancreatic ductal dilation.

\textbf{Adrenal glands:} Enhancing bilateral adrenal masses, including a
moderate-sized right adrenal mass and a moderate-sized left adrenal mass.

\textbf{Kidneys and urinary tract:} Right lower pole renal cyst. Additional
small hypodensities within the kidneys are too small to fully characterize. No
urinary calculus or hydronephrosis.

\textbf{Bowel and appendix:} No evidence of bowel obstruction or inflammation.

\textbf{Peritoneum and omentum:} Peritoneal nodule in the left upper quadrant.
No ascites or pneumoperitoneum.

\textbf{Lymph nodes:} Multiple prominent mesenteric lymph nodes, including a
small mesenteric lymph node, reactive versus metastatic.

\textbf{Vessels:} Normal caliber abdominal aorta. The major vascular structures
are patent. No abdominal vascular calcification or thrombosis.

\textbf{Abdominal wall and soft tissues:} No abdominal wall hernia or soft
tissue mass.

\textbf{Bones:} Aggressive appearing lytic lesion within the right iliac bone
involving the sacroiliac joint. Additional lytic lesion within the left iliac
bone.

\smallskip
\textbf{IMPRESSION:}

1. Aggressive appearing right iliac lytic lesion involving the sacroiliac joint,
highly suspicious for metastasis. Additional left iliac lytic lesion.

2. Enhancing bilateral adrenal masses compatible with metastasis.

3. Multiple hypoattenuating hepatic lesions compatible with metastasis.

4. Peritoneal nodule in the left upper quadrant.

5. Multiple prominent mesenteric lymph nodes, reactive versus metastatic.
\end{aibubble}

\Needspace{12\baselineskip}
\begin{userbubble}
\small\sffamily
\textbf{User:} Are the liver lesions solitary or multiple?
\end{userbubble}

\begin{aibubble}
\small\sffamily
\textbf{NV-REASON-CT:}

\smallskip
\texttt{\textless think\textgreater} The question asks about the number and
distribution of liver lesions. Multiple hypoattenuating liver lesions on
contrast CT can suggest metastases, especially when distributed in both lobes.
I inspect the CT image and see multiple non-cystic hypoattenuating foci. They
involve both the right and left hepatic lobes, with a dominant right hepatic
lesion. \texttt{\textless/think\textgreater}

\smallskip
Multiple. The liver lesions are distributed in both the right and left hepatic
lobes, with a dominant lesion in the right lobe.
\end{aibubble}

\Needspace{12\baselineskip}
\begin{userbubble}
\small\sffamily
\textbf{User:} What is the most likely differential for the liver lesions?
\end{userbubble}

\begin{aibubble}
\small\sffamily
\textbf{NV-REASON-CT:}

\smallskip
\texttt{\textless think\textgreater} The question asks for the leading
diagnostic possibilities for the liver lesions. I review the CT pattern,
including multiple hypoattenuating liver lesions, enhancing adrenal masses,
peritoneal nodule, and lytic iliac bone lesions. On CT, this combination most
strongly suggests metastatic malignancy. Multifocal primary hepatic malignancy
or inflammatory lesions would be less favored from the overall pattern alone.
\texttt{\textless/think\textgreater}

\smallskip
The leading consideration is metastatic malignancy, given the multiple liver
lesions plus adrenal, peritoneal, and iliac bone lesions. Less likely
possibilities, depending on clinical context, could include multifocal primary
hepatic malignancy or inflammatory lesions, but the CT pattern most favors
metastases.
\end{aibubble}

\Needspace{12\baselineskip}
\begin{userbubble}
\small\sffamily
\textbf{User:} which associated findings support it?
\end{userbubble}

\begin{aibubble}
\small\sffamily
\textbf{NV-REASON-CT:}

\smallskip
\texttt{\textless think\textgreater} The question asks how the associated
findings strengthen the suspected diagnosis. I review the CT beyond the liver
lesions. The CT shows bilateral enhancing adrenal masses, a left upper quadrant
peritoneal nodule, prominent mesenteric lymph nodes, and iliac lytic bone
lesions. This multifocal pattern across organs and bones is highly suspicious
for metastatic disease. The answer should describe the pattern without
overclaiming tissue diagnosis. \texttt{\textless/think\textgreater}

\smallskip
The associated findings that support suspected metastatic disease include
bilateral enhancing adrenal masses, a left upper quadrant peritoneal nodule,
prominent mesenteric lymph nodes, and lytic iliac bone lesions. CT alone cannot
provide histology, but the pattern is concerning for metastases.
\end{aibubble}

\Needspace{10\baselineskip}
\subsection{Abdominal CT reasoning output}
\label{app:abdomen_reasoning_output}

\medskip
\begin{userbubble}
\small\sffamily
\textbf{User:} full abdominal CT reasoning analysis
\end{userbubble}

\begin{aibubble}
\footnotesize\sffamily
\textbf{NV-REASON-CT:}

\smallskip
\texttt{\textless think\textgreater} I will review the CT volume systematically,
use the valid technique context, identify the key imaging findings, and then
provide the final labels. \texttt{\textless/think\textgreater}

\smallskip
We are reviewing an IV contrast-enhanced CT of the abdomen and pelvis. I do not
see any medical device in the imaged abdomen. We will go through this in a
systematic abdominal search pattern, starting with the solid organs and then
moving to bowel, peritoneum, vessels, lymph nodes, abdominal wall, and bones.

First, looking at the liver, there are multiple focal lesions involving both
hepatic lobes. The dominant abnormality is in the right hepatic lobe, and it has
a central hypoattenuating component. That morphology is meaningful because it
does not look like a simple cyst; it has a masslike appearance with internal
lower attenuation. In this context, the appearance is most suggestive of
metastatic disease, with possible central necrotic change.

Now we check the biliary system and gallbladder. Despite the liver lesions, I do
not see intrahepatic or extrahepatic biliary ductal dilatation. The gallbladder
appears unremarkable, without imaging features to suggest an acute inflammatory
gallbladder process.

The spleen appears unremarkable. Moving to the pancreas, there is a mass in the
peripancreatic region. It is important to stress that there is no pancreatic
ductal dilatation, but the soft tissue mass itself is still a clinically
significant finding, especially with the other metastatic-appearing
abnormalities in the abdomen.

Next, we evaluate the adrenal glands. Both adrenal glands contain masses. In
isolation, adrenal lesions can sometimes be difficult to characterize, but when
we consider the multiple liver lesions, peripancreatic mass, renal mass, and
osseous lesions, these adrenal findings are very concerning for metastatic
involvement.

Now we look at the kidneys and urinary tract. There is a mass involving the
right kidney. I do not see hydronephrosis, and there is no urinary calculus. So
even though there is a significant renal mass, there is no obstructive urinary
pattern on this exam. The urinary tract otherwise does not show an acute
obstructive process.

We then move to the bowel. I do not see evidence of bowel obstruction. That is
important because despite the presence of peritoneal and mesenteric masses, the
bowel loops are not dilated in a way that would suggest mechanical obstruction.
There is no pneumoperitoneum, so there is no imaging evidence of free
intraperitoneal air.

Now we focus on the peritoneum and mesentery. There are multiple peritoneal and
mesenteric masses, including a dominant mass in the left upper abdomen. This
pattern is very concerning for peritoneal metastatic disease. I do not see
ascites, but the absence of ascites does not make these peritoneal masses less
meaningful; the nodular masslike deposits themselves are the key finding.

Looking at the lymph node regions, there is a large right inguinal lymph node. In
the setting of the other metastatic-appearing findings, this is also suspicious
rather than incidental. The abdominal aorta is normal in caliber, and the
abdominal vasculature appears patent.

The abdominal wall and superficial soft tissues do not show an abdominal wall
hernia or soft tissue mass. Finally, we check the bones. There are multiple
lytic and sclerotic osseous lesions, including involvement of the L2 vertebral
body and a small sclerotic lesion in the right sacrum. The mixed lytic and
sclerotic pattern is concerning for osseous metastatic disease. I do not see an
acute fracture.

Overall, the most significant findings are multiple metastatic-appearing lesions
involving the liver, a peripancreatic mass, bilateral adrenal masses, a right
renal mass, multiple peritoneal and mesenteric masses, a large right inguinal
lymph node, and mixed lytic and sclerotic bone lesions. The overall imaging
pattern is most consistent with widespread abdominal and osseous metastatic
disease.
\end{aibubble}

{
  \small
  \bibliographystyle{unsrt}
  \bibliography{paper}
}

\end{document}